\documentclass[10pt,journal,compsoc]{IEEEtran}

\ifCLASSOPTIONcompsoc
  \usepackage[nocompress]{cite}
\else
  \usepackage{cite}
\fi
\ifCLASSINFOpdf
\else
\fi
\usepackage{balance}
\usepackage{amssymb}
\usepackage{cite}
\usepackage{graphicx}
\usepackage{amsmath}
\usepackage{stfloats}
\usepackage{url}
\usepackage{bm}
\usepackage[latin1]{inputenc}
\usepackage{colortbl}
\usepackage{soul}
\usepackage{multirow}
\usepackage{pifont}
\usepackage{color}
\usepackage{alltt}
\usepackage{enumerate}
\usepackage{siunitx}
\usepackage{pbox}
\usepackage{textcomp}
\usepackage{gensymb}
\usepackage{float}
\usepackage{mathtools}
\usepackage{booktabs}
\usepackage{algorithm}
\usepackage{algorithmic}
\usepackage{makecell}
\usepackage[dvipsnames]{xcolor}
\usepackage{bbm}

\usepackage[colorlinks,
            linkcolor=red,
            anchorcolor=blue,
            citecolor=green
            ]{hyperref}

\usepackage{xcolor}
\usepackage[switch]{lineno}

\begin{document}

%
\title{Efficient and Robust Point Cloud Registration via Heuristics-guided Parameter Search}
%
%
%
%

\author{Tianyu Huang, Haoang Li, Liangzu Peng, Yinlong Liu, and Yun-Hui Liu,~\IEEEmembership{Fellow,~IEEE}
\IEEEcompsocitemizethanks{\IEEEcompsocthanksitem T. Huang and Y.-H. Liu are with the Department of Mechanical and Automation Engineering and T Stone Robotics Institute, The Chinese University of Hong Kong, Hong Kong. E-mail: \{tyhuang, yhliu\}@mae.cuhk.edu.hk.
\IEEEcompsocthanksitem H. Li is with the Thrust of Robotics and Autonomous Systems and the Thrust of Intelligent Transportation, The Hong Kong University of Science and Technology (Guangzhou), China. Email: haoangli@hkust-gz.edu.cn.
\IEEEcompsocthanksitem Y. Liu is with State Key Laboratory of Internet of Things for Smart City (SKL-IOTSC), University of Macau, Macau 999078, China. Email: YinlongLiu@um.edu.mo.
\IEEEcompsocthanksitem Corresponding author: Yun-Hui Liu.
}
\thanks{Code: \url{https://github.com/tyhuang98/HERE-release}}
}

\IEEEtitleabstractindextext{%
\begin{abstract}
Estimating the rigid transformation with 6 degrees of freedom based on a putative 3D correspondence set is a crucial procedure in point cloud registration.
Existing correspondence identification methods usually lead to large outlier ratios~($>$ 95 $\%$ is common), underscoring the significance of robust registration methods. 
Many researchers turn to parameter search-based strategies~(e.g., Branch-and-Bround) for robust registration. Although related methods show high robustness, their efficiency is limited to the high-dimensional search space. 
This paper proposes a heuristics-guided parameter search strategy to accelerate the search while maintaining high robustness. 
We first sample some correspondences~(i.e., heuristics) and then just need to sequentially search the feasible regions that make each sample an inlier. 
Our strategy largely reduces the search space and can guarantee accuracy with only a few inlier samples, therefore enjoying an excellent trade-off between efficiency and robustness.
Since directly parameterizing the 6-dimensional nonlinear feasible region for efficient search is intractable, we construct a three-stage decomposition pipeline to reparameterize the feasible region, resulting in three lower-dimensional sub-problems that are easily solvable via our strategy.
Besides reducing the searching dimension, our decomposition enables the leverage of 1-dimensional interval stabbing at all three stages for searching acceleration. Moreover, we propose a valid sampling strategy to guarantee our sampling effectiveness, and a compatibility verification setup to further accelerate our search. 
Extensive experiments on both simulated and real-world datasets demonstrate that our approach exhibits comparable robustness with state-of-the-art methods while achieving a significant efficiency boost.


\end{abstract}

\begin{IEEEkeywords}
Point cloud registration, sampling, parameter search, transformation decomposition, interval stabbing.
\end{IEEEkeywords}}

\maketitle

\IEEEdisplaynontitleabstractindextext

%
\IEEEpeerreviewmaketitle

\ifCLASSOPTIONcompsoc
\IEEEraisesectionheading{\section{Introduction}\label{sec:introduction}}
\else
\fi

\IEEEPARstart{T}{he} problem of rigid point cloud registration refers to estimating the Euclidean/rigid transformation relationship between two 3D point clouds~\cite{tam2012registration}. 
This problem finds a lot of applications in computer vision and robotics, e.g., 3D reconstruction~\cite{huber2003fully}, scene understanding~\cite{belongie2002shape}, robot manipulation~\cite{ten2017grasp}, and mobile robot navigation~\cite{xu2022fast}, etc.


A commonly used pipeline to solve the rigid point cloud registration problem consists of two main procedures: \textit{correspondence identification} and \textit{robust registration}~\cite{fischler1981random, bustos2017guaranteed, yang2020teaser}.
Correspondence identification refers to establishing putative correspondences between two point clouds, and this can be done by adopting handcrafted~\cite{rusu2009fast, salti2014shot} or deep learning-based~\cite{choy2019fully, huang2021predator} point cloud descriptors.
If there are no false matches~(i.e., outliers) in the correspondences, then the rigid transformation with 6 degrees of freedom~(DoF) can be easily solved in closed forms~\cite{arun1987least}. 
Nevertheless, the existence of outliers is inevitable due to the occlusions, noise, and unorganized density in point clouds~\cite{tombari2013performance}.
Therefore one often needs the second step, robust registration, that is to estimate the transformation accurately even if there is a large fraction of outliers.
Many practices have been made to tackle the robust registration problem based on consensus maximization. 
These methods try to find the rigid transformation that fits the largest number of correspondences. 
They have shown impressive results in terms of either efficiency or robustness~\cite{fischler1981random, bustos2017guaranteed}, but rarely both.
RANSAC~\cite{fischler1981random} is one of the most well-known heuristic strategies for consensus maximization and has inspired a lot of variants~\cite{barath2018graph, quan2020compatibility}. 
This strategy iteratively samples minimal correspondence sets for candidate transformations and selects the solution that produces the largest consensus set.
While RANSAC is generally efficient at each iteration, it can not guarantee estimation robustness in case of high outlier ratios due to sampling uncertainty.

To achieve highly-robust registration, many researchers turn to the parameter search.
The representative Branch-and-Bound~(BnB) strategy recursively searches the whole parameter space for the solution and can guarantee global optimality in terms of maximizing the size of the consensus set~\cite{yang2015go}, \cite{chen2022deterministic}. 
However, the 6-DoF high-dimensional space causes low searching efficiency~\cite{yang2015go}. 
To overcome this problem, a widely-used strategy is to decompose the original 6-DoF and search the lower-dimensional spaces separately~(e.g., independently estimating the 3-DoF rotation and 3-DoF translation)~\cite{yang2020teaser, liu2018efficient}. 
Nevertheless, most existing decomposition methods increase the problem size quadratically and result in limited efficiency gains. 
Some methods~\cite{bustos2017guaranteed, zhang2023accelerating} are proposed to embed the 1-dimensional interval stabbing strategy~\cite{preparata2012computational} into BnB for acceleration. With a single sorting and scanning operation, the interval stabbing is quite fast and can achieve global optimality in terms of 1D search~(see Algorithm~\ref{alg: interval_stabbing}).
While reducing 1 dimension could bring certain computational gains, the remaining degrees of freedom still pose a serious challenge.

\begin{figure}[!t]
    \centering
    \includegraphics[width=0.26\textwidth]{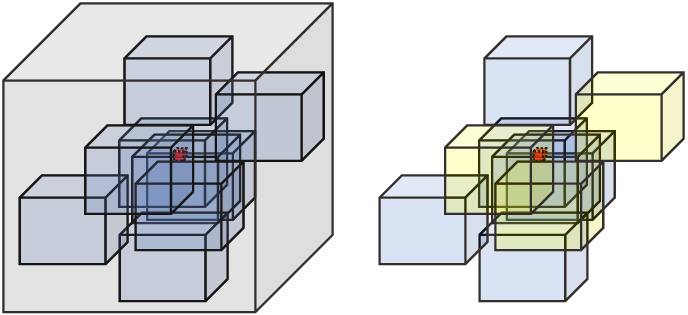}
    \\
    \makebox[0.14\textwidth]{\footnotesize (a)}
    \makebox[0.14\textwidth]{\footnotesize (b)}
    \\[-0.9em]
    \caption{Illustration of the proposed heuristics-guided parameter search strategy. The large \textcolor{Gray}{gray} block represents the whole parameter space. Each middle \textcolor{CadetBlue}{blue}/\textcolor{GreenYellow}{yellow} block represents the feasible region that makes each correspondence an inlier\protect\footnotemark[1]. The small \textcolor{red}{red} block represents the optimal region that fits the most correspondences. (a) The parameter search-based methods~\cite{yang2015go, liu2018efficient, chen2022deterministic} generally search the whole parameter space~(i.e., the large \textcolor{Gray}{gray} block) to find the optimal solution and lead to unsatisfactory efficiency. (b) Our heuristics-guided parameter search method integrates the sampling strategy into parameter search and only needs to search the \textcolor{GreenYellow}{yellow} blocks associated with the samples.} 
    \label{fig: combine_sam_search}
    \vspace{-0.5em}
\end{figure}

Given the high efficiency of sampling-based heuristics and the high robustness of BnB-based parameter search, would it be possible to find something in between that can combine the best of both? With this motivation, we propose a \textit{heuristics-guided parameter search} strategy~(see Fig.~\ref{fig: combine_sam_search}) that enjoys an excellent trade-off between efficiency and robustness.
Specifically, we first sample a specific number of correspondences and then sequentially search the feasible regions that make each sample an inlier.
Compared with the pure search-based methods that search the whole parameter space, our strategy largely reduces the search space and thus is more efficient.
Moreover, once a few samples are inliers, our search regions can contain the solution that falls in the optimal solution region, therefore maintaining high robustness of our approach to sampling uncertainty.
Similar ideas of combining heuristics and parameter search are not rare in other areas~\cite{rossi2015parallel, yiu2004hybrid}, but to our best knowledge, this work first leverages such combination for robust registration.

\footnotetext[1]{In most cases, the feasible region is highly nonlinear and complicated. Here we use the block for simplicity.}

The high-level idea of heuristics-guided parameter search, described above, is fairly straightforward, but crafting an efficient implementation based on this idea is far from trivial. 
For example, parameterizing the feasible region of each sample for efficient search is very difficult if not impossible, as the 6-DoF transformation is highly nonlinear.
We overcome this difficulty by constructing a three-stage decomposition pipeline to reparameterize the feasible region and embed our strategy into the solving progress. In particular, we decompose the original 6-DoF registration problem into three sub-problems with respect to 3-DoF translation, 2-DoF rotation axis, and 1-DoF rotation angle, respectively. 
Given the initial correspondence set, we conduct progressive outlier removal by solving the three sub-problems in turn. 
Our decomposition brings three strengths. 
First, the feasible regions of each correspondence in three sub-problems can be easily parameterized for efficient search, so that our heuristics-guided parameter search can be applied. 
Second, the feasible regions in each sub-problem have specific properties, so that the search part at all three stages can be accelerated by interval stabbing.  
Third, apart from having lower searching dimensions, all three sub-problems in our pipeline avoid increasing the problem size, which ensures the efficiency of our method.

To guarantee the effectiveness of our heuristic part, we propose a valid sampling strategy by leveraging the spatial compatibility constraint of correspondences. Particularly, we design a novel priority computation approach to assign each correspondence a priority. Those correspondences with higher priorities hold higher probabilities to be the inliers.
And we choose the samples with the highest priorities.
Compared to random sampling, our valid sampling strategy can boost the performance of our method in terms of both efficiency and robustness. 
Moreover, we leverage spatial compatibility to improve the search part of our method. 
In particular, we set a compatibility verification before our search on the feasible region of each sampled correspondence. 
This design introduces the constraint of spatial compatibility in our consensus maximization and reduces the problem size by verification, leading to improvements in both robustness and efficiency.

Above all, our main contributions are as follows:
\begin{itemize}
\item (\textit{High-level Idea}) We propose a heuristics-guided parameter search strategy for robust registration. Our strategy largely reduces the search space and can guarantee accuracy with only a few inlier samples~(i.e., heuristics), therefore enjoying a great trade-off between efficiency and robustness.
\item (\textit{Acceleration by Decomposition}) We construct a three-stage decomposition pipeline that results in three lower-dimensional sub-problems, so as to ease the integration of our heuristics-guided parameter search into the solving progress; meanwhile, our decomposition enables the application of interval stabbing at all three stages for searching acceleration.
\item (\textit{Performance Boost}) We propose a valid sampling strategy for the heuristic part and a compatibility verification strategy for the search part. Based on the spatial compatibility constraint of correspondences, both strategies boost the performance of our method in terms of efficiency and robustness.
\end{itemize}

Extensive experiments demonstrate that compared with state-of-the-art robust registration methods, our approach can achieve comparable robustness with a significant increase in efficiency. Significantly, compared to the parameter search-based baselines~\cite{bustos2017guaranteed, chen2022deterministic}, our approach can exhibit up to $10^2\times$ and sometimes exceeding $10^3\times$ speed-up.

The rest of this paper is organized as follows. Section~\ref{sec: related} reviews the related robust registration methods based on heuristics, parameter search, and other strategies. 
Section~\ref{sec: algo_overview} gives a short overview of our method.
Section~\ref{sec: formu} presents the problem formulation and our three-stage decomposition strategy. 
Section~\ref{sec: Muiti-stage} illustrates how we integrate our heuristics-guided parameter search into the solving progress of the decomposed sub-problems. 
Section~\ref{sec: spatial} introduces details of the proposed valid sampling strategy and compatibility verification procedure. 
Section~\ref{sec: complexity} gives the time complexity analysis of our method.
Section~\ref{sec: exper} presents extensive experimental results on both simulated and real-world datasets. 
Finally, we give a conclusion in Section~\ref{sec: conc}.

\section{Related Works}
\label{sec: related}
In this section, we review existing pure heuristics-based and parameter search-based methods for robust registration. Since some methods can not be simply divided by the above two categories, we review them as other methods.

\noindent \textbf{Heuristics-based Methods.}
The heuristic strategy involves utilizing specific rules to select candidate inlier set(s) for initial transformation(s), followed by iteratively refining the registration result until finding a good-enough solution. 
One of the most well-known heuristic strategies is the RANSAC proposed by Fischler and Bolles~\cite{fischler1981random}. 
This strategy demonstrates exceptional performance in several geometric registration problems~\cite{kneip2014opengv}. 
Ever since the advent of RANSAC, many variants are proposed to improve it in terms of sampling~\cite{chum2005matching, barath2019progressive}, inlier/outlier threshold~\cite{torr2000mlesac, barath2019magsac}, and local optimization~\cite{chum2003locally, barath2018graph}, etc. 
However, a common drawback of the RANSAC-family methods is that they cannot ensure a good balance between efficiency and robustness when dealing with high outlier ratios. 
While each iteration of these methods is efficient, the iteration number needs to increase exponentially to guarantee accuracy when the outlier ratio is high. 


Instead of random sampling, some researchers employ specific metrics to select candidate consensus sets. 
A widely-used metric is spatial compatibility~\cite{bustos2019practical, quan2020compatibility}, which leverages the inherent pairwise constraint between two correspondences. 
Yang et al.~\cite{yang2021sac} propose SAC-COT that adjusts random sampling to select three compatible correspondences at each iteration. 
SAC-COT achieves higher robustness than RANSAC thanks to the more reliable minimal sets. 
However, SAC-COT shares the same drawbacks as RANSAC due to uncontrollable iteration numbers. Parra et al.~\cite{parra19pairwise} propose a practical maximum clique algorithm to find the largest set of compatible correspondences and achieve high accuracy. 
Later, Zhang et al.~\cite{Zhang_2023_CVPR} loosen the maximum clique constraint to selecting multiple maximal cliques and further improve the registration robustness. 
Even though their method achieves remarkable accuracy and can be accelerated by certain strategies~(e.g., graph sparsification), their efficiency could be limited when dealing with large-scale (inlier) correspondences since listing all the maximal cliques can be time-consuming in such cases~\cite{Tomita-TCS2006, eppstein2010listing}.
Recently, Chen et al.~\cite{chen2022sc2} introduce a method called SC$^2$-PCR that leverages the second-order spatial compatibility~(SC$^2$) measure to distinguish the consensus inlier set and the outliers. 
While achieving high accuracy, its construction of the SC$^2$ matrix undermines the efficiency when handling large-scale correspondences. 

\noindent \textbf{Parameter Search-based Methods.}
One of the representative classical robust registration techniques based on parameter search is the Hough Transform~(HT)~\cite{hough1962method}. 
Its core idea is to discretize the whole parameter space into a set of bins and select the bin accumulated with the largest support of given correspondences as the solution~\cite{woodford2014demisting, sun2014hough, Chin2017TheMC}. 
However, discretizing the parameter space and finding the optimal bin require significant tuning, which limits the robustness of these methods. 

Recently, researchers have shown great interest in the branch-and-bound (BnB) strategy since BnB can provide the globally optimal solution~\cite{Chin2017TheMC, yang2015go, Campbell_2016_CVPR, cai2019practical, Huang-CVPR2024}. 
Instead of directly searching and voting like HT, BnB recursively partitions the parameter space into smaller branches and prunes those branches that cannot contain the optimal solution by checking the bounds. 
To the best of our knowledge, Yang et al.~\cite{yang2015go} first propose a nested BnB algorithm to solve the point cloud registration problem and their algorithm can achieve global optimality. 
However, the 6-DoF high-dimensional search space leads to exponential time complexity of the BnB in their method. 
Latter, Parra and Chin~\cite{bustos2017guaranteed} propose GORE that combines BnB and 1-Dimensional interval stabbing to perform guaranteed outlier removal based on the 3-DoF rotation constraint. 
Liu et al.~\cite{liu2018efficient} introduce a rotation-invariant constraint to enable the BnB-based search on only 3-DoF translation parameter space. 
These two methods reduce the original 6-DoF problem to a 3-DoF problem, which improves the efficiency of BnB to some extent. 
However, their decomposition strategies both increase the problem size quadratically and therefore limit the efficiency gain. 
More recently, Chen et al.~\cite{chen2022deterministic} propose to decouple the original problem into a (2+1)-DoF sub-problem and a (1+2)-DoF sub-problem. 
These two sub-problems are then sequentially solved by BnB. 
While their decomposition does not increase the problem size, it is still challenging to make BnB efficient for their 3-DoF sub-problems. 

\noindent \textbf{Other Methods.}
Except for the aforementioned methods, one representative category of the other methods resorts to the M-estimation~\cite{le2019deterministic, 10091912, 9528069}. 
Related methods substitute the original least squares objective function with robust functions that are less sensitive to outliers~\cite{sidhartha2023adaptive, Peng-NeurIPS2022,Peng_2023_CVPR} (see \cite[Section 2.4]{Peng-arXiv2023b}). 
Zhou et al.~\cite{zhou2016fast} propose FGR that adopts the Geman-McClure cost function and graduated non-convexity~(GNC) to directly solve the 6-DoF transformation. While FGR shows high efficiency, it fails to guarantee estimation accuracy in case of high outlier ratios. 
Recently, Sidhartha et al.~\cite{sidhartha2023adaptive} propose an adaptive annealing schedule for GNC and achieve higher estimation robustness.
Yang et al.~\cite{yang2020teaser} propose TEASER++ that combines the ideas of consensus maximization and M-estimation. 
Based on the consensus set provided by the maximal clique inlier selection, TEASER++ designs a truncated least squares cost function to formulate the optimization problem. 
It then leverages GNC and adaptive voting to solve the rotation and translation, respectively. 
Despite that TEASER++ is currently one of the fastest robust registration algorithms dealing with high outlier ratios, its maximal clique selection module is time-consuming in case of large-scale inlier correspondences.

Some researchers turn to deep learning for robust registration~\cite{choy2020deep, pais20203dregnet, bai2021pointdsc, lee2021deep, yao2023hunter, jiang2023robust}. 
Choy et al.~\cite{choy2020deep} employs an end-to-end convolutional network to directly predict the inlier confidence of each correspondence and design a differentiable weighted Procrustes algorithm for pose estimation. Later, Lee et al.~\cite{lee2021deep} introduce Hough Voting to cluster consensus sets for votes in the 6D parameter space and design a fully convolutional module to refine the noisy votes. To employ the spatial compatibility property for better alignment, Bai et al.~\cite{bai2021pointdsc} propose PointDSC and formulate a differentiable spectral matching module supervised by spatial consistency to estimate the inlier confidence. Based on the framework of PointDSC, Jiang et al.~\cite{jiang2023robust} design a recurrent network that can aggregate discriminative geometric context information for better inlier/outlier classification. 
While these methods generally perform well on data similar to those in the training sets, their robustness to the unseen data is still restricted~\cite{chen2022sc2, chen2022deterministic}. Meanwhile, these learning-based methods usually require plenty of data for pre-training. In this paper, we focus on non-learning-based geometric methods for robust registration.

\section{Algorithm Overview}
\label{sec: algo_overview}
We aim to find some 6-DOF transformation that fits as many inliers as possible in the putative 3D correspondence set. Instead of searching the whole 6D high-dimensional parameter space for the solution, we propose a three-stage decomposition pipeline and a heuristics-guided parameter search strategy to accelerate the search. 
Our decomposition leads to three lower-dimensional sub-problems~(see Section~\ref{sec: formu}).
We then apply our heuristics-guided parameter search strategy in the solving progress of these sub-problems. When conducting the search at the first and second stages, we first sample some correspondences~(i.e. heuristics) and then sequentially search their feasible regions for the largest consensus set in all correspondences~(see Sections~\ref{subsec: te} and~\ref{subsec: re}). 
To guarantee enough inlier samples, we propose a valid sampling strategy related to assigning each correspondence a priority~(see Section~\ref{sec: spatial}). 
Then at the third stage with only 1-DoF constraint, a single interval stabbing~(see Algorithm~\ref{alg: interval_stabbing}) operation is enough for solving the final consensus set~(see Section~\ref{subsec: thetae}). 
And we can get the final registration result by simply performing Singular Value Decomposition~(SVD)~\cite{arun1987least} on the final consensus set.

\begin{figure*}[!t]
    \centering
    \includegraphics[width=0.86\textwidth]{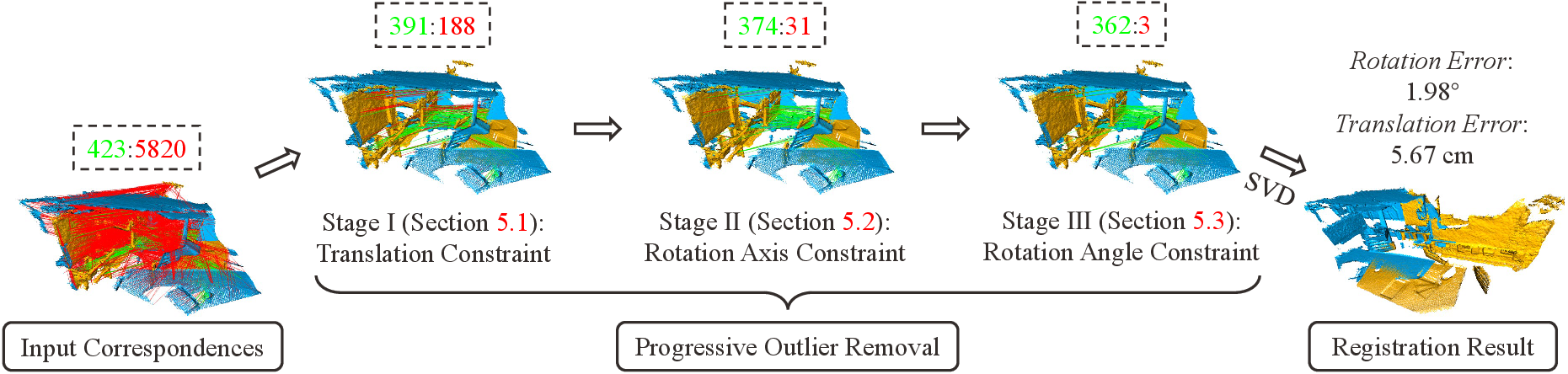}
    \vspace{-1em}
    \caption{Pipeline of the proposed point cloud registration approach. 
    Given a set of putative correspondences, we decompose the original 6-DoF transformation problem into three sub-problems and conduct progressive outlier removal by solving the sub-problems in turn. \textcolor{green}{Green} and \textcolor{red}{red} values denote the numbers of inlier and outlier correspondences, respectively.}
    \label{fig:outline}
\end{figure*}

\section{Problem Formulation and Decomposition}
\label{sec: formu}

\subsection{Consensus Maximization Formulation}
\label{subsec: consen}

Given a set of putative 3D-3D point correspondences $\mathcal{P} = \{  ( \mathbf{x}_{i}, \mathbf{y}_{i}  )\}_{i=1}^N$ where $\mathbf{x}_{i}$ belongs to the source cloud $\mathcal{X}$ and $\mathbf{y}_{i}$ belongs to the target cloud $\mathcal{Y}$, the robust registration problem can be formulated as:
\begin{equation}
\begin{aligned}
\min_{\mathbf{R},\ \mathbf{t}} \ \sum_{i=1}^{N} \rho  (  \| \mathbf{y}_i -  ( \mathbf{R}\mathbf{x}_i+\mathbf{t}  )  \|  ).
\end{aligned}
\label{eq:robust_est}
\end{equation}
where $\mathbf{R}\in$ \textit{SO}(3) is the orthogonal rotation matrix, $\mathbf{t}\in \mathbb{R}^3$ is the translation vector, $\rho  ( \cdot  )$ is a robust function (e.g., M-estimation costs), and ``$ \| \cdot  \|$'' denotes the $L_2$-norm.

Problem~(\ref{eq:robust_est}) can be formulated as the consensus maximization problem~\cite{Chin2017TheMC}, i.e., 
\begin{equation}
\begin{aligned}
&\max_{\mathbf{R},\ \mathbf{t},\ \mathbb{I}\subseteq\mathcal{P}} \  \lfloor \mathbb{I}  \rfloor \\
& s.t. \  \| \mathbf{y}_i -  ( \mathbf{R}\mathbf{x}_i+\mathbf{t}  )  \| \le \xi , \ \forall (\mathbf{x}_i, \mathbf{y}_i) \in \mathbb{I}, \\
\end{aligned}
\label{eq:max_conse}
\end{equation}
where $\mathbb{I}$ refers to an unknown consensus set, ``$ \lfloor \cdot  \rfloor$'' denotes the cardinality of the set, and $\xi$ is a hyperparameter called ``inlier threshold''. The motivation of problem (\ref{eq:max_conse}) is to find the optimal ($\mathbf{R}$, $\mathbf{t}$) that fits the largest consensus set. Given the ground-truth ($\mathbf{R}^*$, $\mathbf{t}^*$), we denote the correspondences satisfying $\| \mathbf{y}_i -  ( \mathbf{R}^*\mathbf{x}_i+\mathbf{t}^*  )  \| \le \xi$ the \textit{true inliers}, otherwise the \textit{true outliers}.

\subsection{Consensus Maximization Decomposition}
\label{subsec: decom}

As introduced in Section~\ref{sec: related}, some methods are proposed to decouple the rotation and translation in problem~\ref{eq:max_conse} for searching acceleration~\cite{bustos2017guaranteed, liu2018efficient}. While their decomposition reduces the parameter search space, they increase the problem size quadratically and lead to limited efficiency gain. 

In contrast, we propose a three-stage decomposition pipeline that decouples problem~(\ref{eq:max_conse}) into three sub-problems with respect to translation, rotation axis, and rotation angle, respectively. Our decomposition does not increase the problem size. More importantly, our decomposition eases the leverage of our heuristics-guided parameter search, as will be introduced in Section~\ref{sec: Muiti-stage}. We solve the three sub-problems one by one to conduct progressive outlier removal~(see Fig.~\ref{fig:outline}). We now introduce our decomposition strategy in the following.

\noindent \textbf{First Sub-problem.} The constraint in problem~(\ref{eq:max_conse}) can be rewritten by introducing an inlier threshold vector $\mathbf{\epsilon}_i$, i.e., 
\begin{equation}
\begin{aligned}
\mathbf{y}_i = \mathbf{R}\mathbf{x}_i+\mathbf{t} + \mathbf{\epsilon}_i,\   \| \mathbf{\epsilon}_i  \| \le \xi, \ \mathbf{\epsilon}_i \in \mathbb{R}^3.
\end{aligned}
\label{eq:inlier}
\end{equation}

We can reduce Eq.~(\ref{eq:inlier}) as
\begin{subequations}
\begin{align}
& \quad \ \  \mathbf{y}_i = \mathbf{R}\mathbf{x}_i+\mathbf{t} + \mathbf{\epsilon}_i \notag\\
&\Rightarrow  \|\mathbf{y}_i -\mathbf{t}  \| =  \|\mathbf{R}\mathbf{x}_i + \mathbf{\epsilon}_i  \| \label{eq: te_1}\\
\scriptsize{\makecell{\color{gray}\rm Triangle\\ \rm \color{gray}Inequality}}&\Rightarrow  \| \mathbf{R}\mathbf{x}_i \| -  \| \mathbf{\epsilon }_i \| \le   \|\mathbf{y}_i -\mathbf{t}  \| \le  \| \mathbf{R}\mathbf{x}_i \| +  \| \mathbf{\epsilon }_i \| \label{eq: te_2}\\ 
\scriptsize{\makecell{\color{gray} \rm Rotation\\ \color{gray}\rm Elimination}}&\Rightarrow  -  \| \mathbf{\epsilon }_i \| \le  \|\mathbf{y}_i -\mathbf{t}  \| -  \| \mathbf{x}_i \| \le  \| \mathbf{\epsilon }_i \| \label{eq: te_3}\\ 
&\Rightarrow \big | { \| \mathbf{y}_i -\mathbf{t}  \| -  \| \mathbf{x}_i \| } \big | \le   \| \mathbf{\epsilon }_i \| \label{eq: te_4}\\
&\Rightarrow \big | { \| \mathbf{y}_i -\mathbf{t}  \| -  \| \mathbf{x}_i \| } \big | \le \xi \label{eq: te_5}
\end{align}
\end{subequations}
where Eq.~(\ref{eq: te_3}) holds because $ \| \mathbf{R}\mathbf{a}  \| = \| \mathbf{a}\|$ for $\mathbf{a} \in \mathbb{R}^3$, ``$| \cdot |$'' in Eq.~(\ref{eq: te_4}) denotes the absolute value. Based on the above procedures, the constraint in Eq.~(\ref{eq:inlier}) is relaxed to the constraint defined by Eq.~(\ref{eq: te_5}) which is only with respect to the linear 3-DoF translation vector $\textbf{t}$. We employ the relaxation to derive the first sub-problem, i.e.,
\begin{equation}
\begin{aligned}
&\max_{\mathbf{t},\ \mathbb{I}_1 \subseteq \mathcal{P}} \  \lfloor \mathbb{I}_1  \rfloor \\
& s.t. \ \big | { \|\mathbf{y}_i -\mathbf{t}  \| -  \| \mathbf{x}_i \|} \big | \le \xi , \ \forall (\mathbf{x}_i, \mathbf{y}_i) \in \mathbb{I}_1
\\
\end{aligned}
\label{eq: sub_t}
\end{equation}
where $\mathbb{I}_1$ is an unknown consensus set at the first stage. By solving problem~(\ref{eq: sub_t})~(see Section~\ref{subsec: te}), we can obtain a candidate translation vector $\mathbf{t}'$ associated with a candidate consensus set ${\mathbb{I}_1}'$. 
Since the constraint in problem~(\ref{eq: sub_t}) is relaxed from the constraint in problem~(\ref{eq:max_conse}), the consensus set ${\mathbb{I}_1}'$ might still contain some \textit{true outliers}. 
We therefore conduct further outlier removal based on the remaining 3-DoF rotation.

\noindent \textbf{Second Sub-problem.} By fixing the candidate $\mathbf{t}'$, the constraint in problem~(\ref{eq:max_conse}) can be simplified into
\begin{equation}
\begin{aligned}
\|\mathbf{y}_i -\mathbf{t}^{'} - \mathbf{R}\mathbf{x}_i  \| \le \xi.
\end{aligned}
\label{eq: wahha}
\end{equation}
While only 3-DoF are left for parameter search, the high non-linearity of the rotation still limits the searching efficiency~\cite{parra2015guaranteed, straub2017efficient}. Inspired by recent works~\cite{peng2022arcs, chen2022deterministic} regarding rotation search, we choose to decouple the rotation axis $\mathbf{r}$ and rotation angle $\theta$ and search them separately for step-wise outlier removal. Specifically, we can reduce Eq.~(\ref{eq: wahha}) as
\begin{subequations}
\begin{align}
& \quad \  \|\mathbf{y}_i -\mathbf{t}^{'} - \mathbf{R}\mathbf{x}_i  \| \le \xi \notag\\
\scriptsize{\color{gray}  \|\mathbf{r} \| = 1 \quad \ }& \Rightarrow   \|\mathbf{y}_i -\mathbf{t}^{'} - \mathbf{R}\mathbf{x}_i  \| \cdot   \| \mathbf{r} \| \le \xi \label{eq: re_1}\\
\scriptsize{\makecell{\color{gray} \rm Cauchy\textnormal{-}Schwarz\\ \color{gray} \rm Inequality}} & \Rightarrow | ( \mathbf{y}_i -\mathbf{t}^{'})^{\top}\mathbf{r}  -  (\mathbf{R}\mathbf{x}_i )^{\top}\mathbf{r} | \le \xi \label{eq: re_2}\\
\scriptsize{\color{gray}  ((\mathbf{R}\mathbf{x}_i )^{\top}\mathbf{r} = } & \scriptsize{\color{gray}\mathbf{r}^{\top}\mathbf{R}\mathbf{x}_i = ( \mathbf{R}^{\top}\mathbf{r} ) ^{\top}\mathbf{x}_i = \mathbf{r}^{\top}\mathbf{x}_i = \mathbf{x}_i^{\top}\mathbf{r}} \color{gray}) \notag\\
& \Rightarrow | ( \mathbf{y}_i -\mathbf{t}^{'} - \mathbf{x}_i)^{\top}\mathbf{r} | \le \xi \label{eq: re_3}
\end{align}
\end{subequations}
where from Eq.~(\ref{eq: re_2}) to Eq.~(\ref{eq: re_3}) we rely on the fact that $\mathbf{R}^{\top}\mathbf{r} = \mathbf{r}$~(the proof is presented in the appendix). Based on the above procedures, we obtain the constraint in Eq.~(\ref{eq: re_3}) only with respect to the 2-DoF rotation axis $\mathbf{r}$. We can then derive the second sub-problem, i.e.,
\begin{equation}
\begin{aligned}
&\max_{\mathbf{r}, \ \mathbb{I}_2 \subseteq {\mathbb{I}_1}^{'}} \  \lfloor \mathbb{I}_2  \rfloor \\
& s.t. \ |  (\mathbf{y}_i -\mathbf{t}^{'} - \mathbf{x}_i )^{\top} \mathbf{r} | \le \xi , \ \forall (\mathbf{x}_i, \mathbf{y}_i) \in \mathbb{I}_2 \\
&\quad \quad \quad \quad  \|\mathbf{r}\| = 1, \ r_3 \ge 0\\
\end{aligned}
\label{eq: sub_r}
\end{equation}
where $\mathbb{I}_2$ is an unknown consensus set at the second stage and $r_3$ is the $z$-axis component of $\mathbf{r}$~($r_3\ge0$ is used to avoid ambiguity of $\mathbf{r}$). Note that the $\mathbf{t}'$ and ${\mathbb{I}_1}'$ has been solved in the first sub-problem in Eq.~(\ref{eq: sub_t}). Therefore, problem~(\ref{eq: sub_r}) is only with respect to the 2-DoF rotation axis with lower nonlinearity. 
By solving problem~(\ref{eq: sub_r})~(see Section~\ref{subsec: re}), we obtain a candidate rotation axis $\mathbf{r}'$ associated with the consensus set ${\mathbb{I}_2}'$ which contains only a few \textit{true outliers}~(see Fig.~\ref{fig:outline}). 

\noindent \textbf{Third Sub-problem.} Since now only the 1-DoF rotation angle $\theta$ is left, we can derive the third sub-problem as
\begin{equation}
\begin{aligned}
&\max_{\theta,\ \mathbb{I}_3 \subseteq {\mathbb{I}_2}^{'}} \  \lfloor \mathbb{I}_3  \rfloor \\
& s.t. \  \|\mathbf{y}_i -\mathbf{t}' - \mathbf{R}\mathbf{x}_i  \| \le \xi, \ \forall (\mathbf{x}_i, \mathbf{y}_i) \in \mathbb{I}_3 \\
&\quad \ \  \mathbf{R} = \mathbf{r}'\mathbf{r}'^{\top} +  [ \mathbf{r}'  ]_{\times}\sin{\theta } + (\mathbf{I}_3 - \mathbf{r}'\mathbf{r}'^{\top})\cos(\theta) \\ 
\end{aligned}
\label{eq: sub_theta}
\end{equation}
where $\mathbb{I}_3$ is an unknown consensus set at the third stage, the second constraint is the \textit{Rodrigues' rotation formula}, $[ \mathbf{r}' ]_{\times} \in \mathbb{R}^{3\times3}$ denotes the matrix generating the cross product $\times$ by $\mathbf{r}'$. This sub-problem can be easily solved with at most $O(N\log(N))$ time complexity by directly adopting interval stabbing, which will be shown in Section~\ref{subsec: thetae}. And we can then get the corresponding final consensus set ${\mathbb{I}_3}'$. As shown in Fig.~\ref{fig:outline}, based on our multi-stage decomposition and progressive outlier removal, most \textit{true outliers} have be pruned on the final consensus set ${\mathbb{I}_3}'$.

\section{Heuristics-guided Parameter Search for Progressive Outlier Removal}
\label{sec: Muiti-stage}

In this section, we illustrate how we integrate our heuristics-guided parameter search into the solving process of the decomposed sub-problems presented in Section~\ref{sec: formu} in order to perform progressive outlier removal.

\subsection{Stage \uppercase\expandafter{\romannumeral1}: Outlier Removal Based on 3-DoF Translation Constraint}
\label{subsec: te}

Consider the translation constraint of the first sub-problem (\ref{eq: sub_t}), the feasible region of the translation vector $\mathbf{t}$ that makes the correspondence $ (\mathbf{x}_i, \mathbf{y}_i)$ an inlier describes a 3D\protect\footnotemark[2] spherical shell~(see Fig.~\ref{fig: t_est}(a)), i.e., $\mathcal{S}_i = \{ \mathbf{t}\in \mathbb{R}^3: \big | { \| \mathbf{y}_i -\mathbf{t}  \| -  \| \mathbf{x}_i \| } \big | \le \xi\}$. Therefore, from a geometric perspective, to solve problem (\ref{eq: sub_t}) is exactly to find the $\mathbf{t}$ that falls in most spherical shells. To achieve this objective, an intuitive idea based on parameter search is to search the whole parameter space of $\mathbf{t}$~(i.e., $\mathbb{R}^3$), which is generally time-consuming. By contrast, we leverage our heuristics-guided parameter search strategy to reduce the search space. 
\footnotetext[2]{In this paper, the 1D/2D/3D terms in the context of parameter space indicate the number of independent parameters that define the space, i.e., the degree of freedom.}

\definecolor{illu_orange}{rgb}{0.972,0.796,0.557}

\begin{figure}[!t]
    \centering
    \includegraphics[width=0.33\textwidth]{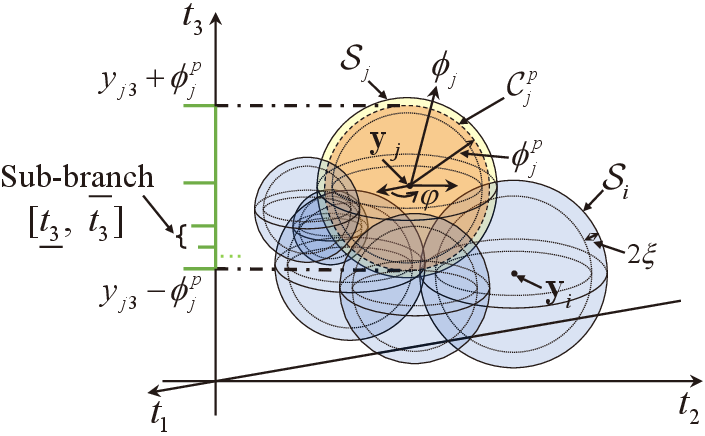}
    \\[-0.3em]
    \makebox[0.18\textwidth]{\footnotesize (a)}
    \\[0.4em]
    \includegraphics[width=0.4\textwidth]{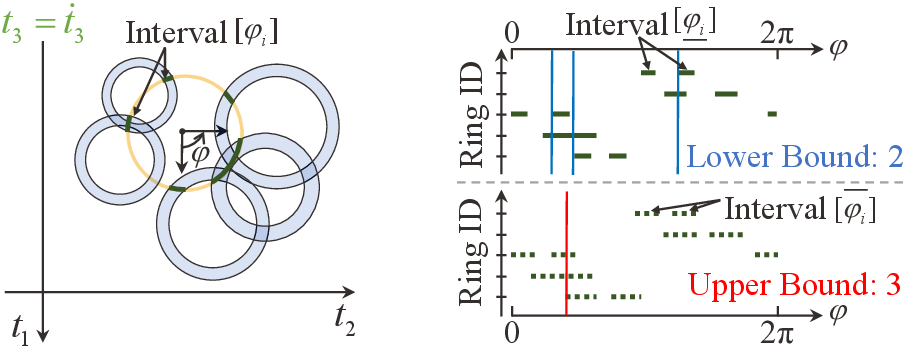}
    \\[-0.5em]
    \makebox[0.2\textwidth]{\footnotesize (b)}
    \makebox[0.24\textwidth]{\footnotesize (c)}
    \\[-0.8em]
    \caption{Illustration of our search strategy for solving the first sub-problem at stage I~(\textit{cf.} Section \ref{subsec: te}). (a) For the feasible region $\mathcal{S}_j$ of $\mathbf{t}$ related to each sampled ($\mathbf{x}_j$, $\mathbf{y}_j$)~(i.e., the \textcolor{GreenYellow}{yellow} shell), we discretize the 3D spherical shell $\mathcal{S}_j$ into $m$ spherical surface $\mathcal{C}_j^p$~(i.e., the \textcolor{illu_orange}{orange} spherical surface)~(\textit{cf.} Section \ref{subsubset: t_dis}). For each $\mathcal{C}_j^p$, we use 1D BnB to search $t_3$. (b) At each branch, when $t_3 = \dot{t}_3 \in [\underline{t_3},\overline{t_3}]$, the intersection of the circle related to ($\mathbf{x}_j$, $\mathbf{y}_j$) and each ring related to ($\mathbf{x}_i$, $\mathbf{y}_i$) leads to the interval [$\varphi_i$]~(\textit{cf.} Section \ref{subsubset: t_1dbnb}). (c) The computations of lower and upper bounds in the outer BnB module only differ in the width of intervals. We adopt interval stabbing as the inner module to compute both bounds~(\textit{cf.} Algorithm~\ref{alg: interval_stabbing}).}
    \label{fig: t_est}
\end{figure}

Particularly, we first sample $k_{\mathbf{t}}$ correspondences from $\mathcal{P}$. Our approach here requires that the $k_{\mathbf{t}}$ samples contain at least a few \textit{true inliers}, and this requirement can be fulfilled, at least empirically, by a valid sampling strategy that we introduce in Section~\ref{sec: spatial}. For each sample $ (\mathbf{x}_j, \mathbf{y}_j)$, $j \in \{j_1, ..., j_{k_{\mathbf{t}}}\} \subseteq \{1, ..., N\}$, we treat it as an \textit{assumptive inlier} and consider the following problem:
\begin{equation}
\begin{aligned}
&\max_{\mathbf{t},\ \mathbb{I}_{1,j}} \  \lfloor \mathbb{I}_{1,j} \rfloor\\
& s.t. \ \big | { \|\mathbf{y}_i -\mathbf{t}  \| -  \| \mathbf{x}_i \|} \big | \le \xi , \ \forall (\mathbf{x}_i, \mathbf{y}_i) \in \mathbb{I}_{1,j} \\
&\quad \quad \quad \quad \quad \quad \quad \ \ \   \mathbf{t} \in \mathcal{S}_j \\
&\quad \quad \quad \quad \quad \mathbb{I}_{1,j} \subseteq \mathcal{P}\setminus \{(\mathbf{x}_j, \mathbf{y}_j)\}
\end{aligned}
\label{eq: quasi_t}
\end{equation}
where $\mathbb{I}_{1,j}$ is an unknown consensus set.
Note that the search space of problem~(\ref{eq: quasi_t}) is the 3D spherical shell $\mathcal{S}_j$.
By solving problem~(\ref{eq: quasi_t}) related to each sample, we can obtain $k_{\mathbf{t}}$ candidate $\mathbf{t}_{j_{1}}', ..., \mathbf{t}_{j_{k_{\mathbf{t}}}}'$ associated with $k_{\mathbf{t}}$ consensus sets $\mathbb{I}_{1,{j_{1}}}'\cup\{(\mathbf{x}_{j_1}, \mathbf{y}_{j_1})\},\ ...,\ \mathbb{I}_{1,{j_{k_{\mathbf{t}}}}}'\cup\{(\mathbf{x}_{j_{k_{\mathbf{t}}}}, \mathbf{y}_{j_{k_{\mathbf{t}}}})\}$, and we choose the one with the maximal cardinality as an approximate solution~(i.e., $\mathbf{t}'$ and ${\mathbb{I}_1}'$) to problem~(\ref{eq: sub_t}). 
Accordingly, we remove those correspondences in $\mathcal{P}\setminus{\mathbb{I}_1}'$ as \textit{candidate outliers}. From problem~(\ref{eq: sub_t}) to problem~(\ref{eq: quasi_t}), our sampling reduces the search space from $\mathbb{R}^3$ to $k_\mathbf{t}$ 3D spherical shells, therefore largely saving the searching time. In addition, as introduced in Section~\ref{sec:introduction}, our approximate solution can fall in the optimal solution region of the sub-problem~(\ref{eq: sub_t}) once a few samples are \textit{true inliers}. While each shell $\mathcal{S}_j$ can be a very small set, directly searching in $\mathcal{S}_j$ is still challenging due to its non-convex shape. Nevertheless, what makes this search problem tractable is our idea of discretizing the 3D spherical shells to 2D spherical surfaces~(see Fig.~\ref{fig: t_est}(a)); and what makes the search efficient is our idea of 1-DoF BnB blended with interval stabbing, which allows us to search a globally optimal solution in the spherical surfaces~(see Figs.~\ref{fig: t_est}(b, c)). We introduce the two ideas in the next two subsections.


\subsubsection{From 3D Spherical Shells to 2D Spherical Surfaces}
\label{subsubset: t_dis}
We consider the cylindrical coordinate representation of the 3D spherical shell $\mathcal{S}_j$ for parameterization simplicity. Let $\phi_j \in \big[\|\mathbf{x}_j\| - \xi, \|\mathbf{x}_j\| + \xi\big]$ define the spherical radial distance, $t_3$ define the height, and $\varphi \in [0, 2\pi]$ define the angle between the $t_1$-axis and the projection vector of $(\mathbf{t-\mathbf{y}_j})$ onto the $t_1t_2$ plane. Then we can represent $\mathcal{S}_j$ as follows:
\begin{equation}
    \begin{aligned}
        \mathcal{S}_j = \{\mathbf{t} \in \mathbb{R}^3: \mathbf{t} = \begin{bmatrix}
\cos{\varphi}\sqrt{(\phi_j)^2 - (t_3-y_{j3})^2} + y_{j1} \\
\sin{\varphi}\sqrt{(\phi_j)^2 - (t_3-y_{j3})^2} + y_{j2}\\
t_3
\end{bmatrix}\}, 
    \end{aligned}
\end{equation}
where $y_{jk}$ defines the $k^{\textnormal{th}}$ component of $\mathbf{y}_j$.
Since the \textit{thickness}~(i.e., $2\xi$) of $\mathcal{S}_j$ is relatively small to the average shell radius~$\|\mathbf{x}_i\|$, we diecretize the shell $\mathcal{S}_j$ into $m$ spherical surfaces $\{\mathcal{C}_j^p\}_{p=1}^m$ along the spherical radius direction. The radius $\phi_j^p$ of the $p^{\textnormal{th}}$ spherical surface in $\mathcal{S}_j$ is computed by
\begin{equation}
    \begin{aligned}
        \phi_j^p = \| \mathbf{x}_j \| +\frac{2p-m-1}{m-1}\xi,\ p\in[1, m].
    \end{aligned}
\end{equation}
Accordingly, each spherical surface $\mathcal{C}_j^p$ can be represented as
\begin{equation}
\begin{aligned}
\mathcal{C}_j^p = \{\mathbf{t} \in \mathbb{R}^3: \mathbf{t} = \begin{bmatrix}
\cos{\varphi}\sqrt{(\phi_j^p)^2 - (t_3-y_{j3})^2} + y_{j1} \\
\sin{\varphi}\sqrt{(\phi_j^p)^2 - (t_3-y_{j3})^2} + y_{j2}\\
t_3
\end{bmatrix}\}.
\end{aligned} 
\label{eq: sphe_sur}
\end{equation}
In this way, we can reduce the 3-dimensional problem (\ref{eq: quasi_t}) into $m$ 2-dimensional sub-problems. Each sub-problem is related to searching the 2D spherical surface $\mathcal{C}_j^p$ defined in Eq. (\ref{eq: sphe_sur}) for some $\mathbf{t}$ defined by ($t_3$, $\varphi$) leading to the largest consensus set. Based on the above notions we can rewrite problem (\ref{eq: quasi_t}) to get the following sub-problem, i.e., 
\begin{equation}
\begin{aligned}
&\max_{t_3,\ \varphi, \ \mathbb{I}_{1,j}^p} \  \lfloor \mathbb{I}_{1,j}^p \rfloor\\
& s.t. \ \big | { \|\mathbf{y}_i -\mathbf{t}  \| -  \| \mathbf{x}_i \|} \big | \le \xi , \ \forall (\mathbf{x}_i, \mathbf{y}_i) \in   \mathbb{I}_{1,j}^p \\
& \quad \quad \quad \quad \quad \quad \quad \  \mathbf{t} \in \mathcal{C}_j^p \\
&\quad \quad \quad \quad \  \mathbb{I}_{1,j}^p \subseteq \mathcal{P} \setminus \{(\mathbf{x}_j, \mathbf{y}_j)\}
\end{aligned}
\label{eq: solve_t}
\end{equation}
where $\mathbb{I}_{1,j}^p$ is an unknown consensus set. By solving problem~(\ref{eq: solve_t}) for each $\mathcal{C}_j^p$, we can obtain $m$ candidate consensus sets and we choose the one with maximal cardinality as an approximate solution to problem~(\ref{eq: quasi_t}). By leveraging the ``thin'' property of the 3D spherical shell $\mathcal{S}_j$, our discretization can avoid affecting the high robustness of our method even with very small $m$, as will be shown in Section~\ref{subsubsec: para_setup}.

\begin{algorithm}[!t]
\caption{Searching A Globally Optimal Solution for Problem~\ref{eq: solve_t}.}
\label{alg: problem_t}
\textbf{Input:} Initial correspondence set $\mathcal{P}$, inlier threshold $\xi$, index of the sampled correspondence $j$, radius of the spherical surface $\phi_{j,p}$.

\textbf{Output:} Maximum consensus set ${\mathbb{I}_{1,j}^{p}}^*$ with cardinality ${Q}^{*}$, and corresponding ${t_3}^{*}$, $\varphi^{*}$.

    \begin{algorithmic}[1]
        \STATE Initialize the queue $q \gets \emptyset$, the searching domain $\mathbb{T} \gets [y_{j3} - \phi_{j,p}, \ y_{j3}+\phi_{j,p}]$ for $t_3$, and ${Q}^* \gets 0$.
        \STATE Compute $\overline{Q}(\mathbb{T})$ by interval stabbing.
        \STATE Insert $\mathbb{T}$ with $\overline{Q}(\mathbb{T})$ to $q$.
        \WHILE{$q$ is not empty}
            \STATE Reach out the branch $\mathbb{B}$ with the highest upper bound $\overline{Q}(\mathbb{B})$ from $q$.
            \IF{$\overline{Q}(\mathbb{B})$ = ${Q}^*$}
            \STATE Terminate.
            \ENDIF
            \STATE Solve $\underline{Q}(\mathbb{B})$, ${\mathbb{I}_{1,j}^{p}}'$, and $\varphi'$ by interval stabbing.
            \IF{$\underline{Q}(\mathbb{B}) > {Q}^*$}
            \STATE Update ${Q}^* \gets \underline{Q}(\mathbb{B})$, ${\mathbb{I}_{1,j}^{p}}^* \gets {\mathbb{I}_{1,j}^{p}}'$, ${\varphi}^* \gets \varphi'$, and ${t_3}^* \gets$ center of $\mathbb{B}$.
            \ENDIF
            \STATE Divide $\mathbb{B}$ into two sub-branches.
            \FOR{each sub-branch $\mathbb{B}_{sub}$}
		\STATE Compute $\overline{Q}(\mathbb{B}_{sub})$ by interval stabbing.
            \IF{$\overline{Q}(\mathbb{B}_{sub}) < {Q}^*$}
            \STATE Discard the $\mathbb{B}_{sub}$ and continue the loop;
            \ELSE
            \STATE Insert $\mathbb{B}_{sub}$ with upper bound $\overline{Q}(\mathbb{B}_{sub})$ into $q$.
            \ENDIF
 		\ENDFOR 
        \ENDWHILE
    \end{algorithmic}
\end{algorithm}
 
\subsubsection{1D BnB blended with Interval Stabbing to Search $\mathbb{S}^2$}
\label{subsubset: t_1dbnb}
For the 2-dimensional problem~(\ref{eq: solve_t}), we develop an efficient parameter search method to find the globally optimal solution, as described in Algorithm~\ref{alg: problem_t}. Specifically, inspired by the nested-BnB~\cite{yang2015go}, we embed the interval stabbing strategy into a 1D BnB framework to search the spherical surface, i.e., $\mathbb{S}^2$. We use the outer BnB module to search $t_3$ and use the inner interval stabbing module to search $\varphi$. Compared with directly searching the 2D space based on BnB, the interval stabbing strategy that involves only a sorting and scanning operation~(see Algorithm~\ref{alg: interval_stabbing})~\cite{preparata2012computational} can largely enhance the searching efficiency.

\noindent \textbf{Outer 1D BnB Module.}
As shown in Fig.~\ref{fig: t_est}(a), we iteratively divide the search space of $t_3$~(i.e., $[y_{j3} - \phi_{j,p}, \ y_{j3}+\phi_{j,p}]$) to obtain wide-to-narrow sub-branches in the outer module. We follow~\cite{chen2022deterministic} to set the minimal branch width as $\psi = 1\textnormal{e}^{-3}$ for precision control. For each sub-branch, we compute lower and upper bounds of the cost function in problem~(\ref{eq: solve_t}). Let $\dot{t}_3$ represent the center of the sub-branch $[\underline{t_3}, \ \overline{t_3}]$, we choose it to compute a lower bound $\underline{Q}$ by
\begin{equation}
    \begin{aligned}
        \underline{Q}=\max_{\varphi} \sum \mathbbm{1}( \big | { \|\mathbf{y}_i -\mathbf{t}  \| -  \| \mathbf{x}_i \|} \big | \le \xi, \\ \mathbf{t}\in\mathcal{C}_j^p,\ t_3 = \dot{t}_3),
    \end{aligned}
    \label{eq: lower_bound_t}
\end{equation}
where $\mathbbm{1}(\cdot)$ is an indicator function that returns 1 if the inside condition is true and 0 otherwise. To compute upper bounds, consider arbitrary $t_3$ in the sub-branch, we have
\begin{subequations}
\begin{align} 
&\big | \| \mathbf{y}_i - \mathbf{t} \| - \|\mathbf{x}_i\| \big | \notag \\
\le & \max\{ \max_{t_3}(\| \mathbf{y}_i- \mathbf{t}\|) - \|\mathbf{x}_i\|, \notag\\
&\quad \quad \ \ \|\mathbf{x}_i\| - \min_{t_3}(\|\mathbf{y}_i - \mathbf{t}\|)\} \label{eq: upper_1} \\
\color{gray}(&\color{gray} \|\mathbf{y}_k - \mathbf{t} \| = \|\mathbf{y}_k - \mathbf{t}_{c} + \mathbf{t}_{c} - t\| \notag\\ 
\scriptsize{\makecell{\color{gray}\rm Triangle\\ \rm \color{gray}Inequality}} \color{gray} \Rightarrow & \color{gray} \|\mathbf{y}_k - \mathbf{t}_{c}\| - \|\mathbf{t}_c-\mathbf{t} \| \le \|\mathbf{y}_k - \mathbf{t} \| \le \notag\\ 
& \color{gray} \|\mathbf{y}_k - \mathbf{t}_{c}\| - \|\mathbf{t}_c+\mathbf{t} \|) \notag \\
\le & \max\{ \|\mathbf{y}_k - \mathbf{t}_c \| - \|\mathbf{x}_k\| + \max_{t_3}(\|\mathbf{t}_c-\mathbf{t} \|), \notag\\
& \quad \quad \ \ \|\mathbf{x}_k\| - \|\mathbf{y}_k - \mathbf{t}_c \| + \max_{t_3}(\|\mathbf{t}_c-\mathbf{t} \|) \} \label{eq: upper_2}\\
\le & \ \xi + \max_{t_3}(\| \mathbf{t}_c - \mathbf{t} \|) \label{eq: upper_3}\\
= &\  \xi + \max\{\| \mathbf{t}_c - \overline{\mathbf{t}} \|, \ \| \mathbf{t}_c - \underline{\mathbf{t}} \| \} \label{eq: upper_4} \\
= & \ \xi + \delta_{t_3}  \label{eq: upper_5}
\end{align}
\end{subequations}
where $\mathbf{t}_c$ represents the center translation vector corresponding to $t_3 = \dot{t}_3$, $\overline{\mathbf{t}}$ and $\underline{\mathbf{t}}$ represent the two-end translation vectors of the branch, respectively; In Eq.~(\ref{eq: upper_5}), we denote $\delta_{t_3}$ as the maximum distance between $\mathbf{t}_c$ and arbitrary point in the sub-branch~(i.e., the greater of $\| \mathbf{t}_c - \overline{\mathbf{t}} \|$ and $\| \mathbf{t}_c - \underline{\mathbf{t}} \|$ in Eq.~(\ref{eq: upper_4})). According to the derivation from (\ref{eq: upper_1}) to (\ref{eq: upper_5}), the residual~(for any $t_3$ in the sub-branch) will be at most $\xi + \delta_{t_3}$. Therefore, we can compute an upper bound $\overline{Q}$ for the given sub-branch by
\begin{equation}
    \begin{aligned}
        \overline{Q}=\max_{\varphi} \sum \mathbbm{1}(\big | { \|\mathbf{y}_i -\mathbf{t}  \| -  \| \mathbf{x}_i\|} \big | \le \xi + \delta_{t_3} , \\ \mathbf{t}\in\mathcal{C}_j^p,\ t_3 = \dot{t}_3).
    \end{aligned}
    \label{eq: upper_bound_t}
\end{equation}

\noindent \textbf{Inner Interval Stabbing Module.}
In the inner module, given a specific $\dot{t}_3$, we aim to solve problem~(\ref{eq: lower_bound_t}) and problem~(\ref{eq: upper_bound_t}) that only differ in the threshold. Fig.~\ref{fig: t_est}(b) shows the cross section of Fig.~\ref{fig: t_est}(a) at $t_3 = \dot{t}_3$. The yellow circle represents the feasible region of $\mathbf{t}$ that makes the sampled correspondence an inlier at $t_3 = \dot{t}_3$. And each ring relates to the feasible region that makes the corresponding correspondence an inlier. From a geometric perspective, solving problem~(\ref{eq: lower_bound_t}) and problem~(\ref{eq: upper_bound_t}) is equivalent to searching through the circle to find the $\varphi$ related to the maximum number of stabbed rings~(i.e., inliers). We adopt the interval stabbing strategy to solve this problem. 

In particular, we compute the intersection interval~$[\varphi_i]$ of each ring and the circle. Then we sort all the interval endpoints and sequentially scan these endpoints to find the endpoint $\tilde{\varphi}$ that stabs most intervals~(see Algorithm~\ref{alg: interval_stabbing}). Note that different thresholds in problem~(\ref{eq: lower_bound_t}) and problem~(\ref{eq: upper_bound_t}) lead to rings with different thicknesses, and therefore to intervals with different widths~(see Fig.~(\ref{fig: t_est})(c)). We denote the intervals to estimate lower bound in problem~(\ref{eq: lower_bound_t}) by $[\underline{\dot{t}_3}]$ and the intervals to estimate upper bound in problem~(\ref{eq: upper_bound_t}) by $[\overline{\dot{t}_3}]$. We choose the cardinality of the most stabbed intervals as the solution. The interval stabbing strategy can get the optimal solution to problem~(\ref{eq: lower_bound_t}) and problem~(\ref{eq: upper_bound_t}), therefore guaranteeing the accuracy of our inner module. 

Based on the upper and lower bounds computed in the inner module, we can prune those sub-branches that have upper bounds smaller than the so-far largest lower bound during each iteration in our outer BnB module. Accordingly, the outer BnB module is guaranteed to converge to a globally optimal solution (${t_3}'$,\  $\varphi'$) of problem~(\ref{eq: solve_t}).

\begin{algorithm}[!t]
\caption{Interval Stabbing.}
\label{alg: interval_stabbing}
\textbf{Input:} Intervals $\mathcal{I}$ = $\{[\gamma^l_i,\ \gamma^r_i]\}_{i=1}^{M}$.

\textbf{Output:} Best stabber $\gamma^{*}$ and the related number of stabbed intervals $T^{*}$.

    \begin{algorithmic}[1]
        \STATE $\hat{\mathcal{I}}$ $\gets$ Sort all the endpoints in $\mathcal{I}$. 
        \STATE \textit{count} $\gets$ 0, $T^{*}$ $\gets$ \textit{count}.
        \FOR{i $\gets$ 1 $\mathbf{to}$ $2M$}
            \IF{$\hat{\mathcal{I}}(i)$ is a left endpoint}
            \STATE \textit{count} $\gets$ \textit{count} + 1.
                \IF{\textit{count} $> T^{*}$}
                    \STATE $\gamma^{*} \gets \hat{\mathcal{I}}(i)$, $T^{*}$ $\gets$ \textit{count}.
                \ENDIF
            \ELSE
            \STATE \textit{count} $\gets$ \textit{count} - 1.
            \ENDIF
 		\ENDFOR
    \end{algorithmic}
\end{algorithm}

\subsection{Stage \uppercase\expandafter{\romannumeral2}: Outlier Removal Based on 2-DoF Rotation Axis Constraint}
\label{subsec: re}

\begin{figure}[!t]
    \centering
    \includegraphics[width=0.37\textwidth]{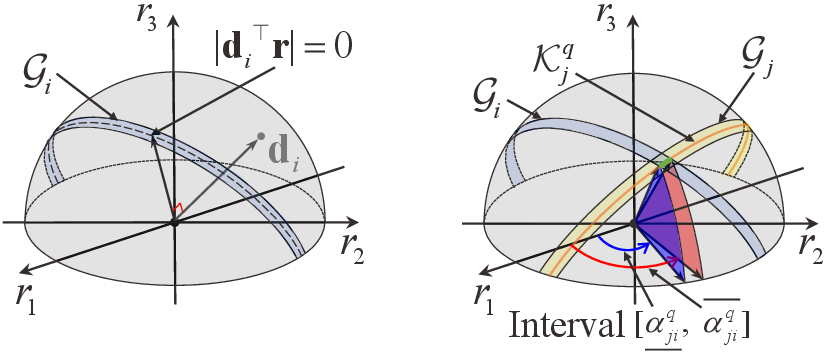}
    \\
    \makebox[0.2\textwidth]{\footnotesize \quad \quad (a)}
    \makebox[0.24\textwidth]{\footnotesize (b)\ }
    \\[-0.8em]
    \caption{Illustration of our search strategy for solving the second sub-problem at stage II~(\textit{cf.} Section \ref{subsec: re}). (a) The constraint defined by Eq.~(\ref{eq: r_cons_simp}) results in a girdle-like feasible region $\mathcal{G}_i$. (b) We discretize $\mathcal{G}_j$ related to the sampled ($\mathbf{x}_j$, $\mathbf{y}_j$) into $n$ half-circles $\mathcal{K}_j^q$~(\textit{cf.} Section \ref{subsubset: r_dis}). Note that the intersection of $\mathcal{K}_j^q$ and each $\mathcal{G}_i$ leads to a interval~(\textit{cf.} Section \ref{subsubset: r_search}). Therefore we apply interval stabbing to find the candidate ${\mathbf{r}_{j}^q}'$ corresponding to each 1D $\mathcal{K}_j^q$.}
    \label{fig: r_est}
\end{figure}

By fixing the translation vector $\mathbf{t}'$ estimated at stage I, we conduct further outlier removal for the related consensus set ${\mathbb{I}_1}'$ by solving the sub-problem~(\ref{eq: sub_r}). Consider the rotation axis constraint (i.e., Eq.~(\ref{eq: re_3})), we define $\mathbf{d}_i \doteq \frac{\mathbf{y}_i -\mathbf{t}^{'} - \mathbf{x}_i}{\|\mathbf{y}_i -\mathbf{t}^{'} - \mathbf{x}_i \|}$ and $\xi_i \doteq \frac{\xi}{\|\mathbf{y}_i -\mathbf{t}^{'} - \mathbf{x}_i \|}$ to rewrite it by 
\begin{equation}
\begin{aligned}
|{\mathbf{d}_i}^{\top} \mathbf{r}| \le \xi_i.
\end{aligned}
\label{eq: r_cons_simp}
\end{equation}
Accordingly, problem~(\ref{eq: sub_r}) can be rewritten by
\begin{equation}
\begin{aligned}
&\max_{\mathbf{r}, \ \mathbb{I}_2 \subseteq {\mathbb{I}_1}^{'}} \  \lfloor \mathbb{I}_2  \rfloor \\
& s.t. \ |  {\mathbf{d}_i}^{\top} \mathbf{r} | \le \xi_i , \ \forall (\mathbf{x}_i, \mathbf{y}_i) \in \mathbb{I}_2 \\
&\quad \quad \quad \ \ \ \|\mathbf{r}\| = 1, \ r_3 \ge 0\\
\end{aligned}
\label{eq: sub_r_trans}
\end{equation}
As shown in Fig.~\ref{fig: r_est}(a), the constraints defined by Eq.~(\ref{eq: r_cons_simp}) results in a girdle-like feasible region $\mathcal{G}_j = \{\mathbf{r} \in \mathbb{S}^2: |{\mathbf{d}_j}^{\top} \mathbf{r}| \le \xi_j,\ \|\mathbf{r}\| = 1, \ r_3 \ge 0 \}$ in the unit hemisphere (i.e., the whole parameter space of $\mathbf{r}$). And the girdle is bisected by a plane with the normal $\mathbf{d}_i$. Therefore, problem~(\ref{eq: sub_r_trans}) reads: find a unit vector~(i.e., the optimal $\mathbf{r}$) in the unit hemisphere that stabs the most number of girdles defined by Eq.~(\ref{eq: r_cons_simp}). To solve problem~(\ref{eq: sub_r_trans}), an intuitive strategy is to search the whole 2D parameter space of $\mathbf{r}$. By contrast, we again adopt our heuristics-guided parameter search strategy to reduce the search space. 

Specifically, we first sample $k_{\mathbf{r}}$ correspondences from ${\mathbb{I}_1}'$. Similarly to stage I, we adopt our valid sampling strategy introduced in Section~\ref{subsec: sampling} to guarantee enough number of sampled correspondences are the inliers. We treat each sampled correspondence $ (\mathbf{x}_j,\  \mathbf{y}_j)$, $j \in \{j_1, ..., j_{k_{\mathbf{r}}}\}$ as the \textit{assumptive inlier} and consider the following problem:
\begin{equation}
\begin{aligned}
&\max_{\mathbf{r}, \ \mathbb{I}_{2,j}} \  \lfloor \mathbb{I}_{2,j}  \rfloor \\
& s.t. \ |  {\mathbf{d}_i}^{\top} \mathbf{r} | \le \xi_i , \ \forall (\mathbf{x}_i, \mathbf{y}_i) \in \mathbb{I}_{2,j} \\
&\quad \quad \quad \quad \quad \ \ \mathbf{r} \in \mathcal{G}_j\\
&\quad \quad \quad \  \mathbb{I}_{2,j} \subseteq {\mathbb{I}_1}^{'}\setminus \{ (\mathbf{x}_j, \mathbf{y}_j)\}
\end{aligned}
\label{eq: quasi_r}
\end{equation}
where $\mathbb{I}_{2,j}$ is an unknown consensus set. 
Note that the search space of problem~(\ref{eq: quasi_r}) is the 2D girdle $\mathcal{G}_j$.
By solving problem~(\ref{eq: quasi_r}) for each sample, we obtain $k_{\mathbf{r}}$ candidate $\mathbf{r}_{j_{1}}', ..., \mathbf{r}_{j_{k_{\mathbf{r}}}}'$ associated with $k_{\mathbf{r}}$ consensus sets $\mathbb{I}_{2,{j_{1}}}'\cup\{(\mathbf{x}_{j_1}, \mathbf{y}_{j_2})\},\ ...,\ \mathbb{I}_{2,{j_{k_{\mathbf{r}}}}}'\cup\{(\mathbf{x}_{j_{k_{\mathbf{r}}}}, \mathbf{y}_{j_{k_{\mathbf{r}}}})\}$, and we choose the one with the maximal cardinality as an approximate solution~(i.e., $\mathbf{r}$ and ${\mathbb{I}_2}'$) to problem~(\ref{eq: sub_r_trans})~(i.e., the problem~(\ref{eq: sub_r})). Accordingly, we remove those correspondences in ${\mathbb{I}_1}'\setminus{\mathbb{I}_2}'$ as \textit{candidate outliers}. Based on our sampling, the parameter search space at stage II is largely reduced, i,e., from $\mathbb{S}^2$ to $k_{\mathbf{r}}$ 2-dimensional girdles. For efficient parameter search on the 2-dimensional $\mathcal{G}_j$ to solve problem~(\ref{eq: quasi_r}), we propose to reduce the search by discretizing the 2D girdles to 1D half-circles and accordingly adopt the 1D interval stabbing to estimate the solution. We introduce our discretization and 1D search in the following two subsections, respectively.

\subsubsection{From 2D Girdles to 1D Half-circles}
\label{subsubset: r_dis}
Consider the \textit{width} of the 2D girdle $\mathcal{G}_j$ is generally small~(approximately $2\xi_j$), we diecretize the girdle~$\mathcal{G}_j$ into $n$ half-circles $\mathcal{K}_j^1,\ ...,\ \mathcal{K}_j^n$, as shown in Fig.~\ref{fig: r_est}(b). The $q^{\textnormal{th}}$ half-circle of $\mathcal{G}_j$ can be represented as
\begin{equation}
\begin{aligned}
\mathcal{K}_j^q = \{\mathbf{r} \in \mathbb{S}^2:\
&{\mathbf{d}_j}^{\top}\mathbf{r} = \xi_j^q,\ \|\mathbf{r}\|=1, \ r_3 \ge0\\
&\xi_j^q = \frac{2q-n-1}{n-1}\xi_j,\ q\in[1, n] \},
\end{aligned} 
\label{eq: half_circle}
\end{equation}
where $\xi_j^q$ is a threshold constraining the position of $\mathcal{K}_j^q$. 
Consider the geographic coordinate representation of $\mathbf{r}$, i.e., $\mathbf{r} = [\sin{\beta}\cos{\alpha},\ \sin{\beta}\sin{\alpha},\ \cos{\beta}]^{\top}$, where $\alpha$ denotes the angle of longitude and $\beta$ denotes the angle of latitude. Since $\mathbf{r}$ in the half-circle $\mathcal{K}_j^q$ can be simply constrained by one parameter~($\alpha$ or $\beta$), we consider the following problem:
\begin{equation}
\begin{aligned}
&\max_{\alpha} \  \lfloor \mathbb{I}_{2,j}^q  \rfloor \\
& s.t. \ |  {\mathbf{d}_i}^{\top} \mathbf{r} | \le \xi_i , \ \forall i \in \mathbb{I}_{2,j}^q \\
&\quad \quad \quad \quad \ \ \mathbf{r} \in \mathcal{K}_j^q\\
&\quad \quad \  \mathbb{I}_{2,j}^q \subseteq {\mathbb{I}_1}^{'}\setminus \{ \mathbf{x}_j, \mathbf{y}_j\}
\end{aligned}
\label{eq: solve_r}
\end{equation}
where $\mathbb{I}_{2,j}^q$ is an unknown consensus set. Note that the search space of problem~(\ref{eq: solve_r}) is the 1D half-circle $\mathcal{K}_j^q$. For the 2-dimensional problem~(\ref{eq: quasi_r}) of each sampled correspondence, we can resemble it into $n$ 1-dimensional sub-problems (i.e., problem~(\ref{eq: solve_r})) by discretizing the corresponding feasible region~$\mathcal{G}_j$ into $n$ half-circles $\mathcal{K}_j^q$s. By solving problem~(\ref{eq: solve_r}) for each $\mathcal{K}_j^q$, we obtain $n$ candidate consensus set and choose the one with maximal cardinality as an approximate solution to problem~(\ref{eq: quasi_r}). 
By leveraging the ``narrow'' property of the 2D girdle $\mathcal{G}_j$, our discretization can avoid affecting the high robustness of our method even with very small $n$, as will be shown in Section~\ref{subsubsec: para_setup}.

\subsubsection{Interval Stabbing for 1D Search}
\label{subsubset: r_search}
For the 1-dimensional problem~(\ref{eq: solve_r}), we adopt the interval stabbing strategy introduced in Section~\ref{subsec: te} to solve it. For each $\mathcal{G}_i$ of the unsampled correspondences in ${\mathbb{I}_1}^{'}\setminus \{ \mathbf{x}_j, \mathbf{y}_j\}$, we compute the intersection between the girdle and the half-circle $\mathcal{K}_j^q$~(see Fig.~\ref{fig: r_est}(b)) by
\begin{equation}
\begin{aligned}
\begin{Bmatrix}
|  {\mathbf{d}_i}^{\top} \mathbf{r} | \le \xi_i\\
{\mathbf{d}_j}^{\top}\mathbf{r} = \xi_j^q  \\
\mathbf{r} = [\sin{\beta}\cos{\alpha},\ \sin{\beta}\sin{\alpha},\ \cos{\beta}]^{\top}
\end{Bmatrix}\Leftrightarrow \alpha \in [\underline{\alpha_{ji}^{q}}, \ \overline{\alpha_{ji}^{q}}],
\end{aligned}
\end{equation}
where $\big[{\underline{\alpha_{ji}^{q}}}, {\overline{\alpha_{ji}^{q}}}\big]$ represents the intersection interval defined by $\alpha$. Based on these intervals, we can employ interval stabbing to find the optimal region of $\alpha$ in $\mathcal{K}_j^q$ that intersects with the most number of girdles. Therefore, we can get an optimal solution ${\mathbf{r}_{j}^q}'$ of problem~(\ref{eq: solve_r}) by directly adopting Algorithm~\ref{alg: interval_stabbing}. The efficiency of interval stabbing leads to fast parameter search for the $\mathbf{r}'$ and related consensus set ${\mathbb{I}_2}^{'}$.

\begin{figure}[!t]
    \centering
    \includegraphics[width=0.22\textwidth]{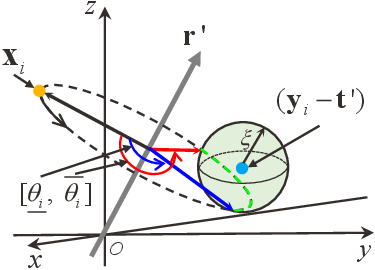}
    \\[-0.9em]
    \caption{At stage III, each correspondence is related to an interval [$\underline{\theta_i}$, $\overline{\theta_i}$]~(\textit{cf.} Section~\ref{subsec: thetae}), and we directly apply interval stabbing to get a globally optimal solution of the third sub-problem~(i.e., problem~(\ref{eq: sub_theta})).}
    \label{fig: theta_est}
\end{figure}

\subsection{Stage \uppercase\expandafter{\romannumeral3}: Outlier Removal Based on 1-DoF Rotation Angle Constraint}
\label{subsec: thetae}

After the above two-stage outlier removal, most \textit{true outlier}s can be removed~(see Fig.~(\ref{fig:outline})). We now consider the sub-problem~(\ref{eq: sub_theta}) to solve for the remaining 1-DoF rotation angle and conduct the final-stage outlier removal. As shown in Fig.~\ref{fig: theta_est}, the constraint in the sub-problem~(\ref{eq: sub_theta}) for each correspondence leads to a feasible interval~$[\underline{\theta_i}, \overline{\theta_i}]$ of $\theta$. Such interval results from the intersection between the \textit{trajectory} of $\mathbf{x}_i$ that rotates around the axis $\mathbf{r}'$ and the \textit{spherical region} with center $(\mathbf{y}_i - \mathbf{t}')$ and radius $\xi$. Since each correspondence is associated with an interval (no intersection leads to $\emptyset$), solving the sub-problem~(\ref{eq: sub_theta}) is equivalent to finding the $\theta'$ that stabs a maximal number of intervals. Therefore we adopt the interval stabbing strategy~(i.e., Algorithm~\ref{alg: interval_stabbing}) introduced before to solve this problem. Based on interval stabbing, we can get a globally optimal solution~(i.e, $\theta'$ and ${\mathbb{I}_3}'$) of the sub-problem~(\ref{eq: sub_theta}). We remove the correspondences in ${\mathbb{I}_3}' \setminus {\mathbb{I}_3}'$ as \textit{candidate outliers}. Given $N$ putative correspondences, the interval stabbing leads to at most $N\log(N)$ time complexity and guarantees the efficiency of this stage.

\section{Boost Performance Based on Spatial Compatibility}
\label{sec: spatial}

While we can simply follow the pipeline introduced above to get the solution, our formulation provides further opportunities to boost the performance of our approach by leveraging the spatial compatibility constraint. We begin with introducing the spatial compatibility.

The spatial compatibility describes the fact that the distance between two keypoints in a point cloud is invariant to any rigid transformations. Specifically, consider the inlier constraint in Eq.~(\ref{eq:inlier}) for two different correspondences $(\mathbf{x}_u, \mathbf{y}_u)$ and $(\mathbf{x}_v, \mathbf{y}_v)$, we have
\begin{subequations}
\begin{align}
&\left\{ {\begin{array}{*{20}{c}}
{\mathbf{y}_u = \mathbf{R}\mathbf{x}_u+\mathbf{t} + \mathbf{\epsilon}_u,\   \| \mathbf{\epsilon}_u  \| \le \xi } \\
{\mathbf{y}_v = \mathbf{R}\mathbf{x}_v+\mathbf{t} + \mathbf{\epsilon}_v,\   \| \mathbf{\epsilon}_v  \| \le \xi }
\end{array}} \right. \label{eq: sc_1}\\
&\Rightarrow  \mathbf{y}_u - \mathbf{y}_v = \mathbf{R}(\mathbf{x}_u - \mathbf{x}_v) + \mathbf{\epsilon}_u - \mathbf{\epsilon}_v \label{eq: sc_2}\\
\scriptsize{\makecell{\color{gray} \rm Triangle\\ \color{gray} \rm Inequality}} &\Rightarrow | \|\mathbf{y}_u - \mathbf{y}_v \| - \|\mathbf{x}_u - \mathbf{x}_v\| | \le \| \mathbf{\epsilon}_u - \mathbf{\epsilon}_v \| \label{eq: sc_3}\\
&\Rightarrow | \|\mathbf{y}_u - \mathbf{y}_v \| - \|\mathbf{x}_u - \mathbf{x}_v\| | \le 2\xi \label{eq: sc_4}
\end{align}
\end{subequations}
Here the two correspondences $(\mathbf{x}_u,\ \mathbf{y}_u)$ and $(\mathbf{x}_v,\ \mathbf{y}_v)$ are called spatially compatible if they satisfy the constraint defined in Eq.~(\ref{eq: sc_4}). 
That two correspondences being spatially compatible does not necessarily mean both of them are \textit{true inliers}, but violating spatial compatibility immediately implies that at least either of them is a \textit{true outlier}~\cite{parra19pairwise}.
We leverage the compatibility constraint to boost the performance of our method in terms of both efficiency and robustness from the following two aspects.

\subsection{Valid Sampling via Spatial Compatibility}
\label{subsec: sampling}
In Sections~\ref{subsec: te} and \ref{subsec: re}, we adopt our heuristics-guided parameter search strategy and sample a specific number of correspondences for search space reduction. Note that the sampling holds a significant impact on the robustness and efficiency of our method. 
In particular, the sampling needs to be ``valid'' in the sense that a few sampled correspondences are~\textit{true inlier}s, for otherwise our method may search in the totally wrong region. 
However, to guarantee ``valid'' sampling, the classic idea of random sampling may require large sampling numbers which jeopardize the efficiency~\cite{fischler1981random}.
To address this issue, we improve our sampling by leveraging the spatial compatibility of 3D correspondences.


\begin{figure}[!t]
    \centering
    \includegraphics[width=0.4\textwidth]{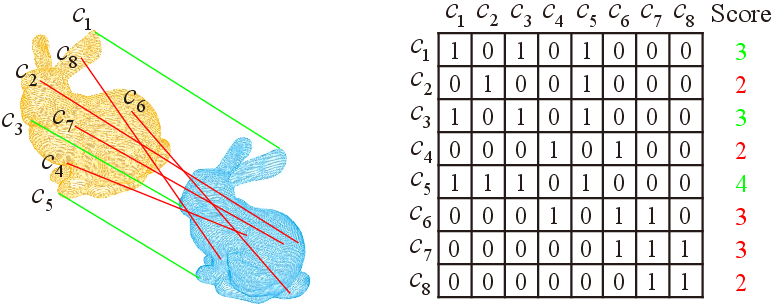}
    \\
    \makebox[0.2\textwidth]{\footnotesize (a)}
    \makebox[0.2\textwidth]{\footnotesize (b)}
    \\[0.6em]
    \includegraphics[width=0.4\textwidth]{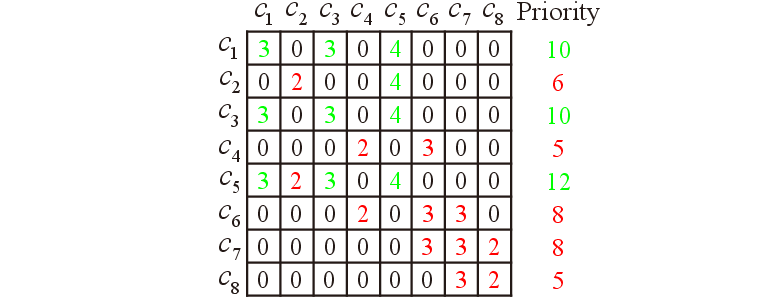}
    \\
    \makebox[0.2\textwidth]{\footnotesize (c)}
    \vspace{-0.8em}
    \caption{(a) A set of putative correspondences in which outliers and inliers are represented by \textcolor{red}{red} and \textcolor{green}{green} line segments, respectively. (b) Using the number of compatible correspondences~(i.e., the scores) to distinguish inliers and outliers can easily fail. (c) We propose to take the scores of the compatible correspondences into consideration and compute the priorities. The priorities show higher robustness to distinguish the inliers and outliers.}
    \label{fig: valid_sampling}
\end{figure}

Given the initial correspondences, we first construct an affinity matrix $W$ based on checking the compatibility between every two correspondences~(see Fig.~\ref{fig: valid_sampling}(b)). For each correspondence $c_i$, we compute the number of correspondences that are compatible with it~(including itself) and take the number as the score, i.e., 
\begin{equation}
\begin{aligned}
\textnormal{Score}(c_i) = \sum_{j=1}^{N} W(c_i, c_j).
\end{aligned} 
\label{eq: score}
\end{equation}
Since the \textit{true inliers} are mutually compatible while the \textit{true outliers} are random, correspondences with higher scores may hold a higher possibility to be \textit{true inlier}s~\cite{9373914}. Therefore, an intuitive sampling strategy is to choose correspondences with higher scores~\cite{quan2020compatibility}. However, when the inlier ratio is relatively low, such a sampling strategy can easily fail. For example, Fig.~\ref{fig: valid_sampling}(b) shows that some \textit{true outliers} have the same scores as the \textit{true inliers}. 

In contrast, we propose to take the scores of the compatible correspondences of each correspondence into consideration. The intuition is that the \textit{true inliers} should not only have higher scores, they should also be compatible with the correspondences with higher scores~(i.e., other \textit{true inliers}). Accordingly, for each correspondence, we sum up the scores of its compatible correspondences and take it as the priority:
\begin{equation}
\begin{aligned}
\textnormal{Priority}(c_i) = \sum_{j=1}^{N} W(c_i, c_j)\cdot\textnormal{Score}(c_j). 
\end{aligned} 
\label{eq: priority}
\end{equation}
As shown in Fig.~\ref{fig: valid_sampling}(c), our operation leads to easily distinguishable priorities between \textit{true inliers}~(higher priorities) and \textit{true outliers}~(lower priorities). We then rank all the correspondences by the related priorities and sample correspondences with the highest priorities. In particular, we choose the top-$m$ correspondences from $\mathcal{P}$ at stage I and the top-$n$ correspondences from ${\mathbb{I}_1}'$ at stage II, respectively. Based on spatial compatibility, the improved sampling strategy results in a large inlier ratio in the samples, which boosts both efficiency and accuracy of our method. First, the sampling number that guarantees ``valid'' sampling is largely reduced, which leads to smaller search regions and less time cost. 
In addition, compared to~\cite{chen2022sc2}, our priority computation does not contain the multiplication of large matrices and thus is more efficient. 
Second, more inliers in the sampled correspondences make our parameter search focus on the regions containing optimal solutions and reduce the disturbance of outliers. Related analysis experiments are presented in Section~\ref{subsubsec: para_setup} and the appendix.

\subsection{Compatibility Verification for Interval Computation}
\label{subsec: com_interval}
To solve problem~(\ref{eq: solve_t}) and problem~(\ref{eq: solve_r}) in Section~\ref{sec: Muiti-stage}, a basic procedure of our method is to compute the intersection intervals between the sampled correspondence and each left correspondences in the given set. Based on these intervals, we can then conduct the outlier removal based on interval stabbing. In the following, we will show that we can leverage spatial compatibility to improve this procedure in terms of both efficiency and robustness.

Particularly, instead of directly computing the intersection intervals between the sample and \textit{each} left correspondence, we can first conduct a compatibility verification based on the constraint in Eq.~(\ref{eq: sc_4}). We only compute the intervals of the correspondences that are compatible with the sample. For the correspondence that fails in the verification, we just ignore it since it and the sample cannot be both inliers. We therefore only need to conduct interval stabbing on the purified interval set. Such verification can improve not only the efficiency but also the robustness of our interval stabbing, as will be shown in Section~\ref{subsubsec: verfi}. First, for each sampled correspondence, we only need to consider the compatible correspondences instead of all correspondences. And the compatibility verification can be easily conducted by checking the affinity matrix $W$ constructed in Section~\ref{subsec: sampling}, which largely saves the computation cost. Second, our verification adds an additional constraint~(i.e., spatial compatibility) to the original problem~(\ref{eq: solve_t}) and problem~(\ref{eq: solve_r}), which additionally rejects outliers and thus leads to higher robustness.

\section{Time Complexity Analysis}
\label{sec: complexity}
The proposed method can achieve quadratic time complexity with respect to the correspondence number $N$. Specifically, the time complexity of each module are as follows:
(1) stage I leads to $O(k_{\mathbf{t}}m\cdot\frac{1}{\psi}N\log N)$ time complexity, where $k_{\mathbf{t}}$ and $m$ are the number of sampled correspondences and the number of spherical surfaces~(feasible region) related to each sample. In Section~\ref{subsubsec: para_setup} we show that $k_{\mathbf{t}}=15$ and $m=2$ are enough for high estimation robustness;
(2) stage II leads to $O(k_{\mathbf{r}}n \cdot N\log N)$ time complexity, where $k_{\mathbf{r}}$ and $n$ are the number of sampled correspondences and the number of half-circles~(feasible region) related to each sample. In Section~\ref{subsubsec: para_setup} we show that $k_{\mathbf{r}}=8$ and $n=2$ are enough for high estimation robustness;
(3) stage III has $O(N\log N)$ time complexity with a single interval stabbing operation;
(4) the priority computation for valid sampling has $O(N^2)$ time complexity; 
(5) the post-refinement based on SVD has $O(N)$ time complexity.
Therefore, the worst-case time complexity of our method is $O(N^2 + \tau N\log N)$, where $\tau := k_{\mathbf{t}}m\cdot\frac{1}{\psi} + k_{\mathbf{r}}n + 1$ defines a constant coefficient.

\begin{figure*}[!t]
    \centering
    \includegraphics[width=0.98\textwidth]{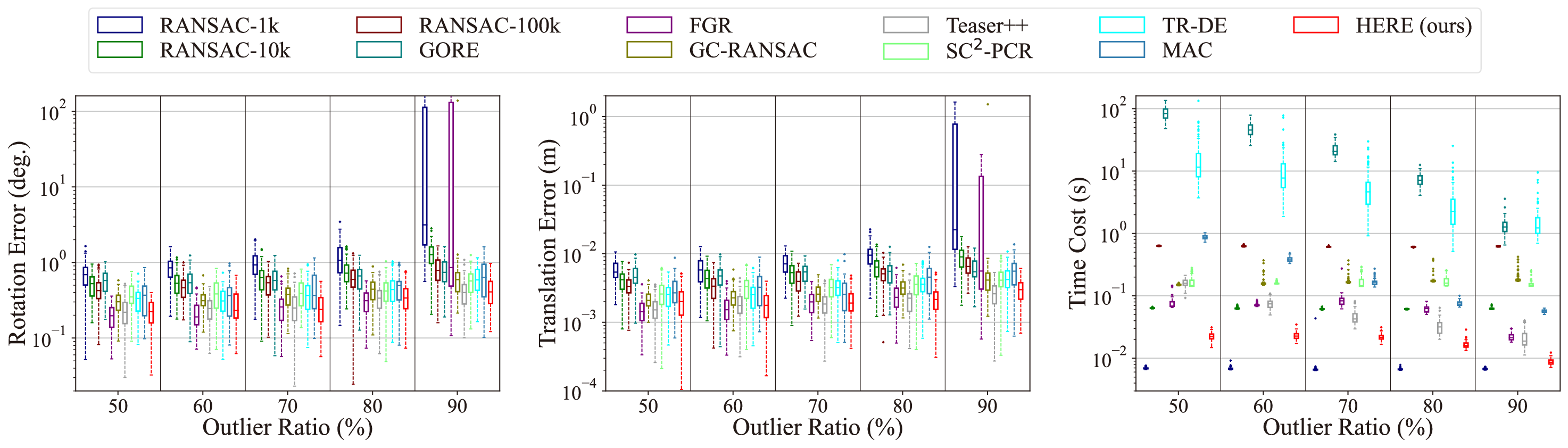}
    \\[-0.5em]
    \makebox[0.327\textwidth]{\footnotesize \quad \quad \ \ (a-1)}
    \makebox[0.327\textwidth]{\footnotesize \quad \quad \ \ (a-2)}
    \makebox[0.327\textwidth]{\footnotesize \quad \quad \ \ (a-3)}
    \\[0.2em]
    \includegraphics[width=0.98\textwidth]{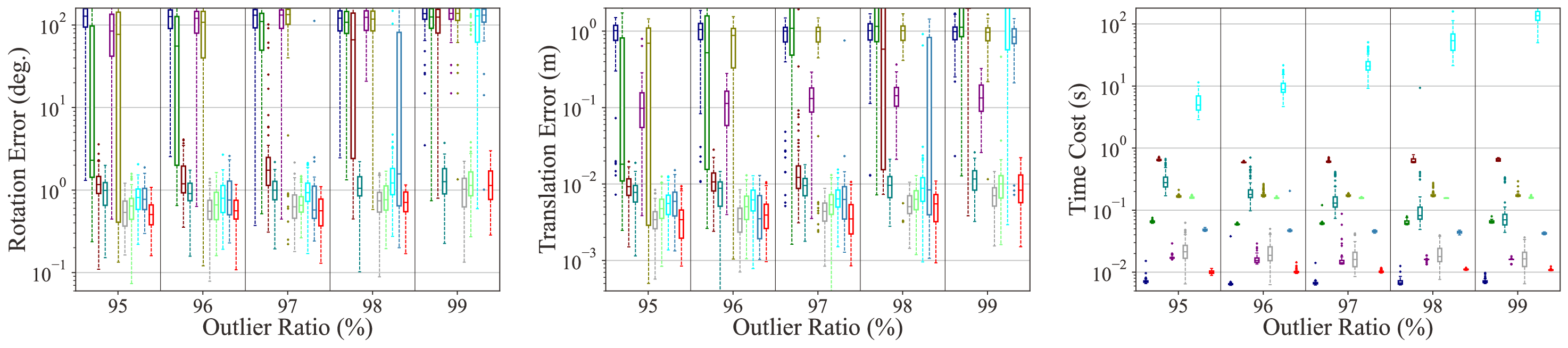}
    \\[-0.4em]
    \makebox[0.327\textwidth]{\footnotesize \quad \quad \ \ (b-1)}
    \makebox[0.327\textwidth]{\footnotesize \quad \quad \ \ (b-2)}
    \makebox[0.327\textwidth]{\footnotesize \quad \quad \ \ (b-3)}
    \\
    \includegraphics[width=0.98\textwidth]{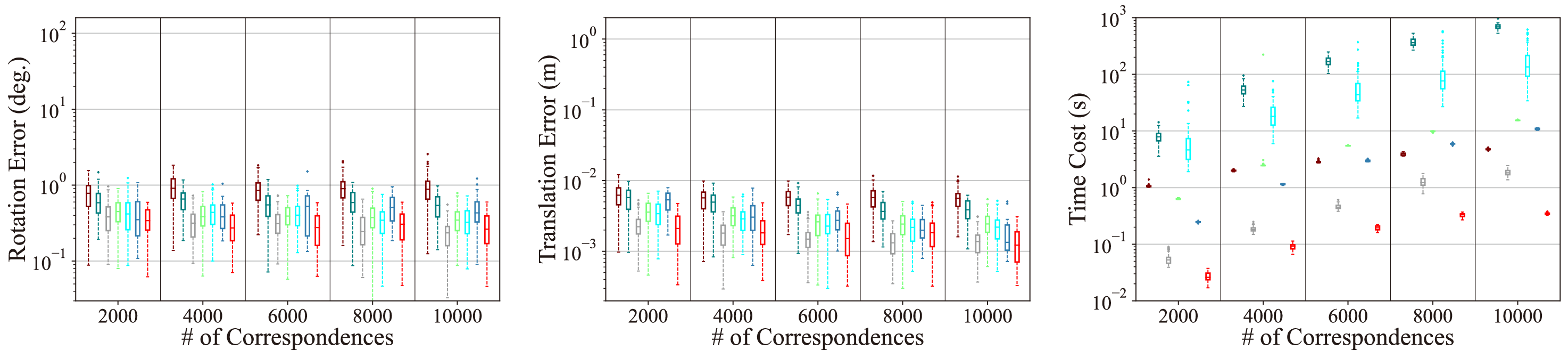}
    \\[-0.5em]
    \makebox[0.327\textwidth]{\footnotesize \quad \quad \ \ (c-1)}
    \makebox[0.327\textwidth]{\footnotesize \quad \quad \ \ (c-2)}
    \makebox[0.327\textwidth]{\footnotesize \quad \quad \ \ (c-3)}
    \vspace{-0.8em}
    \caption{Evaluation results on simulated datasets. (a) Boxplots of rotation errors, translation errors, and time costs of all compared methods for basic evaluation in different outlier ratios. (b) Same as before, but for robustness evaluation regarding extreme outlier ratios. (c) Boxplots of rotation errors, translation errors, and time costs of the six highly-robust methods for efficiency evaluation regarding large-scale correspondences.}
    \label{fig: simu_eval}
\end{figure*}

\section{Experiments}
\label{sec: exper}
We compare our method with several baselines and state-of-the-arts that aim at rigid point cloud registration on both simulated~(see Section~\ref{subsec: exper_simu}) and real-world~(see Section~\ref{subsec: exper_real}) datasets. In addition, we conduct analysis experiments for our method~(see Section~\ref{subsec: ana_exper}).

The compared methods include: (1) two parameter search-based methods: \textsf{GORE}~~\cite{bustos2017guaranteed} and \textsf{TR-DE}~\cite{chen2022deterministic}; (2) \textsf{RANSAC}\protect\footnotemark[3]~\cite{fischler1981random} and its variant \textsf{GC-RANSAC}~\cite{barath2021graph}; (3) the fast method \textsf{FGR}~\cite{zhou2016fast}; (4) \textsf{TEASER++}~\cite{yang2020teaser} and \textsf{MAC}~\cite{Zhang_2023_CVPR} that both leverage the compatibility graph; (5) the \textsf{SC$^2$-PCR}~\cite{chen2022sc2} that relies on the second-order spatial compatibility. We denote the proposed HEuristics-guided paRameter sEarch method by \textsf{HERE} for simplicity.
\footnotetext[3]{For a comprehensive comparison, we follow~\cite{chen2022deterministic} to run RANSAC for 1k~(resp. 10k, 100k) iterations, and denote the corresponding methods by \textsf{RANSAC}-xk~(x = 1, 10, 100).}

\noindent \textbf{Implementation Details.}
We implement the proposed method based on C++ and use shared-memory parallelism to accelerate it following \textsf{TEASER++}. For the other methods, we use the provided implementations. Since most of the compared methods~(except \textsf{SC$^2$-PCR}) are implemented for only CPU execution, all methods are tested on CPUs for a fair comparison. For \textsf{SC$^2$-PCR}, we slightly adjust their PyTorch-based codes so that we can run it on CPUs; note that this only affects the running time, not accuracy, and the readers can find its precise running time on GPUs in~\cite{chen2022sc2}. Most experiments are evaluated on a laptop equipped with an Intel Core i7-10875H CPU@2.3 GHz and 16GB of RAM. In addition, since \textsf{MAC} needs to list all the maximal cliques\protect\footnotemark[4] and runs out of memory in most cases, we test it on the same laptop with 32GB of RAM.
\footnotetext[4]{In the worst case, a graph with $N$ vertexes can have as many as $3^{N/3}$ maximal cliques \cite{Moon-IJM1965,Tomita-TCS2006}.}

\subsection{Comparison on Simulated Datasets}
\label{subsec: exper_simu}

\noindent \textbf{Data Generation.} Similarly to~\cite{bustos2017guaranteed, yang2020teaser}, we first downsample the Bunny model from the Stanford 3D Scanning Repository~\cite{curless1996volumetric} to $N$ points. We resize it to fit inside the unit cube and obtain the \textit{source cloud} $\mathcal{X}$. We then generate the \textit{target cloud} $\mathcal{Y}$ by applying a random rotation~($\mathbf{R}\in SO(3)$) and translation~($\|\mathbf{t}\|\le 1$) to $\mathcal{X}$. For each point in $\mathcal{Y}$, we add uniformly distributed noise inside a sphere centered at the point with a radius of 0.02. Each pair of the original point in $\mathcal{X}$ and the related transformed point in $\mathcal{Y}$ defines a \textit{true inlier} correspondence. To generate \textit{true outliers}, we substitute a fraction~(i.e., the outlier ratio $\rho$) of transformed points in $\mathcal{Y}$ with points uniformly sampled inside the sphere of radius 5. We vary the number of correspondences $N$ and the outlier ratio $\rho$ for different evaluation experiments.

\subsubsection{Basic Evaluation in Different Outlier Ratios}
We first set $N=1000$ and $\rho$ = $\{$50$\%$, 60$\%$, 70$\%$, 80$\%$, 90$\%\}$ to evaluate all the methods. We conduct 100 Monte Carlo runs for each setup and each method to avoid accidental errors. Following~\cite{chen2022deterministic}, we compute the rotation error~($\textit{E}_\mathbf{R}$) and translation error~($\textit{E}_\mathbf{t}$) by $\textit{E}_\mathbf{R} = \arccos{\frac{\textnormal{Tr}(\mathbf{\hat{R}}^{\top}\mathbf{R}^{*})-1}{2}}$ and $\textit{E}_\mathbf{t} = \|\mathbf{\hat{t}}-\mathbf{t}^{*}\|$, where $\mathbf{\hat{R}}$ and $\mathbf{\hat{t}}$ denote the estimated rotation and translation, respectively; $\mathbf{R}^{*}$ and $\mathbf{t}^{*}$ are the ground-truth rotation and translation, respectively; $\textnormal{Tr}(\cdot)$ represents the trace of a matrix. Figs.~\ref{fig: simu_eval}(a-1)-(a-3) show the rotation error, translation error, and timing results of all the methods under different outlier ratios. \textsf{RANSAC}-1k shows the highest efficiency in each outlier ratio, but it leads to the worst accuracy compared with other methods and fails in the outlier ratio of 90\% due to sampling uncertainty. Thanks to the robust function, \textsf{FGR} can produce more accurate estimation results, but it also fails to deal with the 90$\%$ outliers. By contrast, \textsf{TEASER++} and the proposed \textsf{HERE} enjoy both higher registration accuracy and lower time cost. And \textsf{HERE} runs faster than all the methods except for \textsf{RANSAC}-1k. Significantly, compared with the parameter search-based methods \textsf{GORE} and \textsf{TR-DE}, the efficiency of \textsf{HERE} increases more than 2 orders of magnitude.

\subsubsection{Robustness regarding Extreme Outlier Ratios}
\label{subsubsec: simu_extre}
To further evaluate the robustness, we compare all the methods by setting extreme outlier ratios. We set $N=1000$ and $\rho=\left\{95\%,\ 96\%,\ 97\%,\ 98\%,\ 99\%\right\}$. And we conduct 100 Monte Carlo runs for each setup and each method. Figs.~\ref{fig: simu_eval}(b-1)-(b-3) show the rotation error, translation error, and timing results of all the methods under different extreme outlier ratios. Based on random sampling, RANSAC can fail in the outlier ratio of 98$\%$ even with 100k iterations. And the variant \textsf{GC-RANSAC} also starts to fail in the outlier ratio of 95$\%$. Without surprise, \textsf{FGR} fails in all the five extreme outlier ratios. Even the parameter search-based \textsf{TR-DE} shows unreliable registration results at the outlier ratio of 98$\%$ and fails at 99$\%$, and so does the recent method \textsf{MAC}. Note that only \textsf{GORE}, \textsf{TEASER++}, \textsf{SC$^2$-PCR}, and \textsf{HERE} can survive the 99$\%$ outliers and show high robustness. In addition, \textsf{HERE} still shows higher efficiency than other robust methods. Note that \textsf{HERE} is more than 10 times faster than \textsf{GORE} and \textsf{SC$^2$-PCR}. 
This is because \textsf{GORE} relies on total parameter search and \textsf{SC$^2$-PCR} on the construction of large-scale SC$^2$ matrices, while our method is carefully designed to avoid both issues.

\subsubsection{Efficiency Regarding Large-scale Correspondences}
To further evaluate the efficiency, we conduct experiments by setting large-scale correspondences. We set $\rho=95\%$ and $N=\left\{2000,\ 4000,\ 6000,\ 8000,\ 10000 \right\}$. And we compare the seven methods with enough robustness~(i.e., surviving the outlier ratio of 95$\%$ in the robustness experiment). We conduct 100 Monte Carlo runs for each setup and each method.
As shown in Fig.~\ref{fig: simu_eval}(c-3), the running times of \textsf{GORE} and \textsf{TR-DE} can go extremely high as the number of correspondences goes larger. And the \textsf{SC$^2$-PCR} leads to time costs larger than \textsf{RANSAC}-100k due to its construction of the SC$^{2}$ matrices. By contrast, the proposed \textsf{HERE} shows the highest efficiency among the six robust methods. Specifically, \textsf{HERE} runs more than 3 orders of magnitude faster than \textsf{GORE}, 2 orders of magnitude faster than \textsf{TR-DE}, 15 times faster than \textsf{SC$^2$-PCR}, 10 times faster than \textsf{MAC}, and 3 times faster than \textsf{TEASER++}. In addition, as shown in Figs.~\ref{fig: simu_eval}(c-1) and (c-2), our method and the state-of-the-art \textsf{TEASER++} both show higher accuracy than other methods. Above all, apart from keeping comparable robustness as the state-of-the-arts, \textsf{HERE} enjoys higher efficiency.

\begin{table*}[!t]
\centering
    \caption{Quantitative comparison results regarding registration accuracy and efficiency on the outdoor ETH dataset~\cite{theiler2014keypoint}. \textbf{Bold} fonts denote the first place, and \underline{underlined} fonts denote the second place. For those methods with \textit{RR} lower than 50$\%$, we don't consider them when comparing efficiency~(marked as \textcolor{gray}{gray}).}
    \vspace{-0.8em}
    \label{tab: ETH_acc}
    \footnotesize
    \renewcommand{\tabcolsep}{1.9pt} 
    \renewcommand\arraystretch{1.5}
    \begin{tabular}{c|ccc|ccc|ccc|ccc|ccc}
        \Xhline{1pt}
        Method & \multicolumn{3}{c|}{Arch} & \multicolumn{3}{c|}{Courtyard} & \multicolumn{3}{c|}{Facade} & \multicolumn{3}{c|}{Office} & \multicolumn{3}{c}{Trees}\\
        \cline{2-16}
         & \textit{RR} $\uparrow$ & \makecell{$\textit{E}_\mathbf{R}$/$\textit{E}_\mathbf{t}$\\($^{\circ}$/cm) $\downarrow$} & \makecell{Time\\(s) $\downarrow$} 
         & \textit{RR} $\uparrow$ & \makecell{$\textit{E}_\mathbf{R}$/$\textit{E}_\mathbf{t}$\\($^{\circ}$/cm) $\downarrow$} & \makecell{Time\\(s) $\downarrow$} & \textit{RR} $\uparrow$ & \makecell{$\textit{E}_\mathbf{R}$/$\textit{E}_\mathbf{t}$\\($^{\circ}$/cm) $\downarrow$} & \makecell{Time\\(s) $\downarrow$} & \textit{RR} $\uparrow$ & \makecell{$\textit{E}_\mathbf{R}$/$\textit{E}_\mathbf{t}$\\($^{\circ}$/cm) $\downarrow$} & \makecell{Time\\(s) $\downarrow$} & \textit{RR} $\uparrow$ & \makecell{$\textit{E}_\mathbf{R}$/$\textit{E}_\mathbf{t}$\\($^{\circ}$/cm) $\downarrow$} & \makecell{Time\\(s) $\downarrow$} \\
        \Xhline{0.5pt}
        \textsf{RANSAC}-1k~\cite{fischler1981random} & 0/10 & - & \color{gray}0.05 & 4/28 & 1.26/26.98 & \color{gray}0.07 & 7/21 & 1.20/0.27 & \color{gray}0.02 & 0/10 & - & \color{gray}0.02 & 0/15 & - & \color{gray}0.10\\
        \textsf{RANSAC}-10k~\cite{fischler1981random} & 0/10 & - & \color{gray}0.46 & 18/28 & 0.23/9.77 & \textbf{0.52} & 17/21 & 0.98/16.82 & \underline{0.08} & 3/10 & 2.47/28.30 & \color{gray}0.06 & 0/15 & - & \color{gray}0.82\\
        \textsf{RANSAC}-100k~\cite{fischler1981random} & 2/10 & 1.02/25.86 & \color{gray}4.48 & \underline{27/28}& 0.30/9.48& 4.99& \underline{20/21} & 0.30/7.52 & 0.81 &8/10 & 1.46/13.47 & 0.62 & 1/15 & 2.80/46.36 & \color{gray}7.71\\
        \textsf{GORE}~\cite{bustos2017guaranteed} & \textbf{9/10} & 0.47/9.16 & 156.27 & \textbf{28/28} & 0.16/7.35 & 153.97 & \textbf{21/21} & 0.32/8.03 & 5.84 & \underline{9/10} & 0.88/7.96 & 13.6& \textbf{13/15} & 0.68/9.61 & 2576.47\\
        \textsf{FGR}~\cite{zhou2016fast} & 0/10 & - & \color{gray}0.17 & 2/28 & 1.15/27.17 & \color{gray}0.19 & 9/21 & 0.58/16.01 & \color{gray}0.05 & 1/10 & 1.91/17.32 & \color{gray}0.03 & 0/15 & - & \color{gray}0.26\\
        \textsf{GC-RANSAC}~\cite{barath2021graph} & 0/10 & - & \color{gray}0.45 & 12/28 & 0.11/3.96 & \color{gray}0.81 & 17/21 & 0.24/4.43 & 0.16 & 5/10 & 0.74/\textbf{5.17} & 0.13 & 0/15 & - & \color{gray}0.98\\        
        \textsf{TEASER++}~\cite{yang2020teaser} & \textbf{9/10} & 0.29/6.99 & \underline{1.25} & \textbf{28/28} & 0.09/3.98 & 1.49 & \textbf{21/21} & \textbf{0.18}/4.17 & 0.09 & \textbf{10/10} & 0.98/7.34 & \textbf{0.06} & \textbf{13/15} & 0.31/7.28 & \underline{3.79}\\
        \textsf{SC$^2$-PCR}~\cite{chen2022sc2} & \underline{8/10} & \textbf{0.24}/\textbf{6.47} & 14.19 & \textbf{28/28} & \textbf{0.03}/\underline{3.65} & 18.73 & \textbf{21/21} & \underline{0.20}/4.02 & 0.32 & \underline{9/10} & \textbf{0.62}/\underline{7.24} & 0.15 & 11/15 & \underline{0.16}/\underline{4.07} & 39.51\\
        \textsf{TR-DE}~\cite{chen2022deterministic} & \underline{8/10} & 0.37/9.15 & 541.11 & 26/28 & 0.19/9.98 & 216.3 & \textbf{21/21} & 0.36/7.72 & 5.18 & \underline{9/10} & 1.26/10.37 & 1.46 & 10/15 & 1.02/10.12 & 786.23\\
        \textsf{MAC}~\cite{Zhang_2023_CVPR} & \underline{8/10} & 0.42/8.65 & 16.73 & \textbf{28/28} & \underline{0.05}/\textbf{3.64} & 291.67 & \textbf{21/21} & 0.23/\underline{3.99} & 0.25 & 8/10 & 0.88/7.38 & 0.10 & \underline{12/15} & \textbf{0.14}/\textbf{3.94} & 57.51\\
        \textsf{HERE}~(ours) & \textbf{9/10} & \underline{0.28}/\underline{6.82} & \textbf{0.54} & \textbf{28/28} & \underline{0.05}/3.79 & \underline{0.65} & \textbf{21/21} & 0.23/\textbf{3.97} & \textbf{0.07} & \textbf{10/10} & \underline{0.71}/7.57 & \underline{0.07} & \textbf{13/15} & 0.30/6.85 & \textbf{0.86}\\
        \Xhline{1pt}
    \end{tabular}
\end{table*}

\begin{table*}[!t]
\centering
    \caption{Quantitative comparison results regarding registration accuracy, outlier removal performance, and efficiency on the outdoor KITTI dataset~\cite{geiger2012we}.}
    \vspace{-0.8em}
    \label{tab: Kitti_nomutual}
    \footnotesize
    \renewcommand{\tabcolsep}{2.6pt} 
    \renewcommand\arraystretch{1.5}
    \begin{tabular}{c|ccccccc|ccccccc}
        \Xhline{1pt}
        Method & \multicolumn{7}{c|}{FPFH Descriptor~\cite{rusu2009fast}} & \multicolumn{7}{c}{FCGF Descriptor~\cite{choy2019fully}} \\
        \cline{2-15}
         & \textit{RR}(\%)$\uparrow$ & $\textit{E}_\mathbf{R}$($^{\circ}$)$\downarrow$ & $\textit{E}_\mathbf{t}$(cm)$\downarrow$ & \textit{IP}(\%)$\uparrow$ &  \textit{IR}(\%)$\uparrow$ & $F_1$(\%)$\uparrow$ & Time(s)$\downarrow$ & \textit{RR}(\%)$\uparrow$ & $\textit{E}_\mathbf{R}$($^{\circ}$)$\downarrow$ & $\textit{E}_\mathbf{t}$(cm)$\downarrow$ & \textit{IP}(\%)$\uparrow$ &  \textit{IR}(\%)$\uparrow$ & $F_1$(\%)$\uparrow$ & Time(s)$\downarrow$\\
        \Xhline{0.5pt}
        \textsf{RANSAC}-1k~\cite{fischler1981random} & 19.09 & 2.31 & 36.64 & 43.30 & 15.13 & 20.38 & \color{gray}0.04 & 97.29 & 0.53 & 23.37 & 80.73 & 86.52 & 83.19 & \textbf{0.05}\\
        \textsf{RANSAC}-10k~\cite{fischler1981random} & 66.66 & 1.67 & 31.25 & 74.15 & 47.56 & 55.97 & 0.39 & 97.47 & 0.41 & 22.66 & 80.58 & 88.18 & 83.89 & 0.51\\
        \textsf{RANSAC}-100k~\cite{fischler1981random} & 95.67 & 1.06 & 23.19 & 86.99 & 77.02 & 81.23 & 3.79 & 98.01 & 0.39 & 21.73 & 80.82 & 89.46 & 84.65 & 4.95\\
        \textsf{FGR}~\cite{zhou2016fast} & 9.73 & 0.58 & 27.84 & 14.43 & 8.97 & 10.20 & \color{gray}0.14 & 97.47 & 0.34 & 19.86 & \underline{81.68} & 89.20 & 84.87 & \underline{0.20}\\
        \textsf{GC-RANSAC}~\cite{barath2021graph} & 79.46 & \textbf{0.39} & \underline{8.02} & 76.62 & 76.49 & 76.02 & 1.24 & 97.47 & \underline{0.32} & \underline{20.50} & 81.41 & 90.09 & 85.21 & 0.84\\
        \textsf{TEASER++}~\cite{yang2020teaser} & 97.84 & 0.43 & 8.39 & 92.75 & 95.01 & 93.73 & \underline{0.36} & \underline{98.02} & 0.34 & 20.74 & 81.05 & 90.36 & 85.36 & 37.81\\
        \textsf{SC$^2$-PCR}~\cite{chen2022sc2} & \textbf{99.64} & \textbf{0.39} & 8.29 & 93.01 & \textbf{95.85} & \textbf{94.26} & 4.33 & 97.66 & \textbf{0.31} & \textbf{20.21} & 81.65 & \textbf{90.44} & \underline{85.48} & 4.06\\
        \textsf{TR-DE}~\cite{chen2022deterministic} & 98.91 & 0.90 & 15.63 & 88.07 & 87.01 & 87.20 & 132.66 & 97.11 & 0.83 & 24.33 & 80.24 & 85.48 & 82.30 & 237.17\\
        \textsf{MAC}~\cite{Zhang_2023_CVPR} & \underline{99.10} & 0.51 & 10.17 & \underline{93.12} & 88.86 & 89.55 & 5.74 & 97.66 & 0.45 & 23.40 & 78.95 & 87.68 & 82.14 & 2.47\\
        \textsf{HERE}~(ours) & \underline{99.10} & \underline{0.42} & \textbf{7.90} & \textbf{93.15} & \underline{95.33} & \underline{94.23} & \textbf{0.28} & \textbf{98.20} & \underline{0.32} & 20.73 & \textbf{81.86} & \underline{90.34} & \textbf{85.50} & 0.25\\
        \Xhline{1pt}
    \end{tabular}
\end{table*}

\begin{figure*}
    \centering
    \footnotesize
    \renewcommand{\tabcolsep}{12pt}
    \begin{tabular}{cccc}
      \makecell{\includegraphics[width=0.15\textwidth]{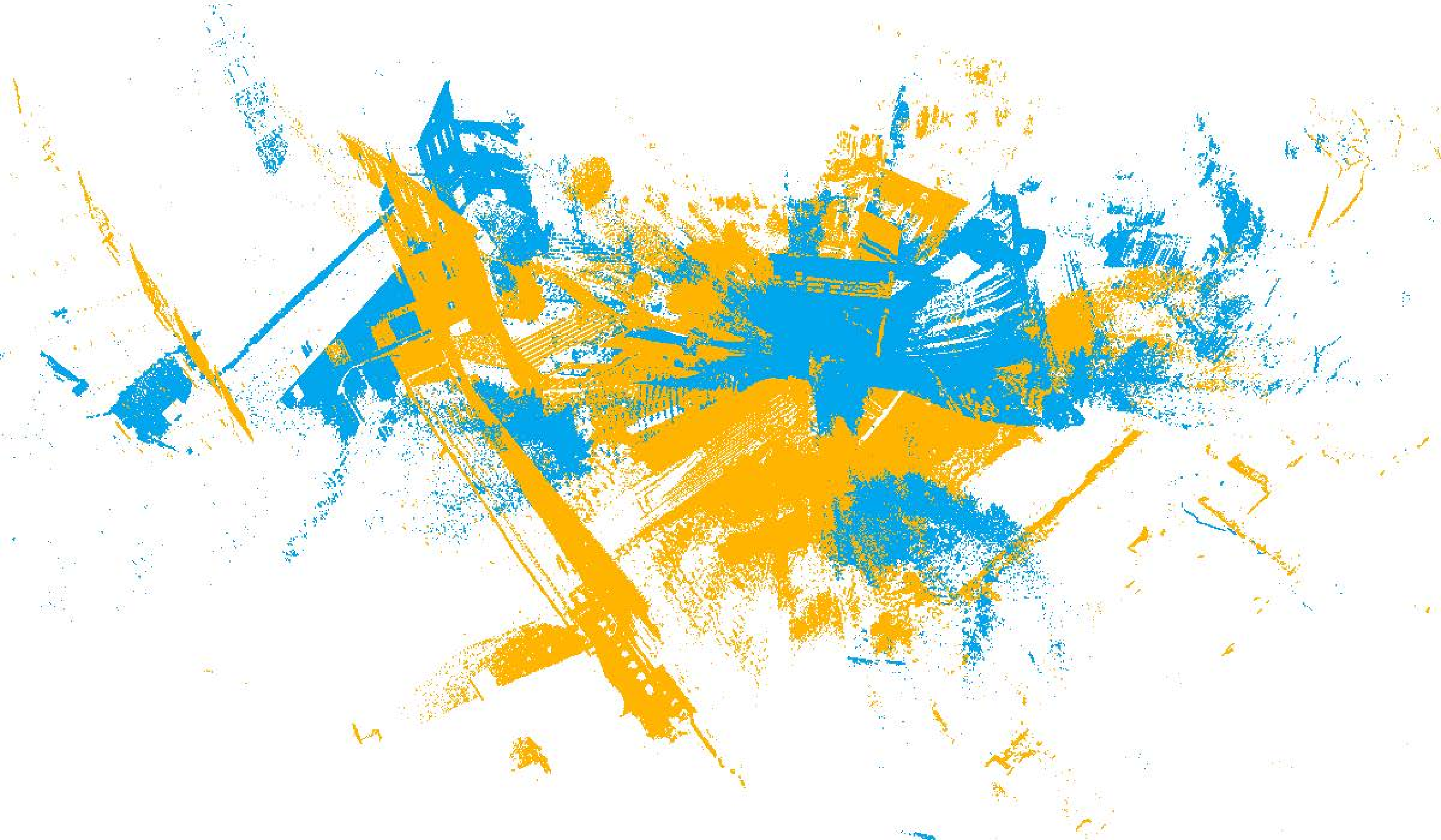} \\ (a-1) Inputs - ETH Dataset \\ N = 10394\\ $\rho$ = 98.91$\%$} &
      \makecell{\includegraphics[width=0.15\textwidth]{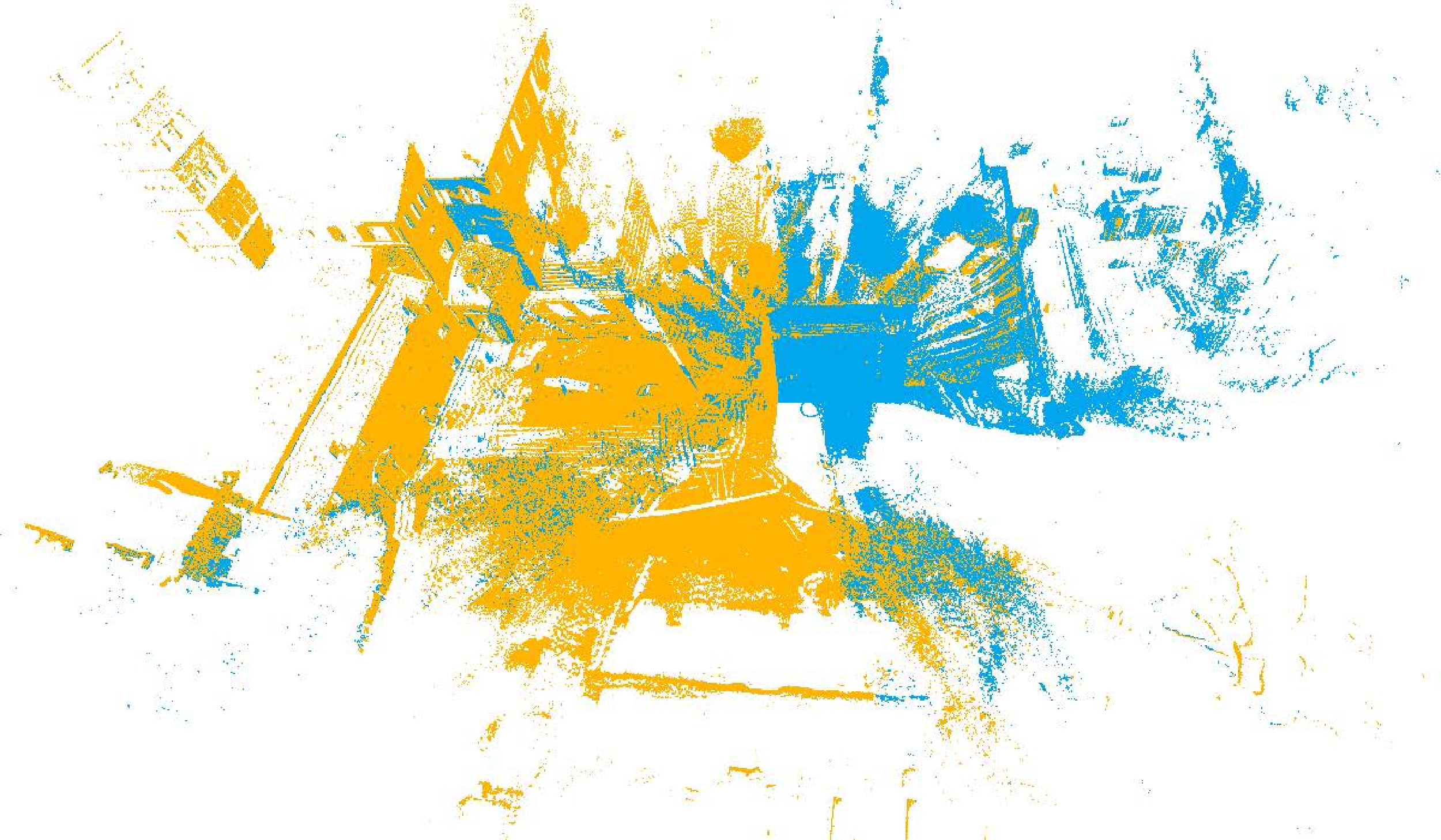} \\ (a-2) \textsf{GORE}~\cite{bustos2017guaranteed} \\ $\textit{E}_\mathbf{R}$ = 0.063$^{\circ}$, \ $\textit{E}_\mathbf{t}$ = 3.2cm \\ t = 221.59s \\[0.5em] \includegraphics[width=0.15\textwidth]{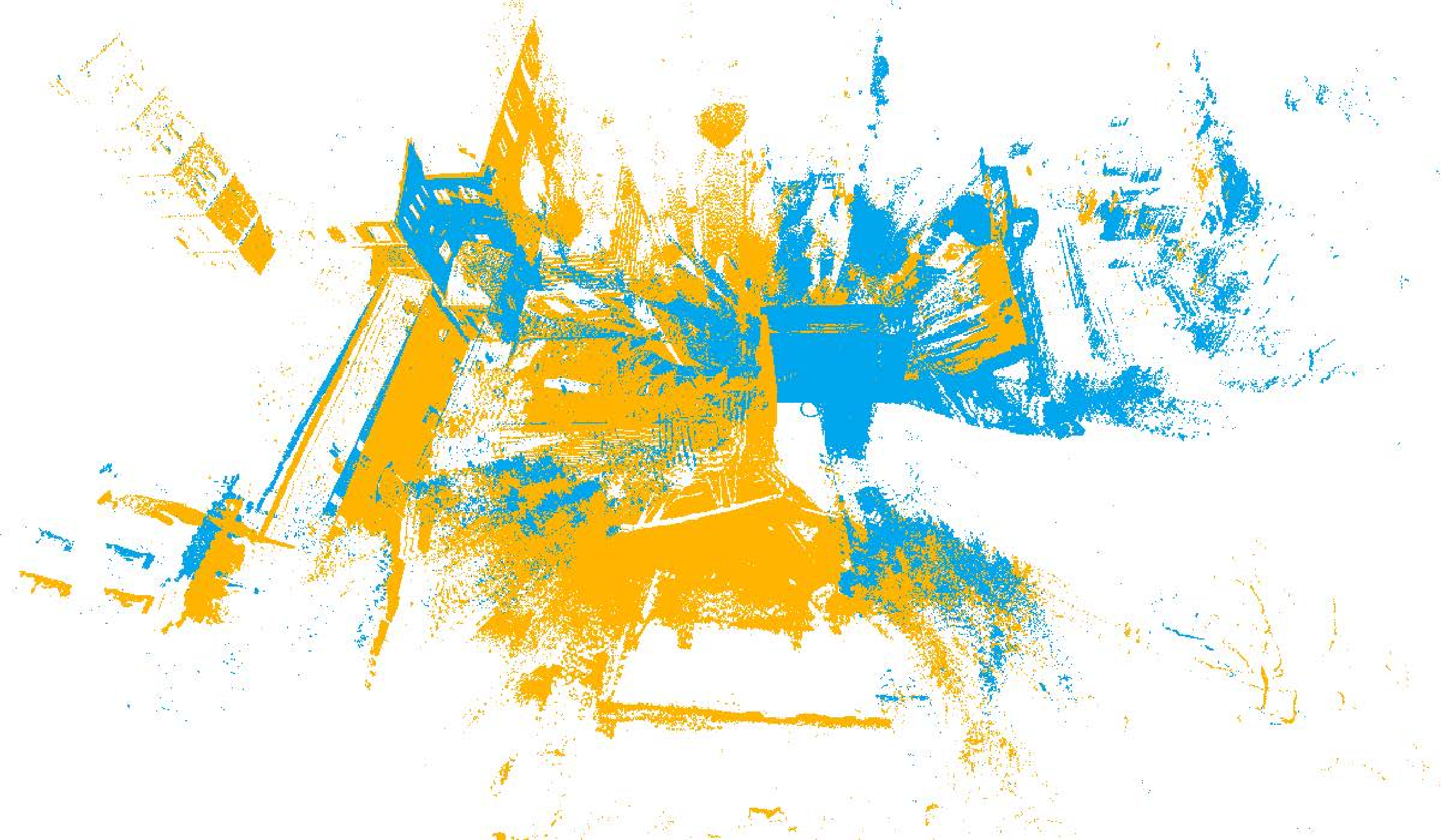} \\(a-5) \textsf{TR-DE}~\cite{chen2022deterministic} \\ $\textit{E}_\mathbf{R}$ = 0.67$^{\circ}$, \ $\textit{E}_\mathbf{t}$ = 8.32cm \\ t = 289.02s} &
      \makecell{\includegraphics[width=0.15\textwidth]{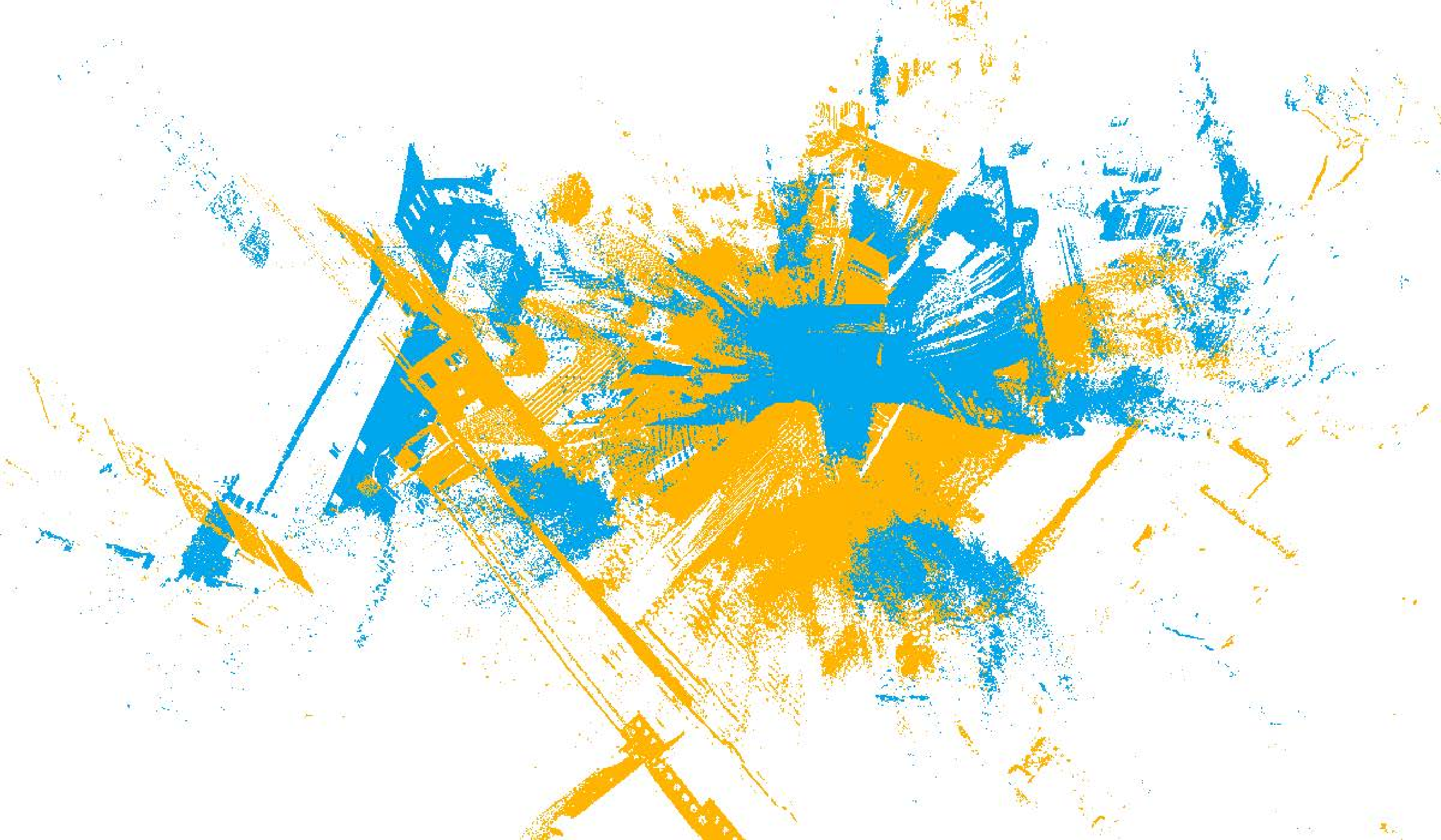} \\ (a-3) \textsf{GC-RANSAC}~\cite{barath2021graph} \\ $\textit{E}_\mathbf{R}$ = 50.87$^{\circ}$, \ $\textit{E}_\mathbf{t}$ = 1128.08cm \\ t = \color{gray}1.02s \\[0.5em]  \includegraphics[width=0.15\textwidth]{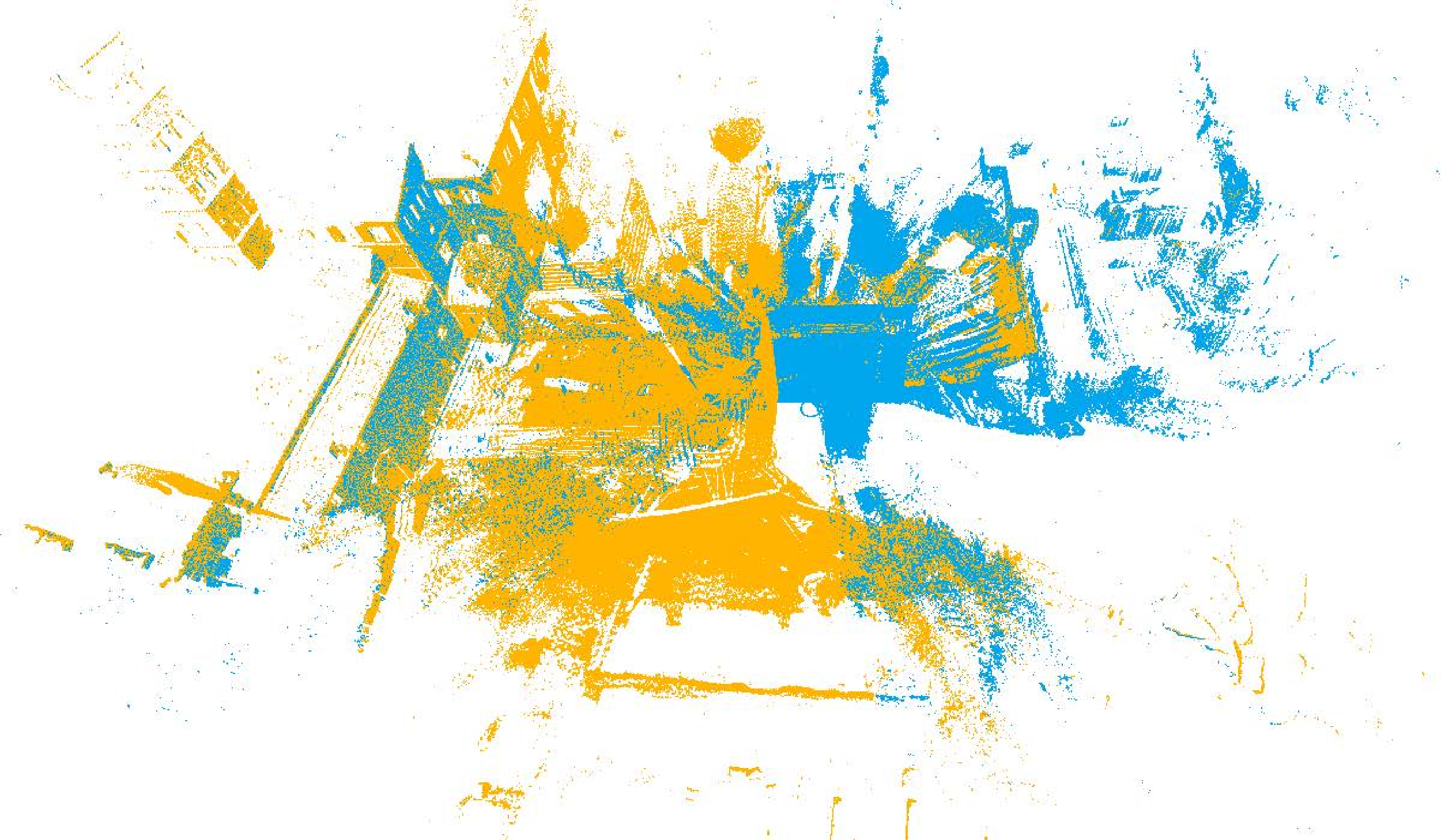} \\ (a-6) \textsf{SC$^2$-PCR}~\cite{chen2022sc2} \\ $\textit{E}_\mathbf{R}$ = \textbf{0.043$^{\circ}$}, \ $\textit{E}_\mathbf{t}$ = \underline{1.9cm} \\ t = 16.05s} &
      \makecell{\includegraphics[width=0.15\textwidth]{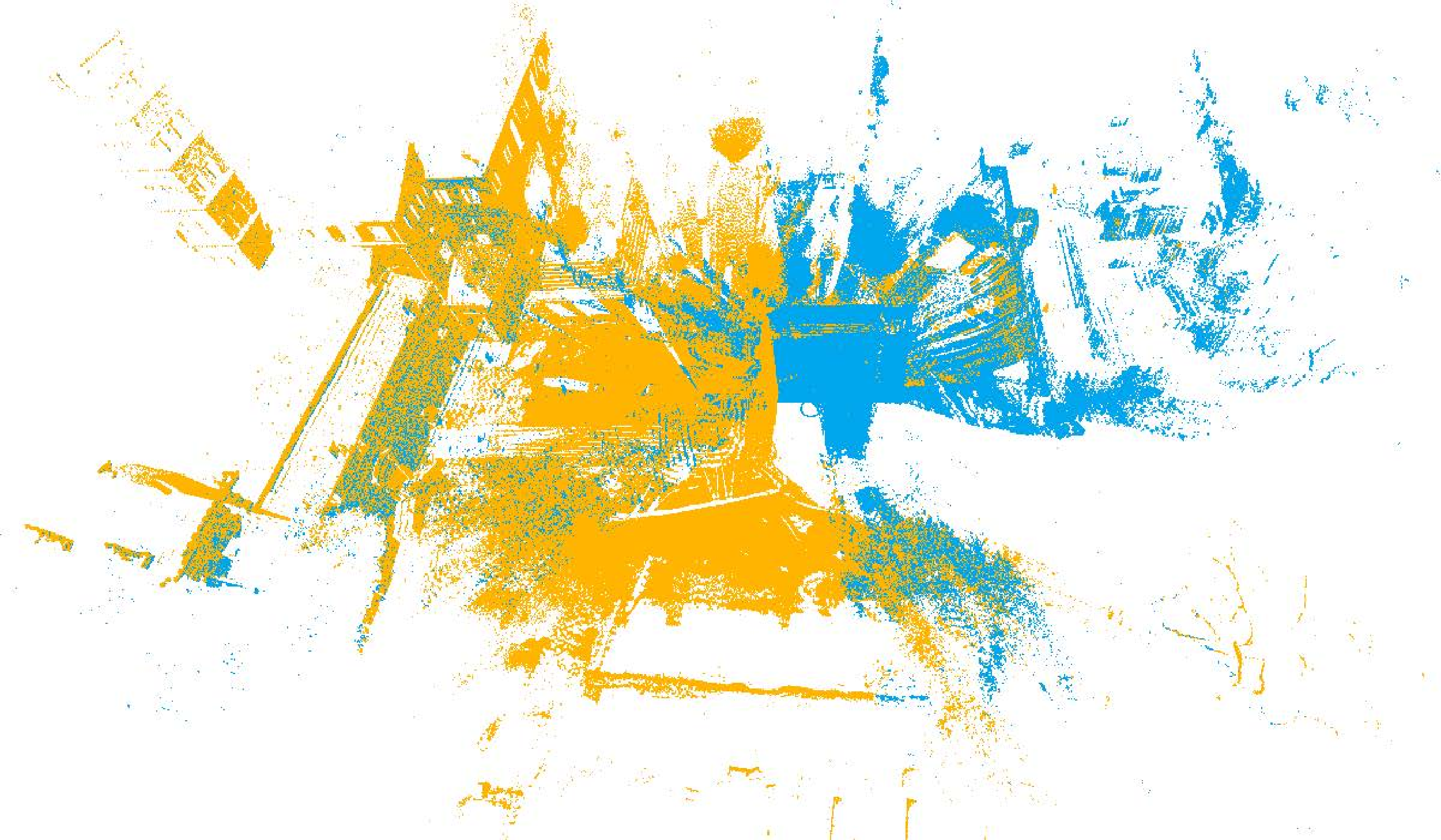} \\ (a-4) \textsf{TEASER++}~\cite{yang2020teaser} \\ $\textit{E}_\mathbf{R}$ = 0.058$^{\circ}$,\ $\textit{E}_\mathbf{t}$ = 2.9cm \\ t = \underline{2.78s} \\[0.5em] \includegraphics[width=0.15\textwidth]{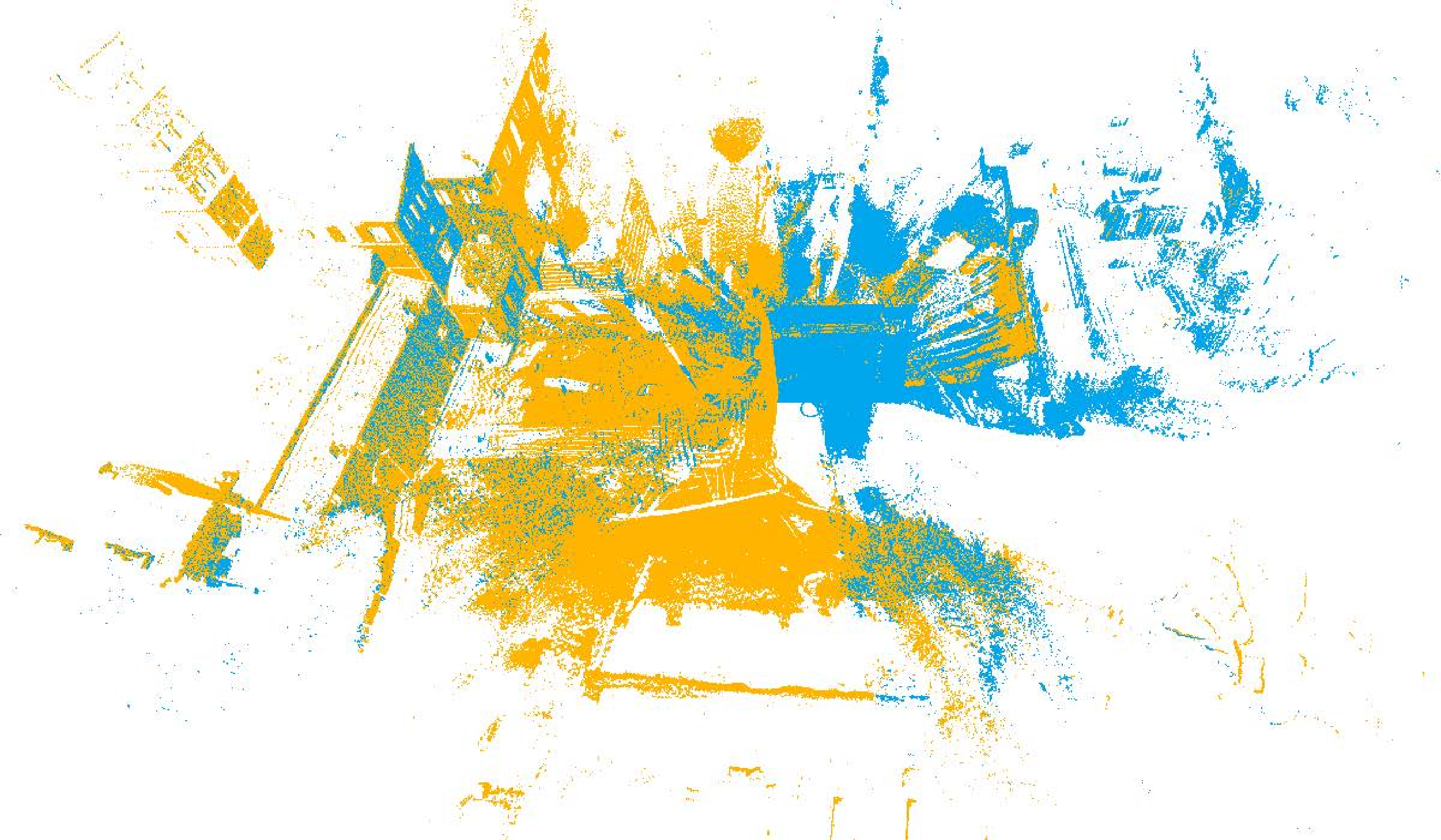} \\ (a-7) \textsf{HERE}~(ours) \\ $\textit{E}_\mathbf{R}$ = \underline{0.045$^{\circ}$}, \ $\textit{E}_\mathbf{t}$ = \textbf{1.6cm} \\ t = \textbf{0.47s}} 
      \\[10em]
      \makecell{\includegraphics[width=0.12\textwidth]{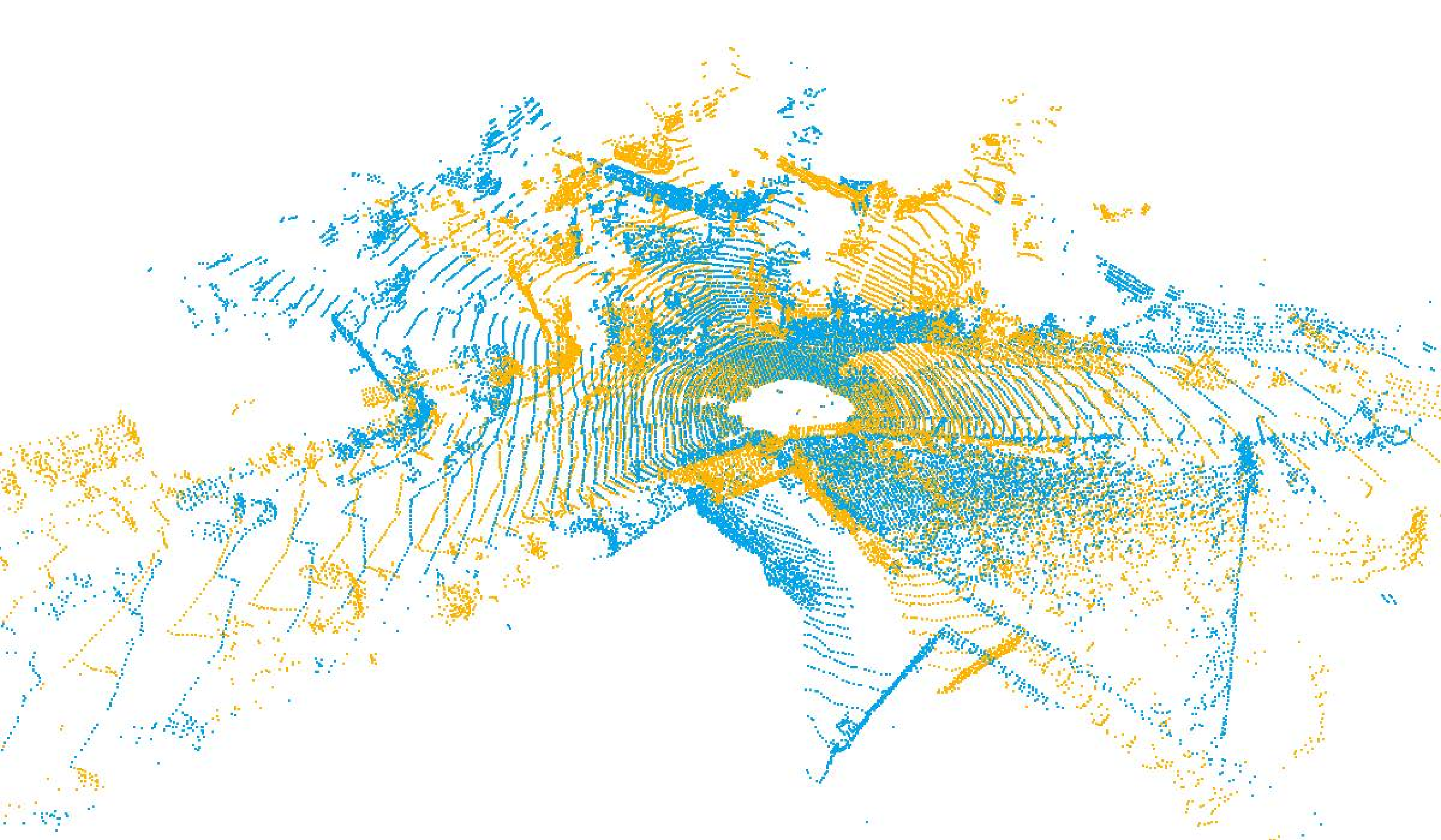} \\ (b-1) Inputs - KITTI Dataset \\ N = 5000\\ $\rho$ = 97.80$\%$} &
      \makecell{\includegraphics[width=0.12\textwidth]{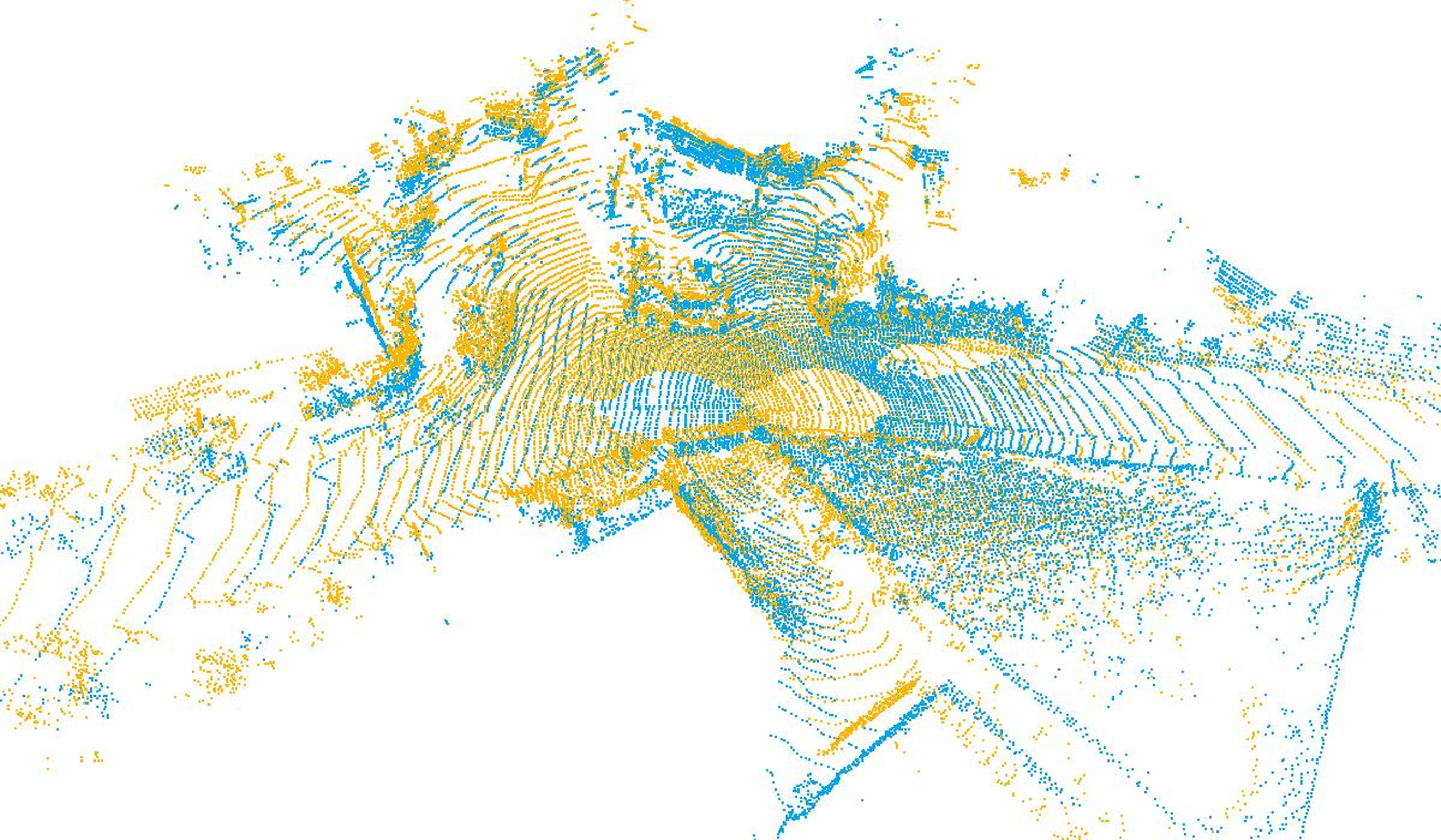} \\ (b-2) \textsf{RANSAC}-10k~\cite{fischler1981random} \\ $\textit{E}_\mathbf{R}$ = 1.55$^{\circ}$, \ $\textit{E}_\mathbf{t}$ = 35.78cm \\ t = \underline{0.25s} \\[0.5em]  \includegraphics[width=0.12\textwidth]{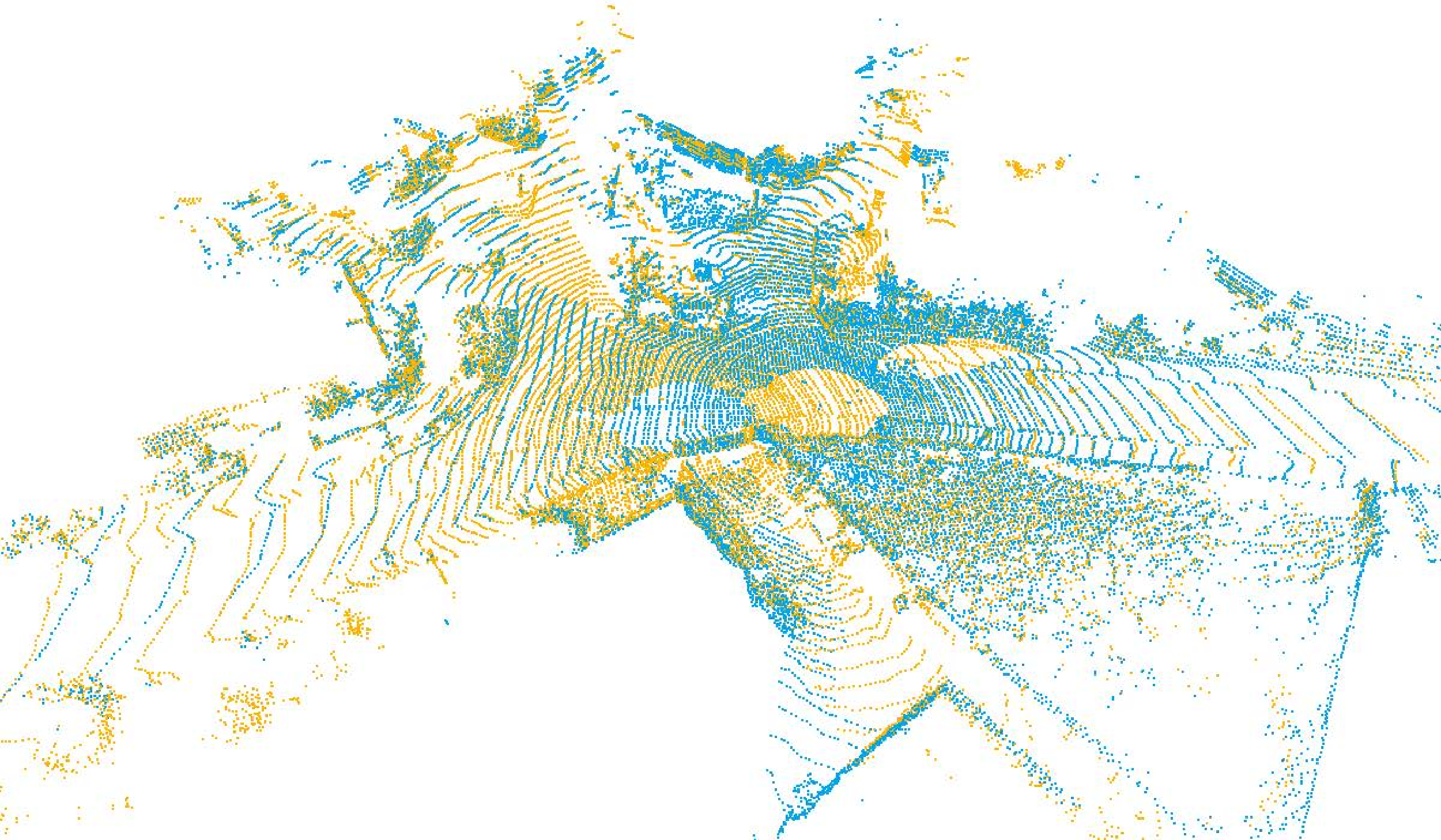} \\  (b-5) \textsf{SC$^2$-PCR}~\cite{chen2022sc2} \\ $\textit{E}_\mathbf{R}$ = 0.44$^{\circ}$, \ $\textit{E}_\mathbf{t}$ = 13.65cm \\ t = 4.14s} &
      \makecell{\includegraphics[width=0.12\textwidth]{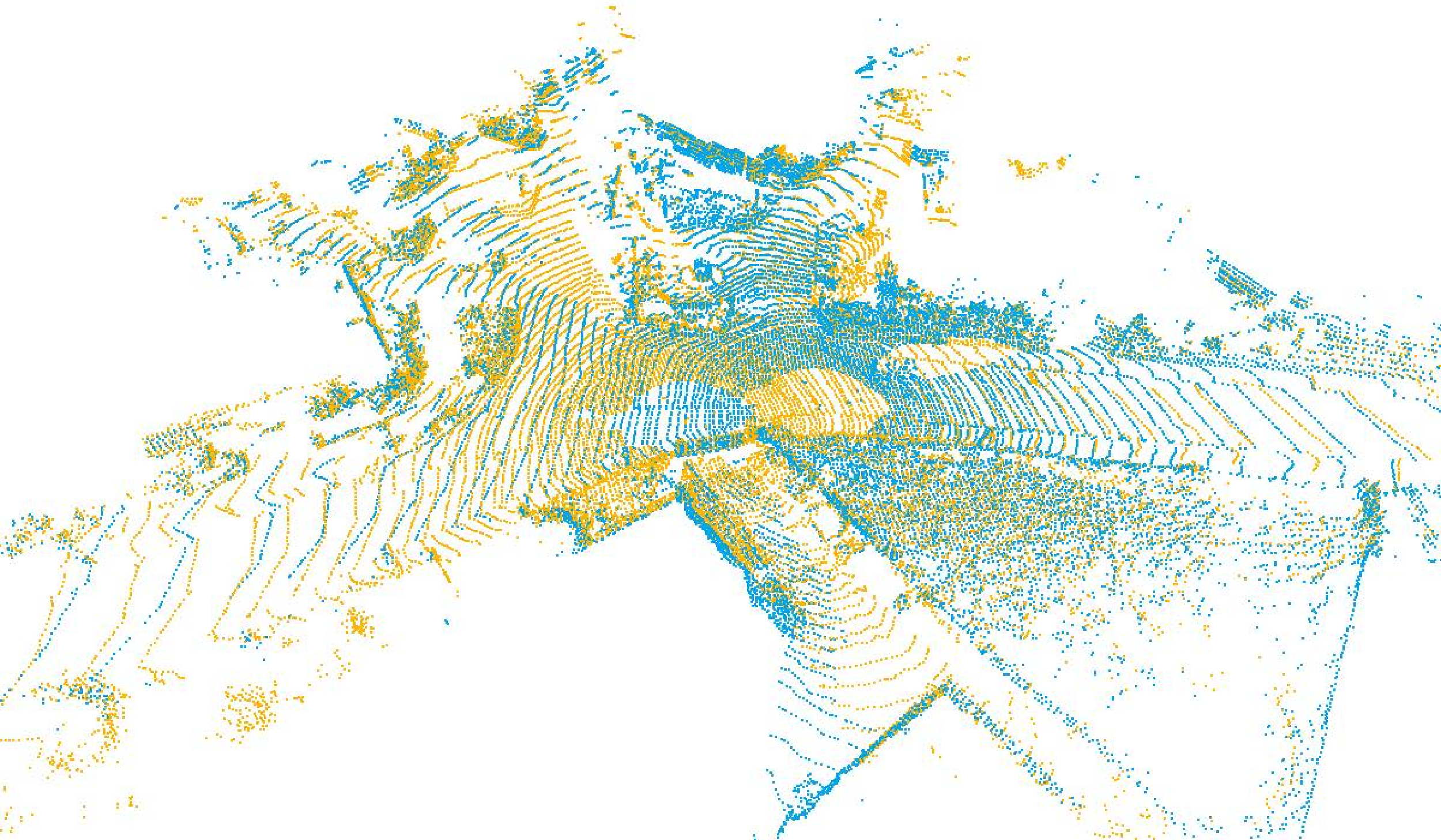} \\ (b-3) \textsf{GORE}~\cite{bustos2017guaranteed} \\ $\textit{E}_\mathbf{R}$ = 0.53$^{\circ}$, \ $\textit{E}_\mathbf{t}$ = 23.34cm \\ t = 603.52s \\[0.5em]  \includegraphics[width=0.12\textwidth]{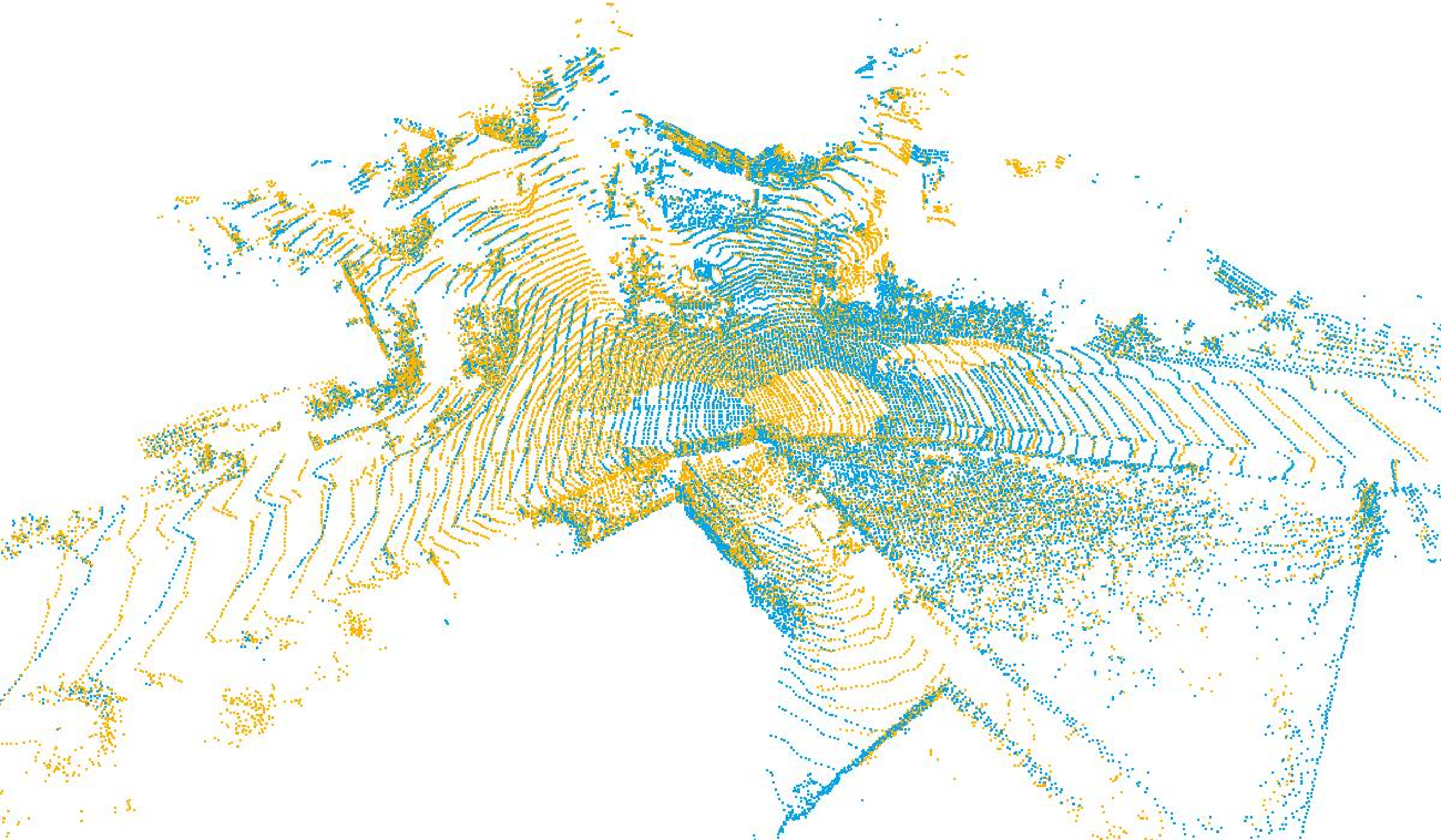} \\(b-6) \textsf{MAC}~\cite{Zhang_2023_CVPR} \\ $\textit{E}_\mathbf{R}$ = \textbf{0.38$^{\circ}$}, \ $\textit{E}_\mathbf{t}$ = 10.17cm \\ t = 2.85s} &
      \makecell{\includegraphics[width=0.12\textwidth]{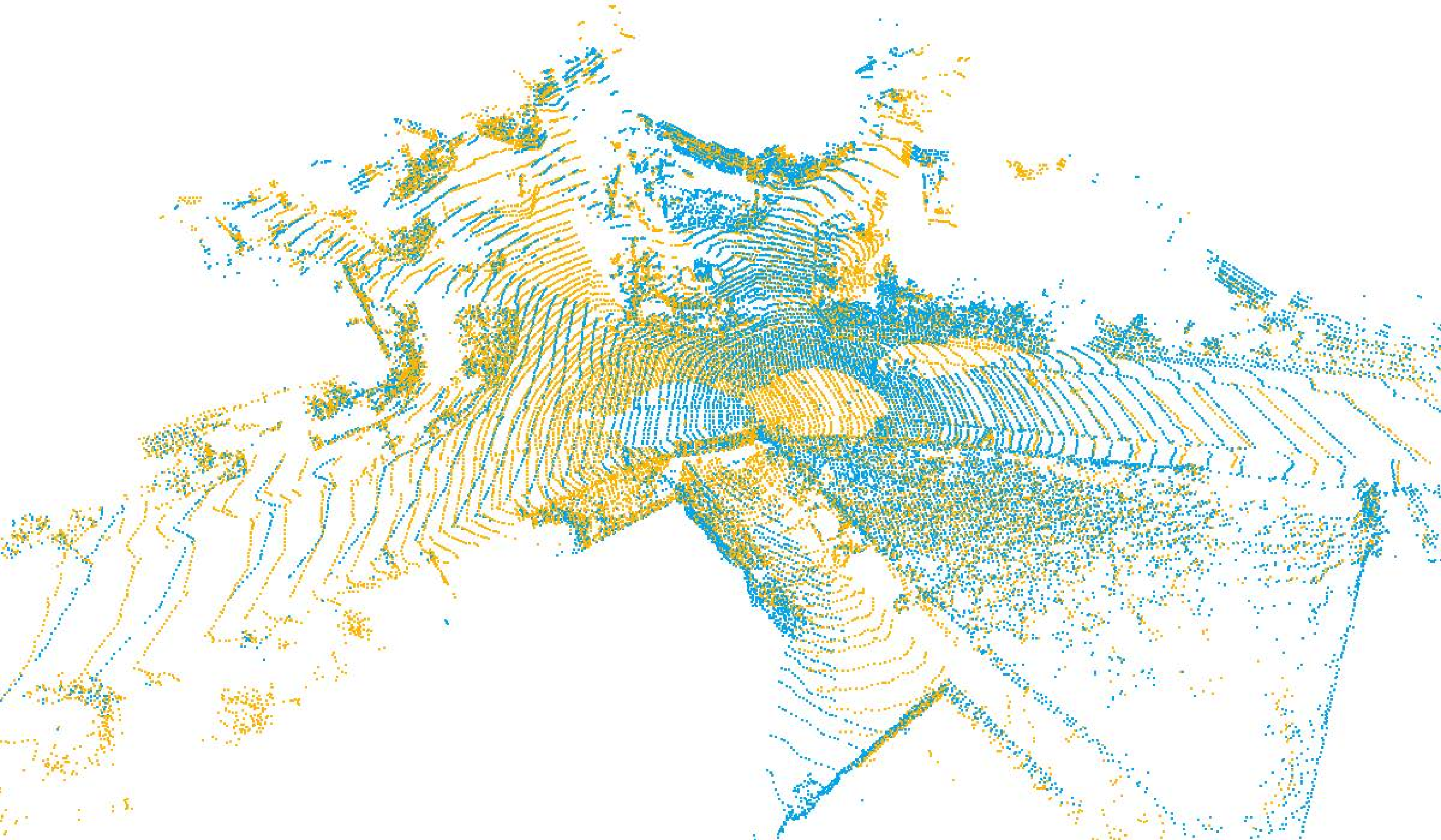} \\ (b-4) \textsf{GC-RANSAC}~\cite{barath2021graph} \\ $\textit{E}_\mathbf{R}$ = 0.43$^{\circ}$,\ $\textit{E}_\mathbf{t}$ = \underline{9.65cm} \\ t = 0.92s \\[0.5em] \includegraphics[width=0.12\textwidth]{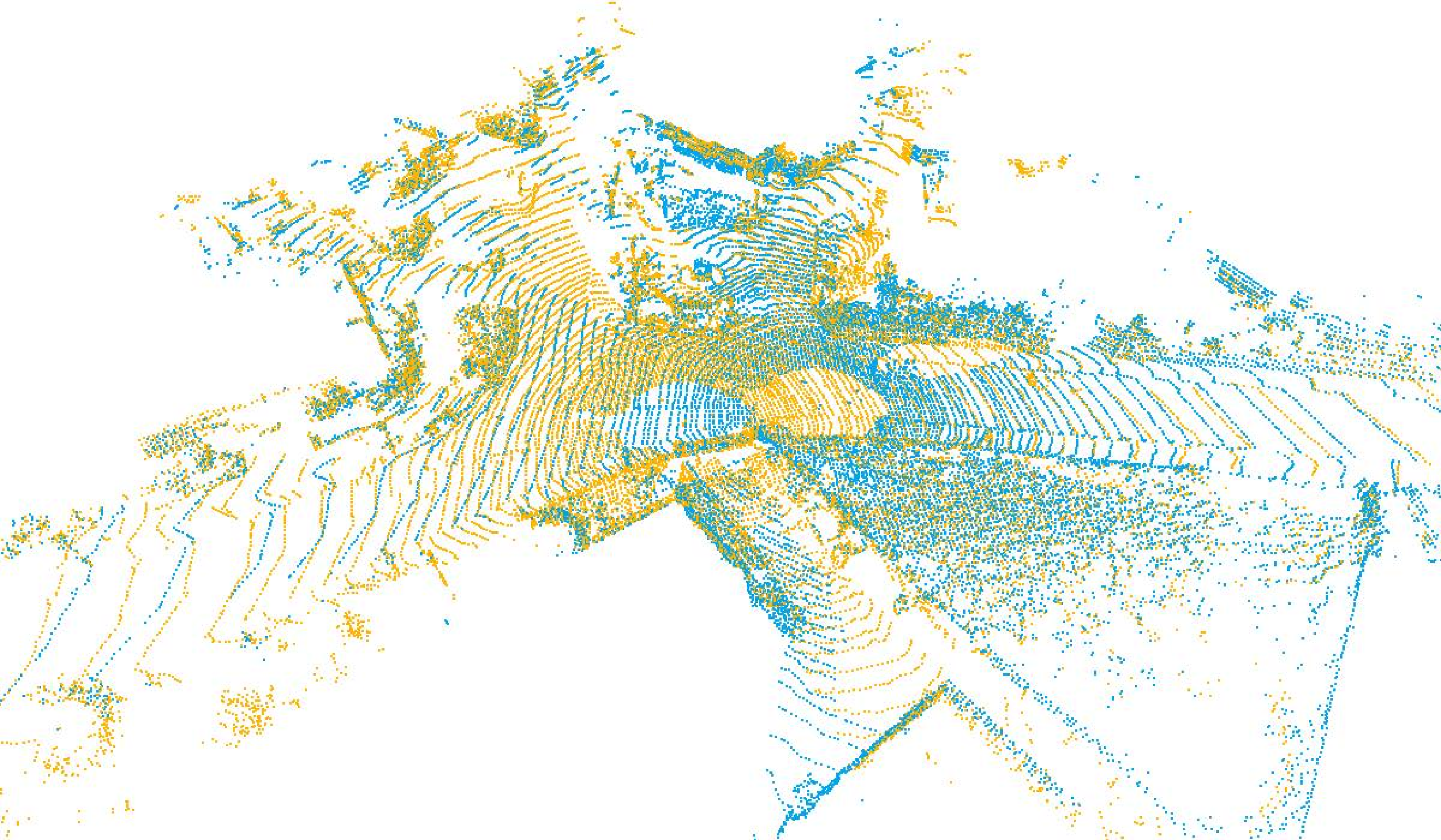} \\ (b-7) \textsf{HERE}~(ours) \\ $\textit{E}_\mathbf{R}$ = \underline{0.39$^{\circ}$}, \ $\textit{E}_\mathbf{t}$ = \textbf{9.12cm} \\ t = \textbf{0.19s}} \\
    \end{tabular}
    \vspace{-0.4em}
    \caption{Two representative comparison cases on the outdoor (a) ETH~\cite{theiler2014keypoint} and (b) KITTI~\cite{geiger2012we} datasets. The first column presents the input cloud pairs from different datasets. The second, third, and fourth columns show the registration results of different methods.}
    \label{fig: quant_outdoor}
\end{figure*}

\subsection{Comparison on Real-World Datasets}
\label{subsec: exper_real}

In this section, we compare all the methods on real-world datasets that contain outdoor and indoor scenes. Regarding the evaluation metrics, we follow~\cite{chen2022sc2, chen2022deterministic} to report the results in terms of both registration accuracy and outlier removal performance. 
To evaluate the registration accuracy, we report the registration recall~(\textit{RR}), i.e., the percentage of successful registration where the rotation and translation errors are below specific thresholds.
And we further report the average $\textit{E}_\mathbf{R}$ and $\textit{E}_\mathbf{t}$ on successfully registered pairs.
To evaluate the outlier removal performance, we report the inlier precision~(\textit{IP})=$\frac{\#\textnormal{kept true inliers}}{\#\textnormal{kept correspondences}}$ and the inlier recall~(\textit{IR})=$\frac{\#\textnormal{kept true inliers}}{\#\textnormal{true inliers}}$. We also report the $F_1$-score that considers both precision and recall, i.e., $F_1 = \frac{2\cdot\textit{IP}\cdot\textit{IR}}{\textit{IP}+\textit{IR}}$.

\subsubsection{Evaluation on Outdoor Scenes}
We conduct comparison experiments on the ETH~\cite{theiler2014keypoint} and KITTI~\cite{geiger2012we} datasets, both of which contain large-scale outdoor scenes.

\noindent \textbf{ETH Dataset.} 
The ETH dataset is a challenging large-scale dataset consisting of five distinct scenes: arch, courtyard, facade, office, and trees. We follow~\cite{theiler2014keypoint} to process the raw point clouds and generate the putative correspondences. Specifically, we first downsample the dense point clouds with a 0.1m voxel grid filter and set the inlier threshold $\xi$ = 0.3m. We then adopt ISS~\cite{zhong2009intrinsic} and FPFH~\cite{rusu2009fast} to extract the feature descriptors. Based on these descriptors, we establish the putative correspondence set based on the top-5 nearest neighbor search in the feature space. Due to moving objects and low overlapping ratios, the outlier ratios of each pair in the dataset can be very high~(averagely $>$ 95$\%$). 
We consider a registration successful if the rotation error $\textit{E}_\mathbf{R} < 3^{\circ}$ and the translation error $\textit{E}_\mathbf{t} < 0.5$m.

Table.~\ref{tab: ETH_acc} reports the registration recall~(\textit{RR}), rotation error~($\textit{E}_\mathbf{R}$), translation error~($\textit{E}_\mathbf{t}$), and timing results of each method evaluated on different scenes in the ETH dataset. 
Regarding the \textit{RR}, \textsf{TEASER++} and the proposed \textsf{HERE} both show higher robustness compared with the other methods. 
However, when the number of correspondences is large~(e.g., over 14k correspondences in the Trees scene), \textsf{HERE} can run more than 4 times faster than \textsf{TEASER++}. In addition, \textsf{HERE} generally leads to lower registration errors than \textsf{TEASER++}. 
Regarding the registration error, \textsf{SC$^2$-PCR} generally shows the highest accuracy, with \textsf{HERE} being the second best. 
However, \textsf{SC$^2$-PCR} can be very slow when dealing with large-scale correspondences~(e.g., averagely 39s in the Trees scene and 14s in the Arch scene).
By contrast, \textsf{HERE} can run more than 10 times faster than \textsf{SC$^2$-PCR}. Moreover, \textsf{HERE} leads to higher registration recall than \textsf{SC$^2$-PCR}. 
As to the recent \textsf{MAC}, even though it also shows high robustness, its efficiency is limited when dealing with large-scale point clouds~(e.g., the Courtyard and Trees scenes). 
Fig.~\ref{fig: quant_outdoor}(a) presents a compelling comparison case in the ETH dataset, where \textsf{HERE} achieves the lowest $\textit{E}_\mathbf{t}$ and is roughly 6 times faster than the second fastest \textsf{TEASER++}.

\noindent \textbf{KITTI Dataset.}
Following~\cite{chen2022sc2, choy2019fully}, we use the 8$^{\textnormal{th}}$ to 10$^{\textnormal{th}}$ scenes of the KITTI dataset to evaluate all the methods. We downsample each pair of dense point clouds with a 30cm voxel grid and set the inlier threshold $\xi=$ 60cm. For putative correspondences, we follow~\cite{chen2022sc2} to adopt FPFH~\cite{rusu2009fast} and FCGF~\cite{choy2019fully} as feature descriptors, respectively. To evaluate successful registration, we set the threshold as $\textit{E}_\mathbf{R}\le5^\circ$ and $\textit{E}_\mathbf{t}\le60$cm. Note that we do not report the quantitative results of \textsf{GORE} since it is slow and requires 100s - 10h to terminate for the dataset under consideration.

\begin{table*}[!t]
\centering
    \caption{Quantitative comparison results regarding registration accuracy, outlier removal performance, and efficiency on the indoor 3DMatch~\cite{zeng20173dmatch} and 3DLoMatch~\cite{huang2021predator} datasets.}
    \vspace{-1em}
    \label{tab: 3DMatch_nomutual}
    \footnotesize
    \renewcommand{\tabcolsep}{2.6pt} 
    \renewcommand\arraystretch{1.5}
    \begin{tabular}{c|ccccccc|ccccccc}
        \Xhline{1pt}
        Method & \multicolumn{7}{c|}{3DMatch with FCGF Descriptor~\cite{choy2019fully}} & \multicolumn{7}{c}{3DLoMatch with Predator Descriptor~\cite{huang2021predator}} \\
        \cline{2-15}
         & \textit{RR}(\%)$\uparrow$ & $\textit{E}_\mathbf{R}$($^{\circ}$)$\downarrow$ & $\textit{E}_\mathbf{t}$(cm)$\downarrow$ & \textit{IP}(\%)$\uparrow$ &  \textit{IR}(\%)$\uparrow$ & $F_1$(\%)$\uparrow$ & Time(s)$\downarrow$ & \textit{RR}(\%)$\uparrow$ & $\textit{E}_\mathbf{R}$($^{\circ}$)$\downarrow$ & $\textit{E}_\mathbf{t}$(cm)$\downarrow$ & \textit{IP}(\%)$\uparrow$ &  \textit{IR}(\%)$\uparrow$ & $F_1$(\%)$\uparrow$ & Time(s)$\downarrow$\\
        \Xhline{0.5pt}
        \textsf{RANSAC}-1k~\cite{fischler1981random} & 85.03 & 3.14 & 9.40 & 77.80 & 79.65 & 78.34 & \textbf{0.05} & 61.20 & 4.10 & 12.45 & 56.73 & 63.87 & 59.34 & \textbf{0.04}\\
        \textsf{RANSAC}-10k~\cite{fischler1981random} & 88.29 & 2.68 & 8.24 & 78.74& 84.42& 81.22& 0.42 & 65.30 & 3.87 & 12.11 & 57.23 & 67.15 & 61.07 & 0.38\\
        \textsf{RANSAC}-100k~\cite{fischler1981random} & 89.22& 2.46& 7.60& 78.78& 86.94& 81.92& 4.05 & 66.03 & 3.76 & 11.82 & 56.96 & 67.86 & 61.20 & 3.75\\
        \textsf{FGR}~\cite{zhou2016fast} & 73.75 & 2.73 & 8.14 & 68.91 & 67.07 & 66.56 & \underline{0.17} & 38.90 & 3.90 & 11.66 & 45.22 & 42.48 & 41.46 & \color{gray}0.13\\
        \textsf{GC-RANSAC}~\cite{barath2021graph} & 89.65 & 2.36 & 7.23 & 77.75 & 84.33 & 80.62 & 0.91 & 64.18 & 3.39 & 11.21 & 55.52 & 65.95 & 59.57 & 1.07\\        
        \textsf{TEASER++}~\cite{yang2020teaser} & 85.77 & 2.91 & 9.40 & 79.28 & 84.27 & 81.24 & - & 63.17 & 4.17 & 10.58 & \underline{59.46} & 69.09 & \underline{62.93} & -\\
        \textsf{SC$^2$-PCR}~\cite{chen2022sc2} & \underline{92.73} & 2.20 & \textbf{6.88} & 79.68 & \textbf{87.15} & \underline{82.94} & 3.95 & 68.73 & \textbf{3.22} & 10.75 & 58.39 & \underline{69.25} & 62.61 & 4.16\\
        \textsf{TR-DE}~\cite{chen2022deterministic} & 86.99 & 2.62 & 8.03 & 77.31 & 86.09 & 82.63 & 134.45 & 66.03 & 4.32 & 11.04 & 57.32 & 66.65 & 61.65 & 243.53\\
        \textsf{MAC}~\cite{Zhang_2023_CVPR} & \textbf{92.79} & \underline{2.18} & 6.89 & \textbf{80.33} & \underline{86.99} & 82.28 & 3.72 & \textbf{69.17} & 3.42 & \underline{10.47} & \textbf{59.75} & \textbf{70.10} & \textbf{63.76} & 2.99 \\
        \textsf{HERE}~(ours) & 91.56 & \textbf{2.17} & \underline{6.93} & \underline{79.88} & 86.97 & \textbf{83.11} & 0.32 & \underline{68.89} & \underline{3.31} & \textbf{10.42} & 58.13 & \underline{69.25} & 62.37 & \underline{0.37}\\
        \Xhline{1pt}
    \end{tabular}
\end{table*}

\begin{figure*}
\centering
    \footnotesize
    \renewcommand{\tabcolsep}{10pt}
    \begin{tabular}{cccc}
      \makecell{\includegraphics[width=0.12\textwidth]{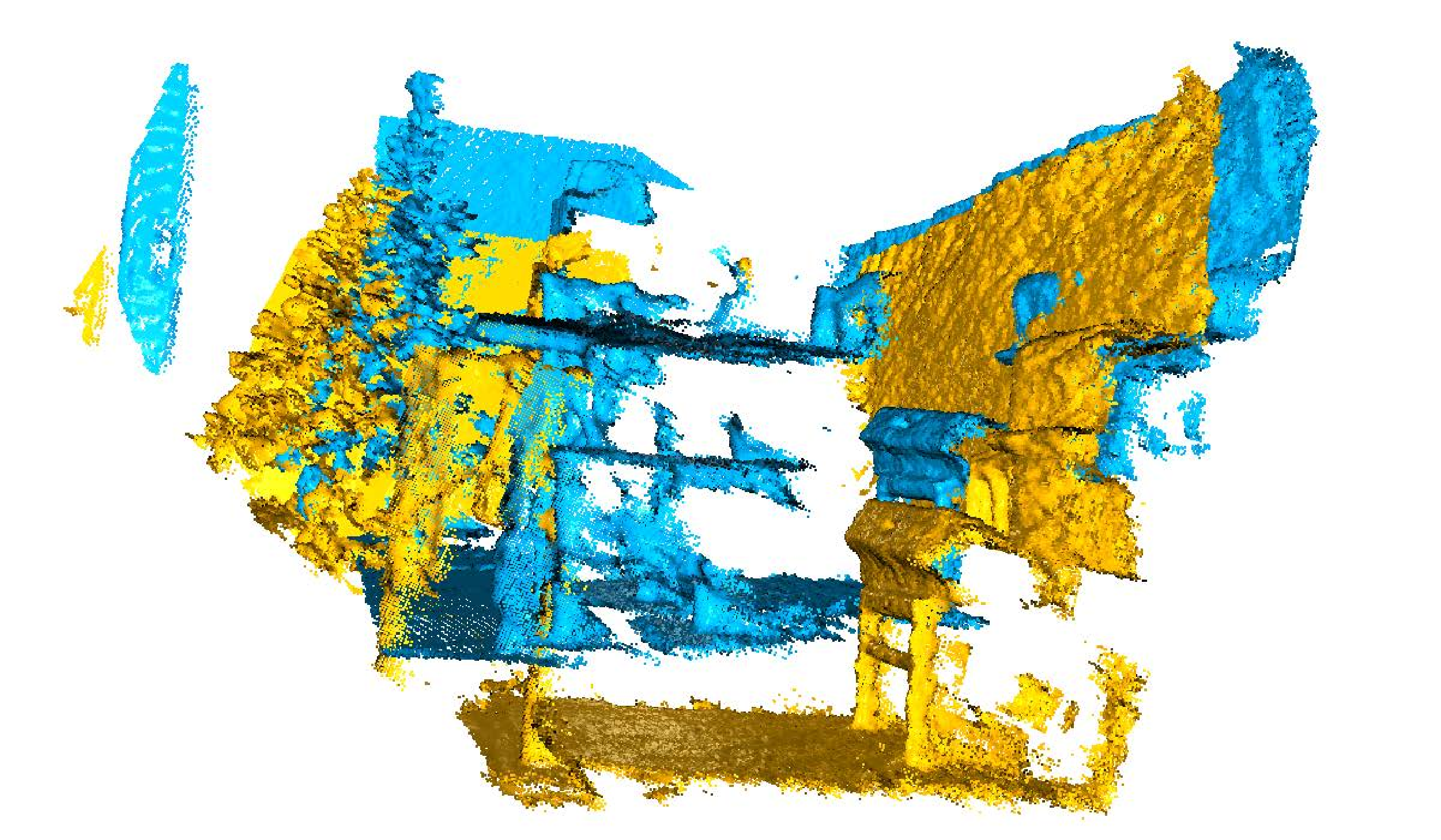} \\ (a-1) Inputs - 3DMatch Dataset \\ N = 7158 \\ $\rho$ = 65.07$\%$} &
      \makecell{\includegraphics[width=0.12\textwidth]{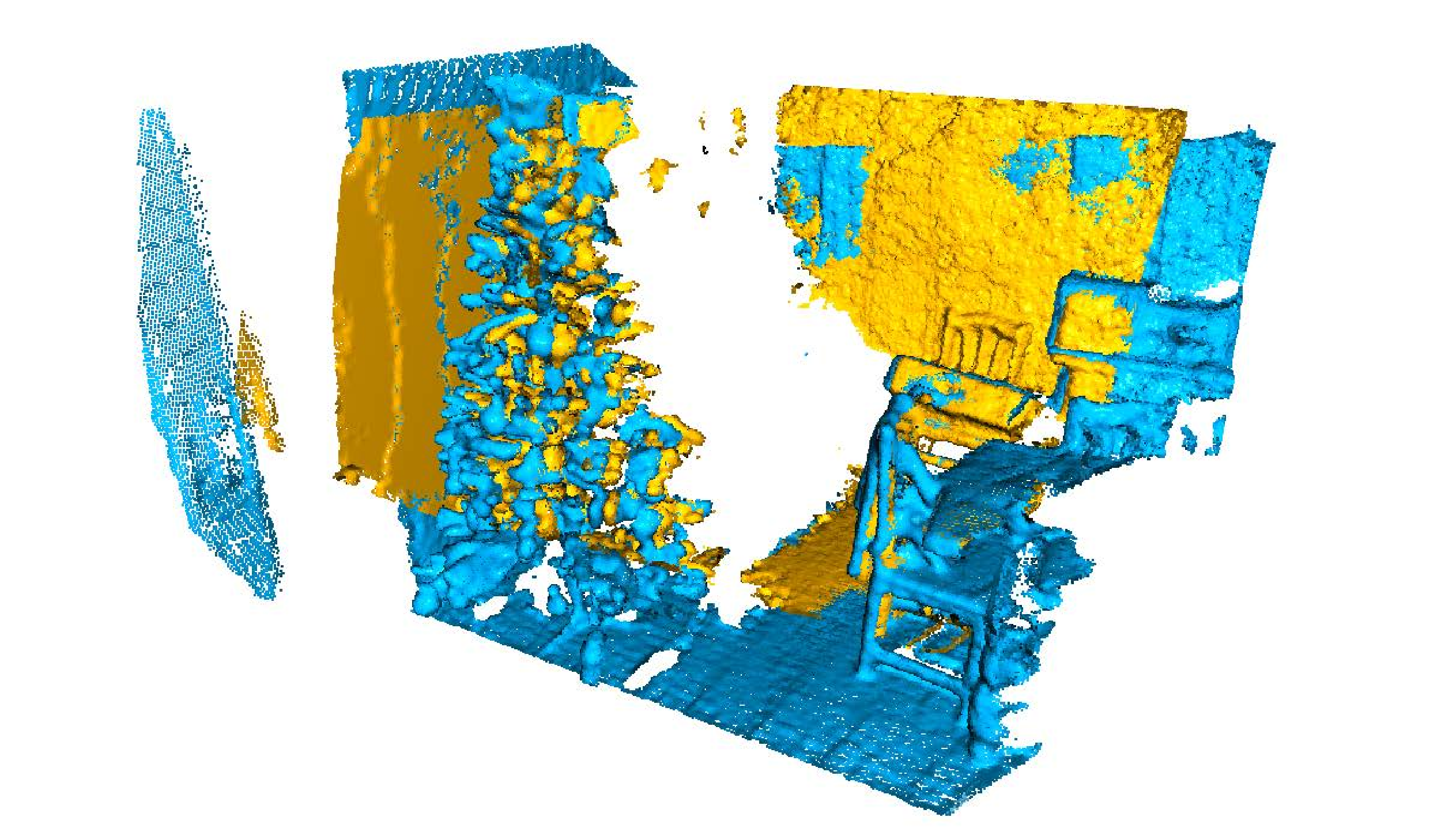} \\ (a-2) \textsf{RANSAC}-10k~\cite{fischler1981random} \\ $\textit{E}_\mathbf{R}$ = 2.98$^{\circ}$, \ $\textit{E}_\mathbf{t}$ = 3.51cm \\ t = \underline{0.46s} \\[0.5em] \includegraphics[width=0.12\textwidth]{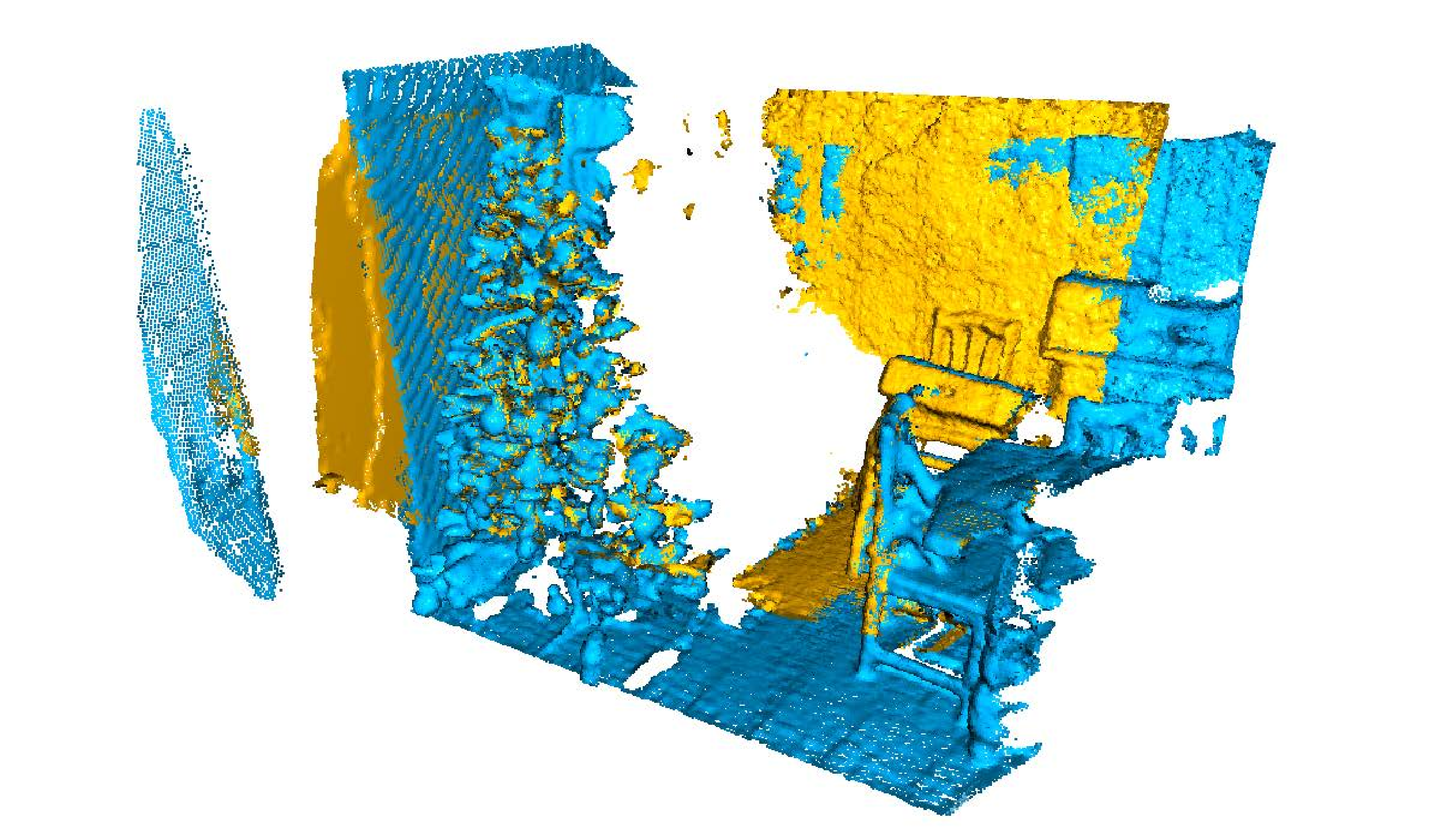} \\(a-5) \textsf{TEASER++}~\cite{yang2020teaser} \\ $\textit{E}_\mathbf{R}$ = 1.01$^{\circ}$, \ $\textit{E}_\mathbf{t}$ = 2.89cm \\ t = 891.85s} &
      \makecell{\includegraphics[width=0.12\textwidth]{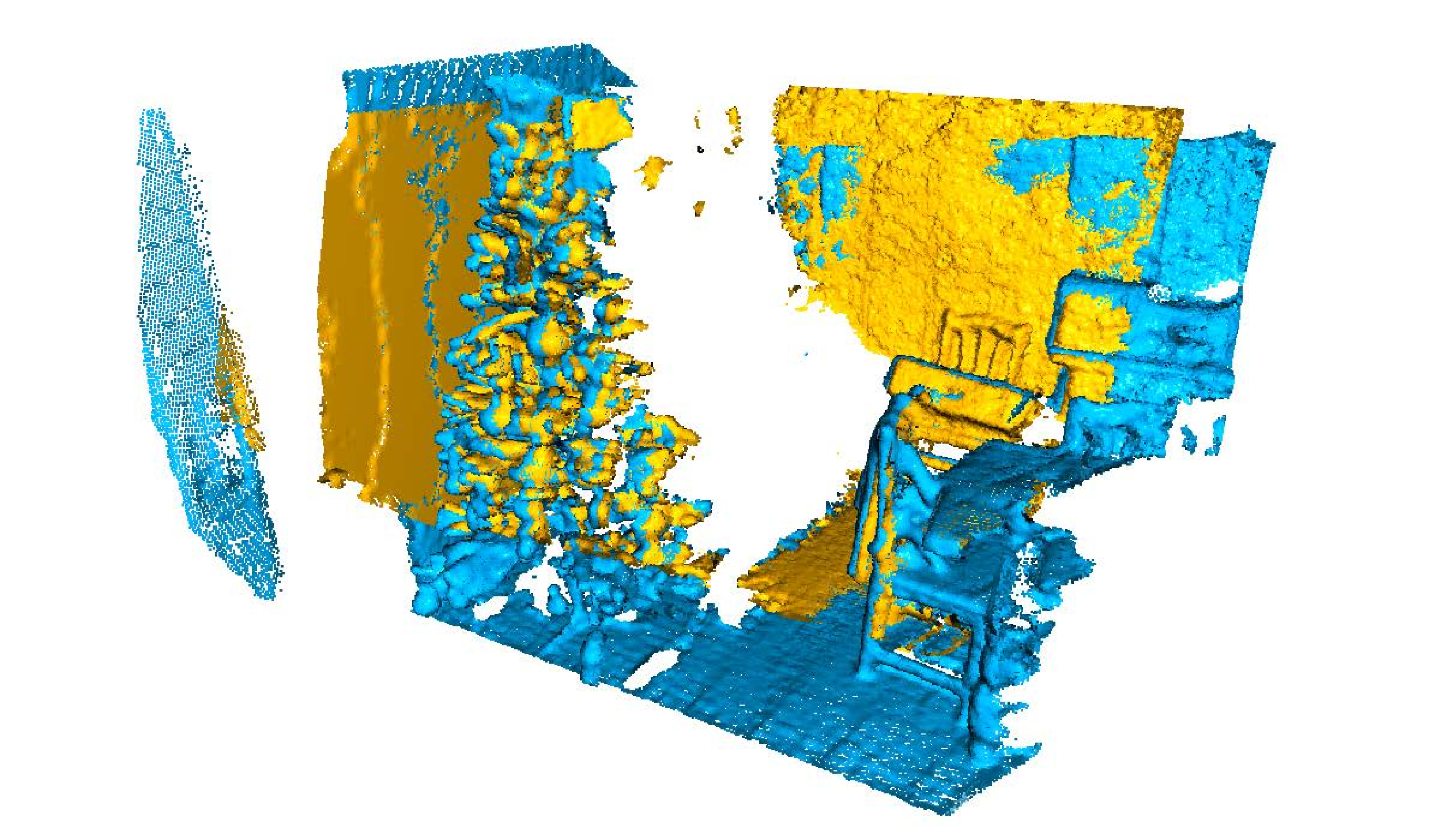} \\ (a-3) \textsf{GORE}~\cite{bustos2017guaranteed} \\ $\textit{E}_\mathbf{R}$ = 0.97$^{\circ}$, \ $\textit{E}_\mathbf{t}$ = 2.25cm \\ t = 18262.57s \\[0.5em]  \includegraphics[width=0.12\textwidth]{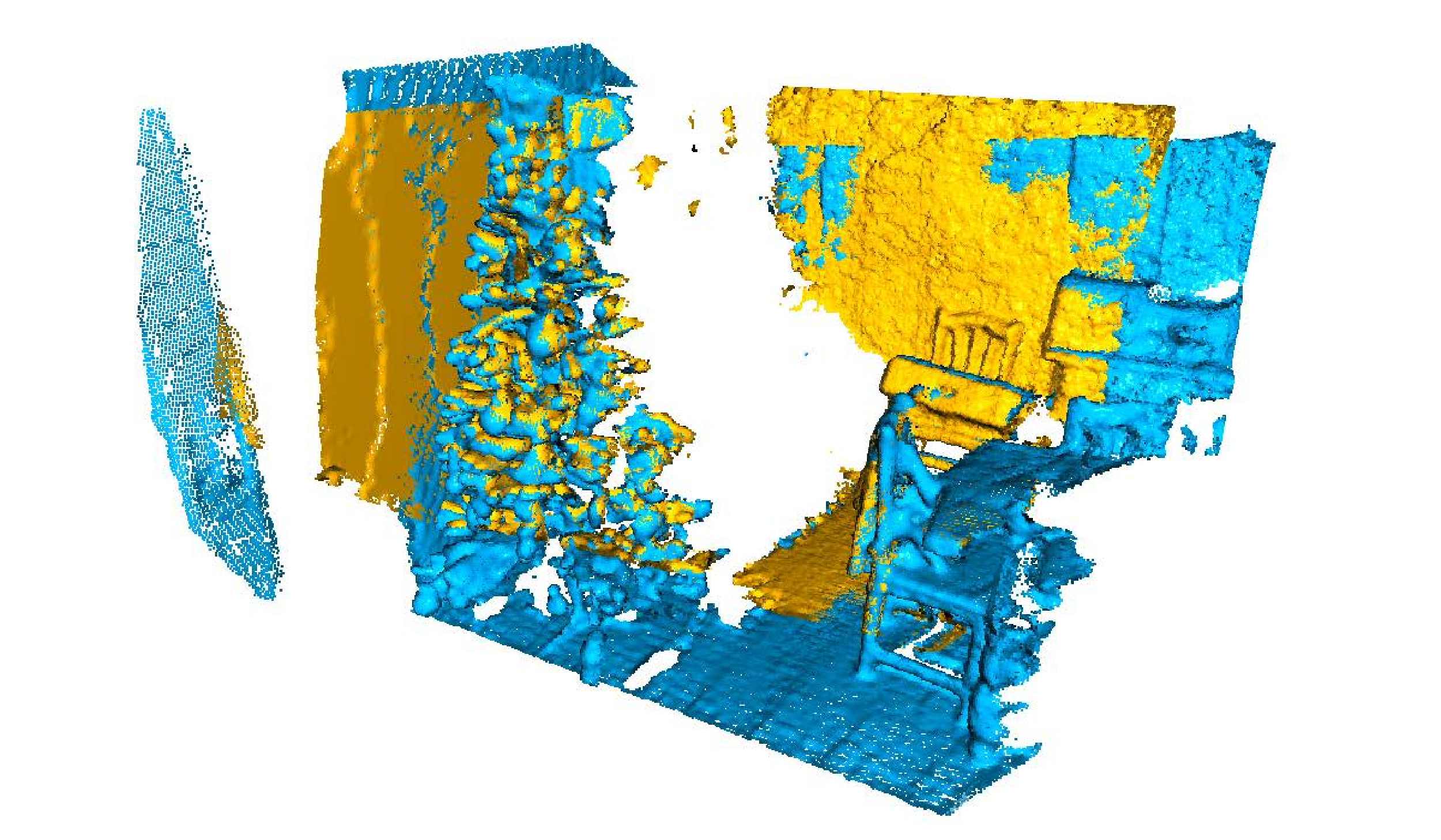} \\ (a-6) \textsf{SC$^2$-PCR}~\cite{chen2022sc2} \\ $\textit{E}_\mathbf{R}$ = \underline{0.89$^{\circ}$}, \ $\textit{E}_\mathbf{t}$ = \textbf{2.17cm} \\ t = 8.49s} &
      \makecell{\includegraphics[width=0.12\textwidth]{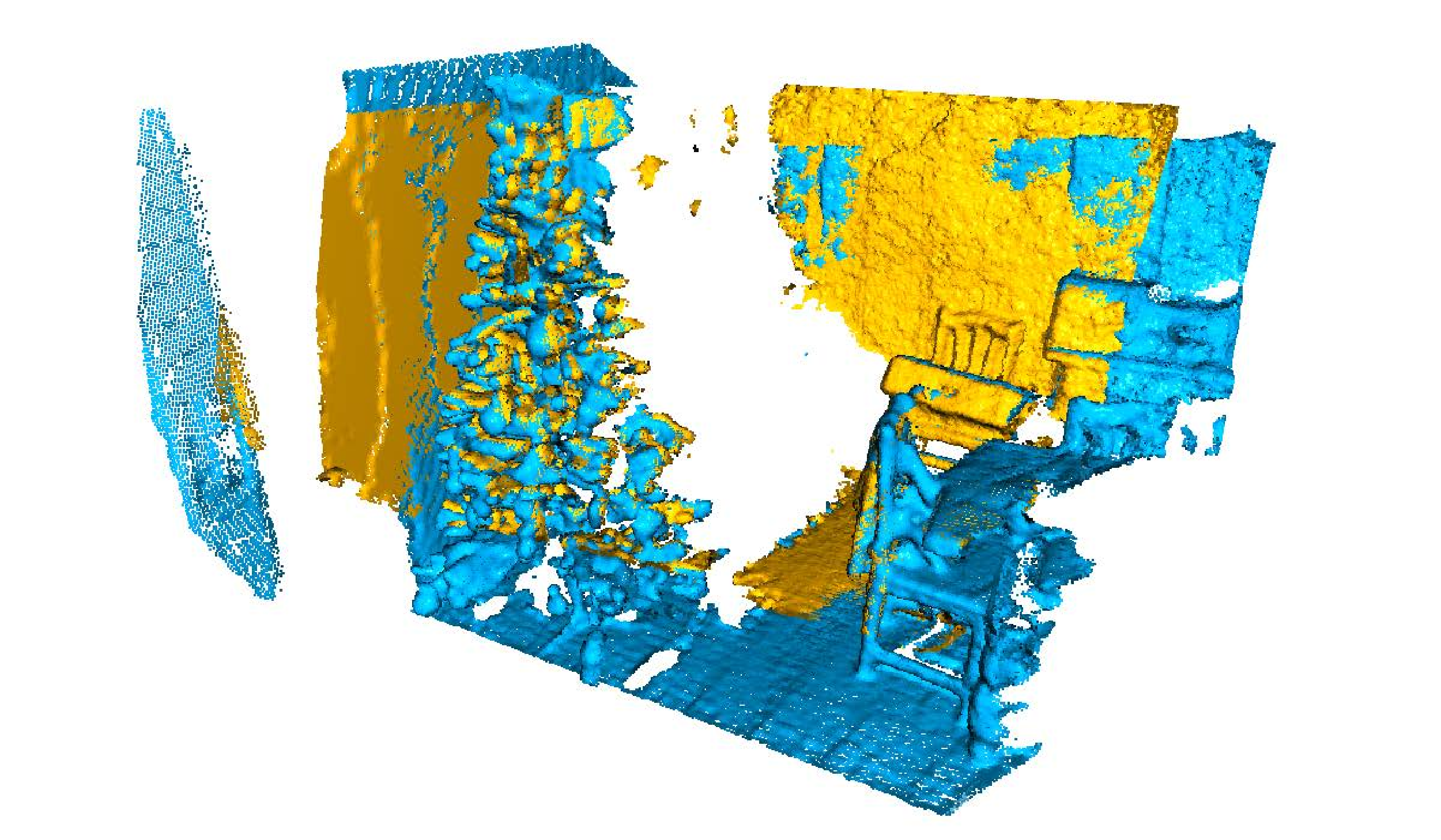} \\ (a-4) \textsf{GC-RANSAC}~\cite{barath2021graph} \\ $\textit{E}_\mathbf{R}$ = 1.07$^{\circ}$,\ $\textit{E}_\mathbf{t}$ = 2.28cm \\ t = 1.06s \\[0.5em] \includegraphics[width=0.12\textwidth]{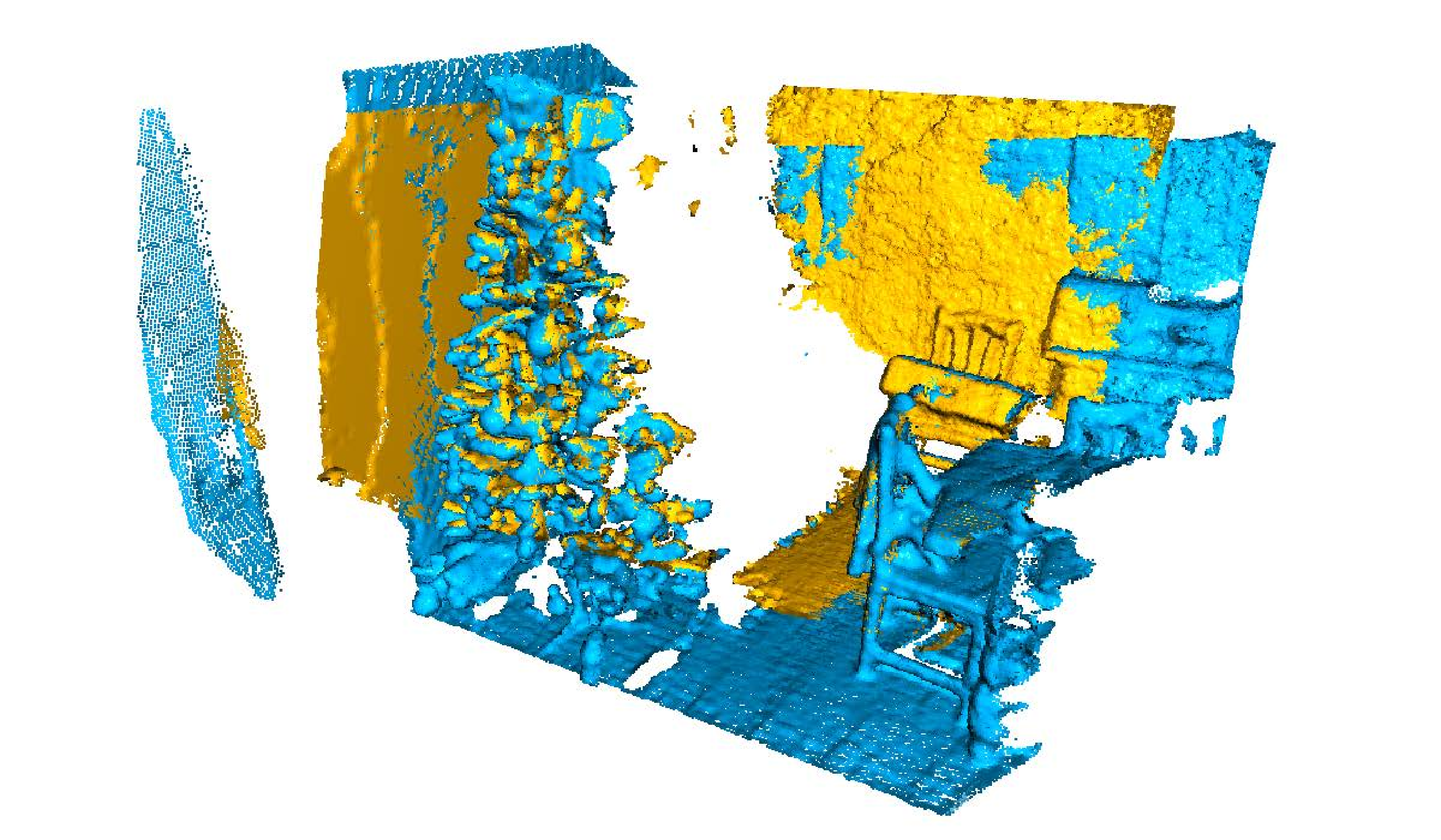} \\ (a-7) \textsf{HERE}~(ours) \\ $\textit{E}_\mathbf{R}$ = \ \textbf{0.85$^{\circ}$}, \ $\textit{E}_\mathbf{t}$ = \underline{2.21cm} \\ t = \textbf{0.35s}}
      \\[9em]
      \makecell{\includegraphics[width=0.12\textwidth]{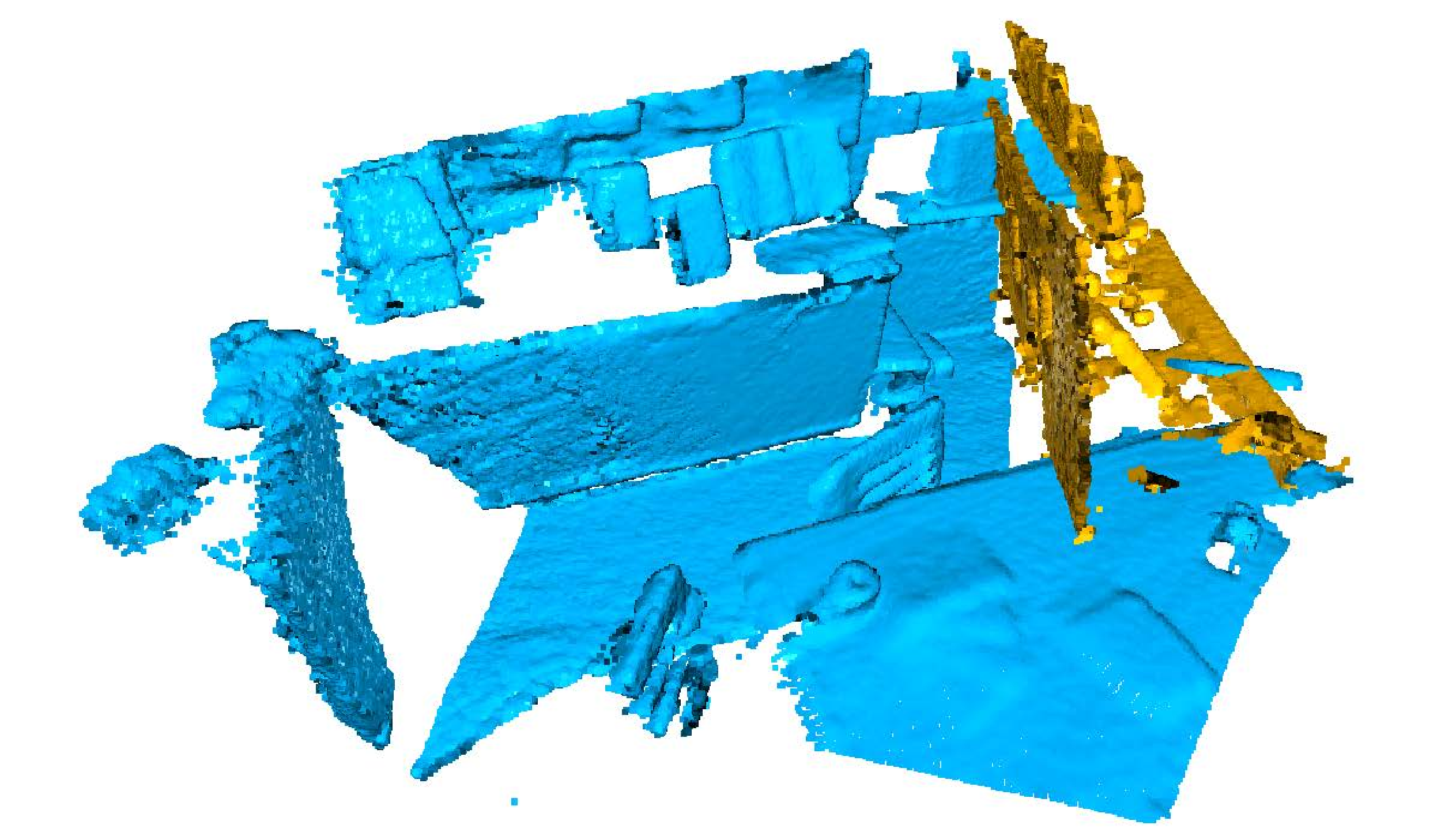} \\ (b-1) Inputs - 3DLoMatch Dataset \\ N = 5000 \\ $\rho$ = 92.02$\%$} &
      \makecell{\includegraphics[width=0.12\textwidth]{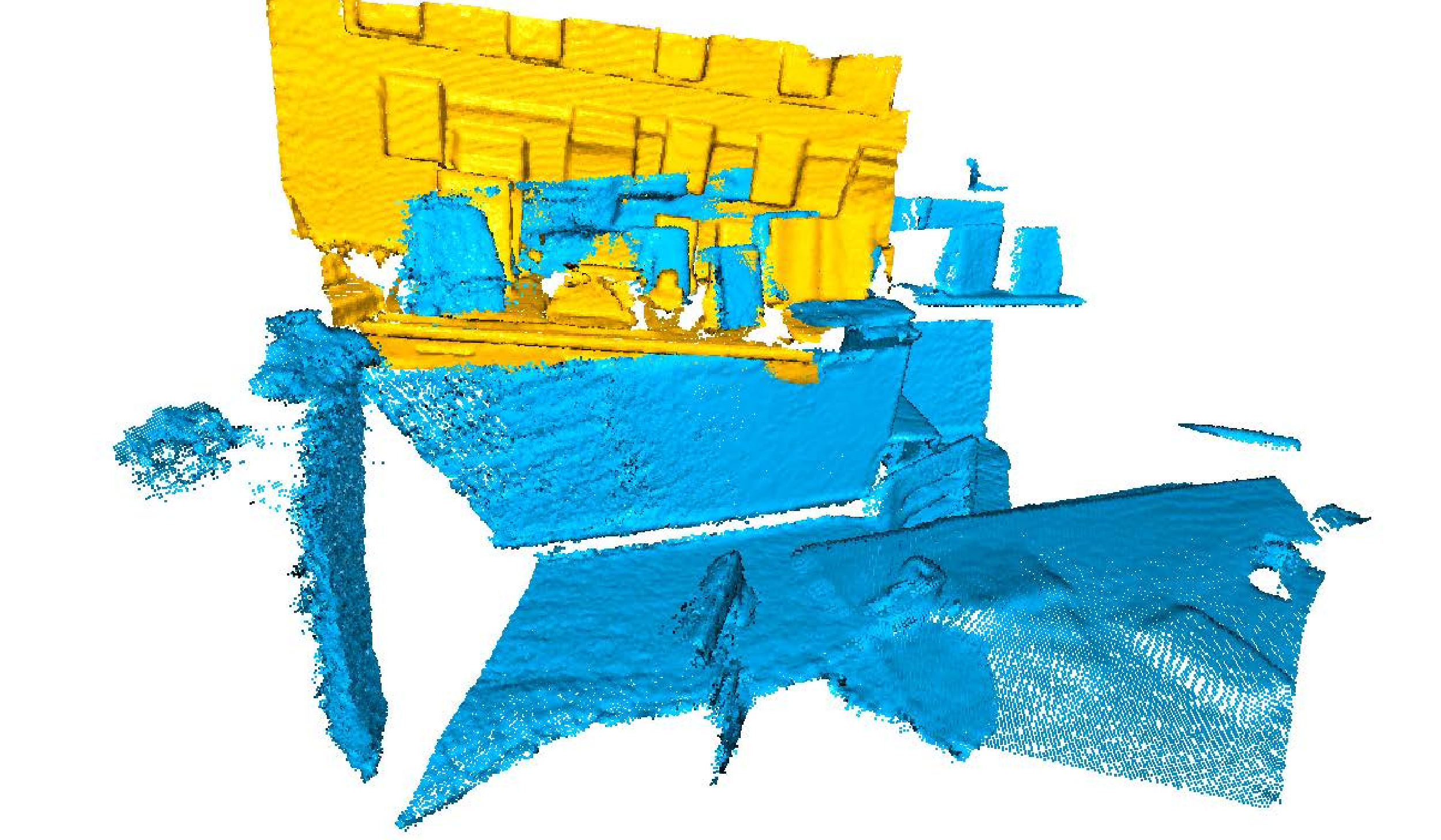} \\ (b-2) \textsf{RANSAC}-100k~\cite{fischler1981random} \\ $\textit{E}_\mathbf{R}$ = 8.44$^{\circ}$, \ $\textit{E}_\mathbf{t}$ = 41.17cm \\ t = \underline{2.71s} \\[0.5em] \includegraphics[width=0.12\textwidth]{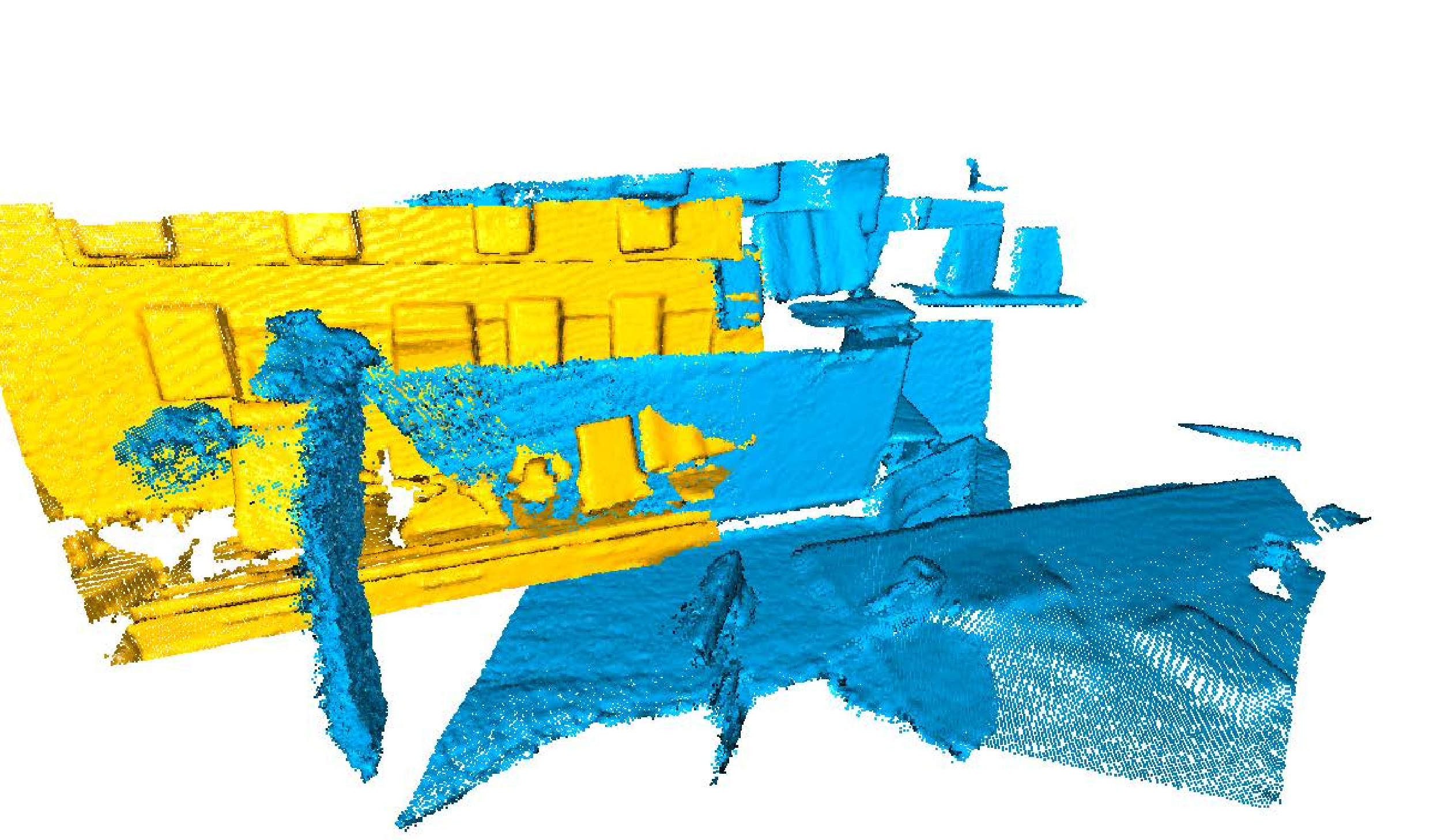} \\(a-5) \textsf{TR-DE}~\cite{chen2022deterministic} \\ $\textit{E}_\mathbf{R}$ = 24.17$^{\circ}$, \ $\textit{E}_\mathbf{t}$ = 43.29cm \\ t = 31.13s} &
      \makecell{\includegraphics[width=0.12\textwidth]{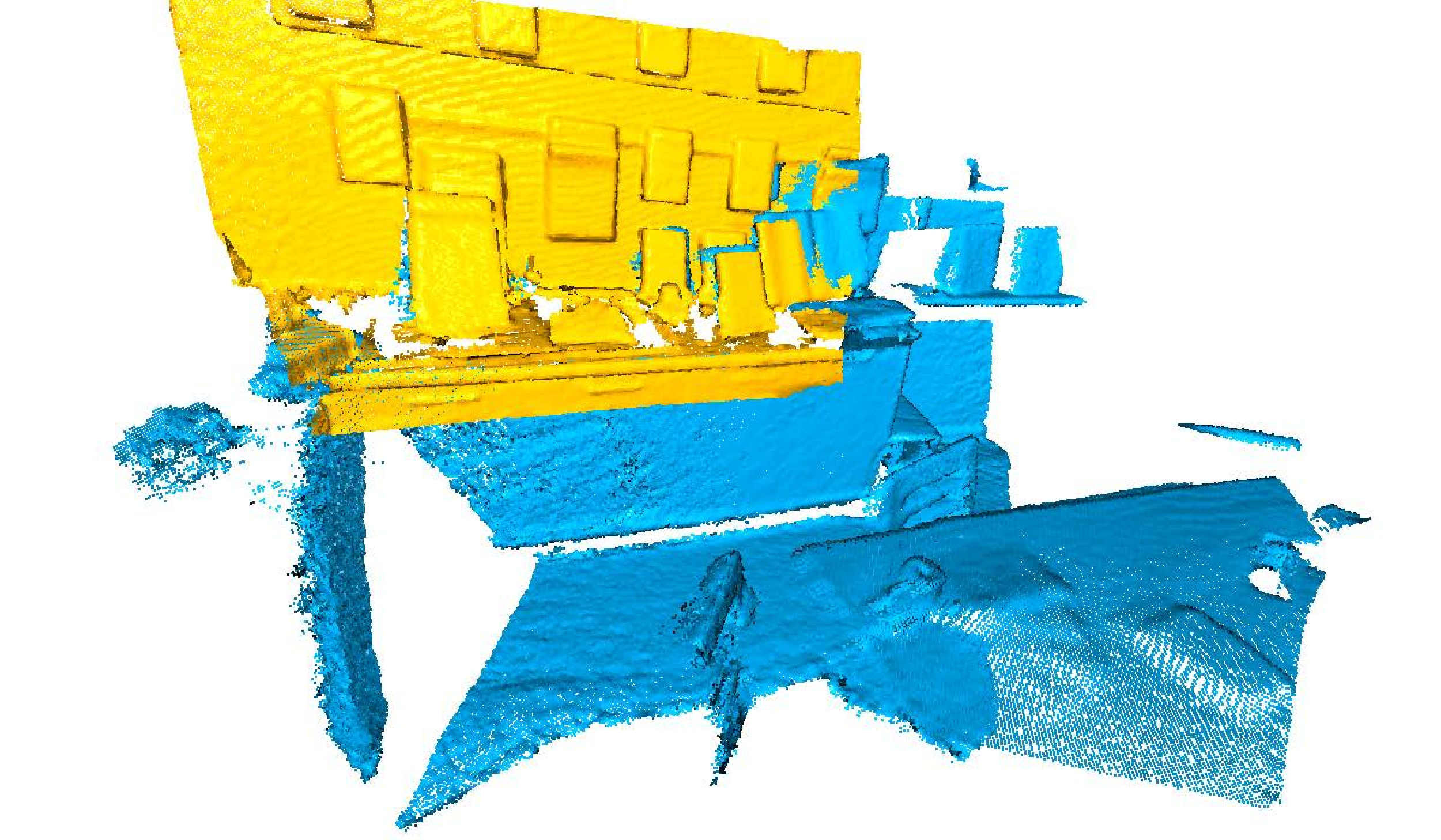} \\ (b-3) \textsf{GORE}~\cite{bustos2017guaranteed} \\ $\textit{E}_\mathbf{R}$ = 8.21$^{\circ}$, \ $\textit{E}_\mathbf{t}$ = 24.75cm \\ t = 6212.93s \\[0.5em]  \includegraphics[width=0.12\textwidth]{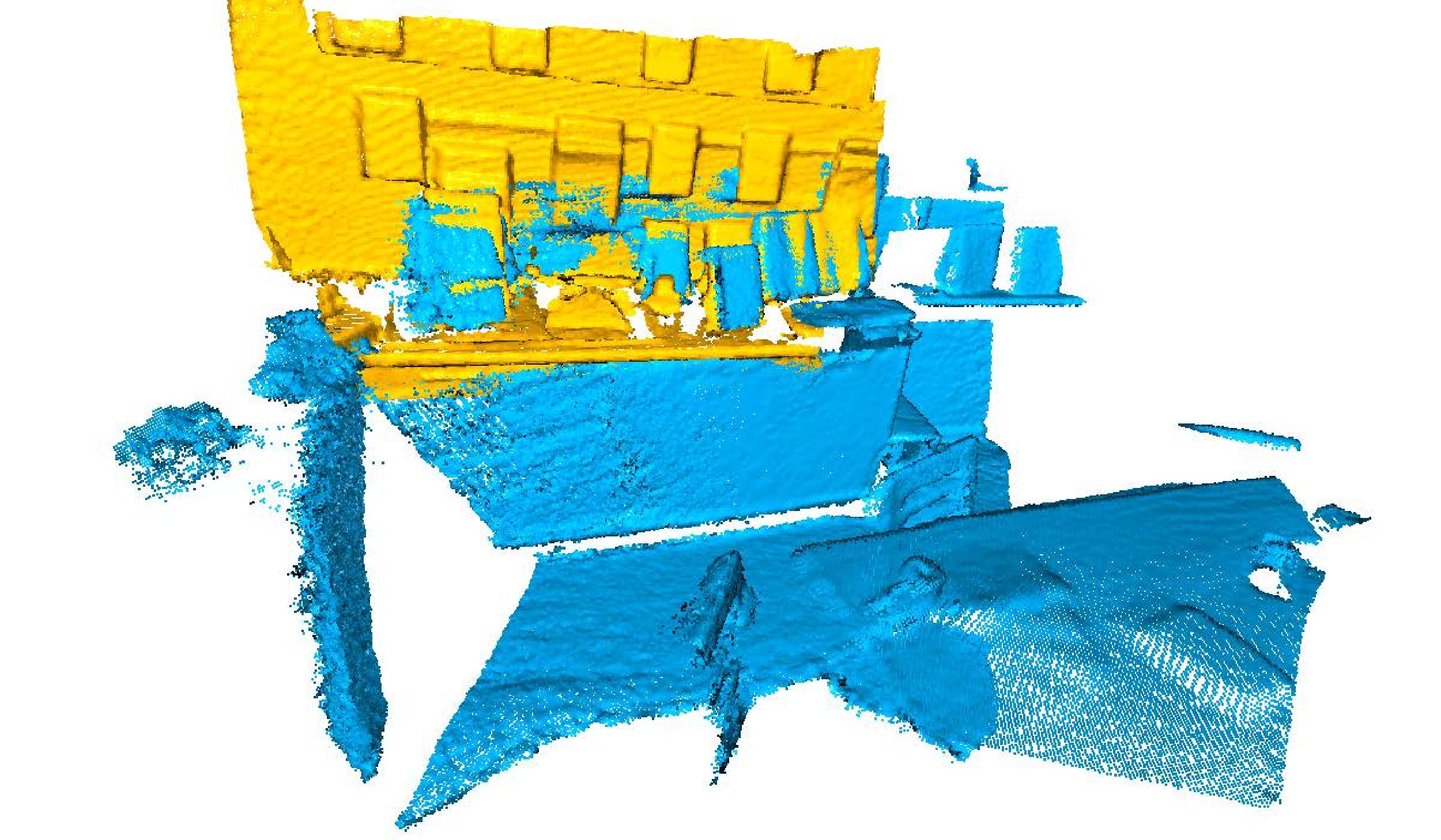} \\ (b-6) \textsf{SC$^2$-PCR}~\cite{chen2022sc2} \\ $\textit{E}_\mathbf{R}$ = \underline{2.79$^{\circ}$}, \ $\textit{E}_\mathbf{t}$ = 16.87cm \\ t = 3.86s} &
      \makecell{\includegraphics[width=0.12\textwidth]{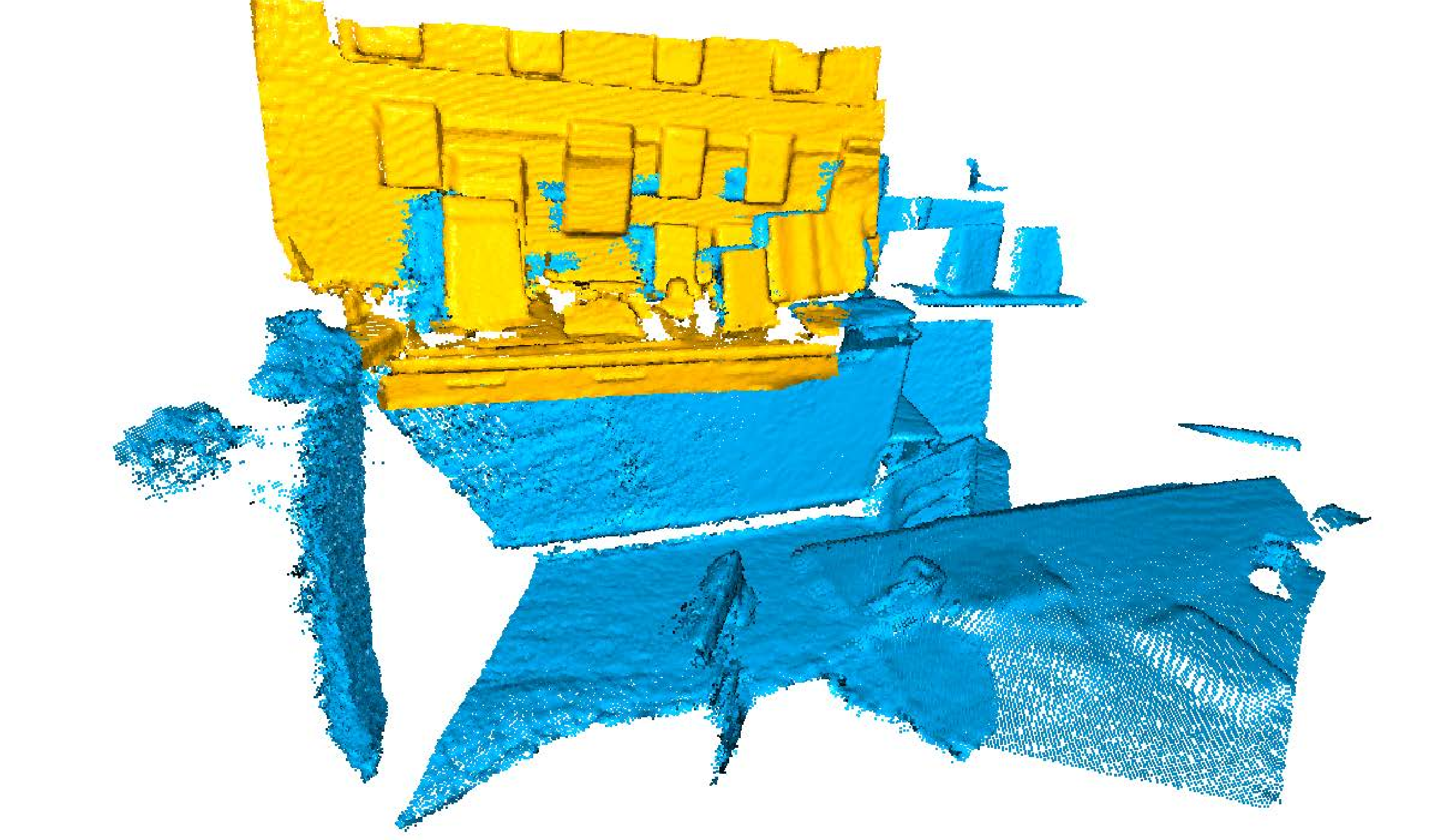} \\ (b-4) \textsf{TEASER++}~\cite{yang2020teaser} \\ $\textit{E}_\mathbf{R}$ = 3.24$^{\circ}$,\ $\textit{E}_\mathbf{t}$ = \textbf{14.37cm} \\ t = 58.23s \\[0.5em] \includegraphics[width=0.12\textwidth]{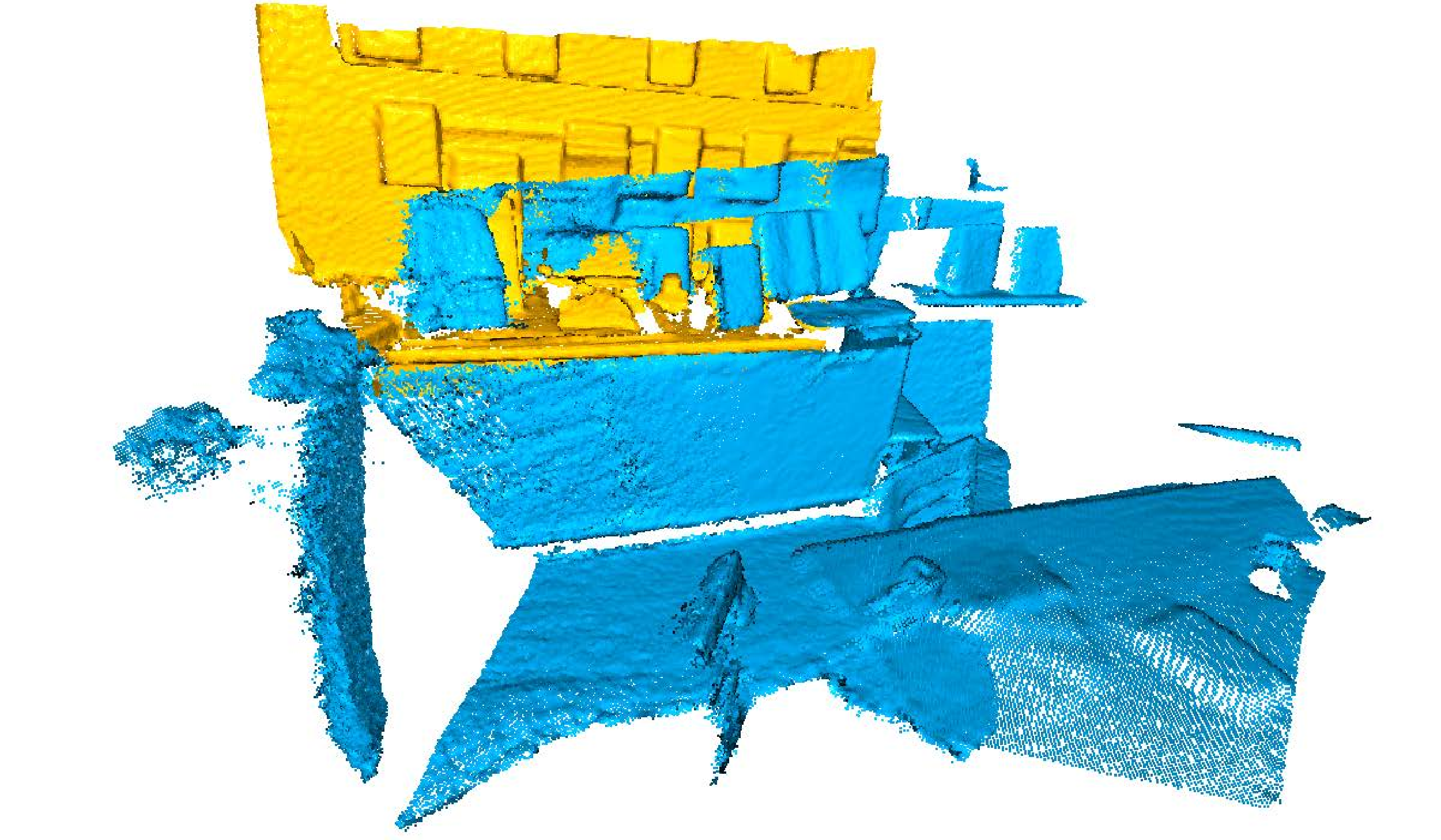} \\ (b-7) \textsf{HERE}~(ours) \\ $\textit{E}_\mathbf{R}$ = \ \textbf{2.41$^{\circ}$}, \ $\textit{E}_\mathbf{t}$ = \underline{15.11cm} \\ t = \textbf{0.31s}}
    \end{tabular}
    \vspace{-0.4em}
    \caption{Two representative comparison  cases on the indoor (a) 3DMatch~\cite{zeng20173dmatch} and (b) 3DLoMatch~\cite{huang2021predator} datasets. First column presents the input cloud pairs from different datasets. Second, third, and forth columns show the registration results of different methods.}
    \label{fig: quant_indoor}
\end{figure*}

Table.~\ref{tab: Kitti_nomutual} shows the evaluation results on the KITTI dataset. With FPFH descriptors, although \textsf{SC$^2$-PCR} has the highest \textit{RR}, the proposed \textsf{HERE} is more than 15 times faster and its \textit{RR} is only slightly lower. 
While \textsf{RANSAC}-1k and \textsf{FGR} shows high efficiency, their registration accuracy seems not very satisfactory, i.e., their \textit{RR}s are both lower than 20$\%$. The parameter search-based \textsf{TR-DE} shows satisfactory \textit{RR}, but it costs more than 130s. By contrast, \textsf{HERE} can not only achieve high robustness, but it also shows the highest efficiency among the methods with \textit{RR} larger than 50$\%$. For example, \textsf{HERE} runs more than 13 times faster than \textsf{RANSAC}-100k, 20 times faster than \textsf{MAC}, 470 times faster than \textsf{TR-DE}, and even faster than \textsf{RANSAC}-10k with 30$\%$ higher registration recall. In addition, \textsf{HERE} achieves the lowest $\textit{E}_\mathbf{t}$ and the highest \textit{IP}. With FCGF descriptors, the outlier ratios of the putative correspondence are relatively low~(averagely $\approx$ 58.7$\%$), therefore almost all methods lead to excellent results. Be that as it may, \textsf{HERE} outperforms other methods in terms of \textit{RR}, \textit{IP}, and \textit{$F_1$}. And for the methods with \textit{RR} $>$ 98$\%$, \textsf{HERE} shows higher efficiency. For example, \textsf{HERE} is more than 19 times faster than \textsf{RANSAC}-100k and more than 150 times faster than \textsf{TEASER++}. Fig.~\ref{fig: quant_outdoor}(b) shows a visual comparison case in the KITTI dataset, where \textsf{HERE} achieves the lowest $\textit{E}_\mathbf{t}$ with the lowest time cost compared with the other five baselines.

\subsubsection{Evaluation on Indoor Scenes}
We conduct comparison experiments on the 3DMatch~\cite{zeng20173dmatch} and 3DLoMatch~\cite{huang2021predator} datasets, both of which contain large-scale indoor scenes.

\noindent \textbf{3DMatch Dataset.}
We follow \cite{chen2022deterministic, chen2022sc2} to choose the test set of the 3DMatch Dataset~\cite{zeng20173dmatch} to evaluate the performance of our method on indoor scenes. We downsample the raw point clouds with a 5cm voxel grid and extract descriptors to match them based on a nearest-neighbor search. Following~\cite{chen2022sc2}, we adopt the FCGF~\cite{choy2019fully} as the feature descriptors. We set the inlier threshold $\xi=$ 10cm based on our downsampling. To evaluate successful registration, we set the threshold as $\textit{E}_\mathbf{R}\le15^\circ$ and $\textit{E}_\mathbf{t}\le30$cm. Note that we follow~\cite{yang2020teaser, chen2022deterministic} to use the correspondences with a reciprocal check for \textsf{TEASER++}, otherwise it costs an extremely long time for estimation~(up to 2h for some matching pairs).
We don't use the reciprocal check for all methods since this leads to similar efficiency performance and thus uninformative evaluation. We don't report the results of \textsf{GORE} since it is slow even after the reciprocal check.

As shown in Table~\ref{tab: 3DMatch_nomutual}, only \textsf{SC$^2$-PCR}, \textsf{MAC}, and \textsf{HERE} achieve more than 90$\%$ \textit{RR}s. While the \textit{RR}s of \textsf{SC$^2$-PCR} and \textsf{MAC} are slightly higher, \textsf{HERE} is 10 times faster. In addition, \textsf{HERE} outperforms all the compared methods in terms of $\textit{E}_\mathbf{R}$ and $F_1$-score. Regarding the efficiency comparison, while \textsf{RANSAC}-1k and \textsf{FGR} are the fastest, the \textit{RR} of \textsf{HERE} is 6.53$\%$ higher than \textsf{RANSAC}-1k and 17.81$\%$ higher than \textsf{FGR}. Compared to \textsf{TR-DE}, our method is more than 400 times faster and enjoys 4.57$\%$ higher \textit{RR}. While the reciprocal check largely reduces the time cost of \textsf{TEASER++}, its accuracy is compromised, with a 5.79$\%$ lower \textit{RR} than \textsf{HERE}. Fig.~\ref{fig: quant_indoor}(a) presents a powerful comparison case in the 3DMatch dataset, where \textsf{GORE} and \textsf{TEASER++} both lead to long time costs with the original correspondences, while \textsf{HERE} achieves the lowest $\textit{E}_\mathbf{R}$ with much less time cost.

\noindent \textbf{3DLoMatch Dataset.}
3DLoMatch~\cite{huang2021predator} is rearranged from the original 3DMatch dataset by choosing pairs with low overlapping ratio~(averagely lower than 30$\%$). We follow~\cite{chen2022deterministic, chen2022sc2} to adopt the Predator~\cite{huang2021predator} descriptors for putative correspondences. 

As shown in Table~\ref{tab: 3DMatch_nomutual}, although \textsf{MAC} has the highest \textit{RR}, the proposed \textsf{HERE} is more than 8 times faster and its \textit{RR} is only slightly lower. 
While \textsf{FGR} and \textsf{RANSAC}-1k show high efficiency, they yield 29.99$\%$ and 7.69$\%$ lower \textit{RR}s than \textsf{HERE}, respectively. 
In addition, compared to other methods with relatively satisfactory robustness~(i.e., with \textit{RR} higher than 66$\%$), \textsf{HERE} shows higher efficiency. For example, \textsf{HERE} is roughly 10 times faster than \textsf{RANSAC}-100k and \textsf{SC$^2$-PCR}, as well as 670 times faster than \textsf{TR-DE}. Fig.~\ref{fig: quant_indoor}(b) presents a representative comparison case in the 3DLoMatch dataset, where \textsf{HERE} achieves the lowest $\textit{E}_\mathbf{R}$ and time cost compared with the other five baselines.

\subsection{Analysis Experiments}
\label{subsec: ana_exper}


\subsubsection{Hyperparameter Setup}
\label{subsubsec: para_setup}

Our method has several hyperparameters: sampling numbers $k_{\mathbf{t}}$ and $k_{\mathbf{r}}$, and discretization numbers $m$ and $n$. At first glance, one might expect $k_{\mathbf{t}}$ and $k_{\mathbf{r}}$ to be large, so that more \textit{true inliers} get sampled, and similarly expect $m$ and $n$ to be large, so that the search space gets sufficiently covered. However, setting either of these parameters to be large would jeopardize the efficiency. In fact, as we will analyze using the KITTI dataset, there are deeper reasons for which all these parameters can be set unexpectedly small. With such analysis, we intend to rebuild some intuition that explains the efficiency of our method.

\begin{figure}[!t]
    \centering
    \includegraphics[width=0.485\textwidth]{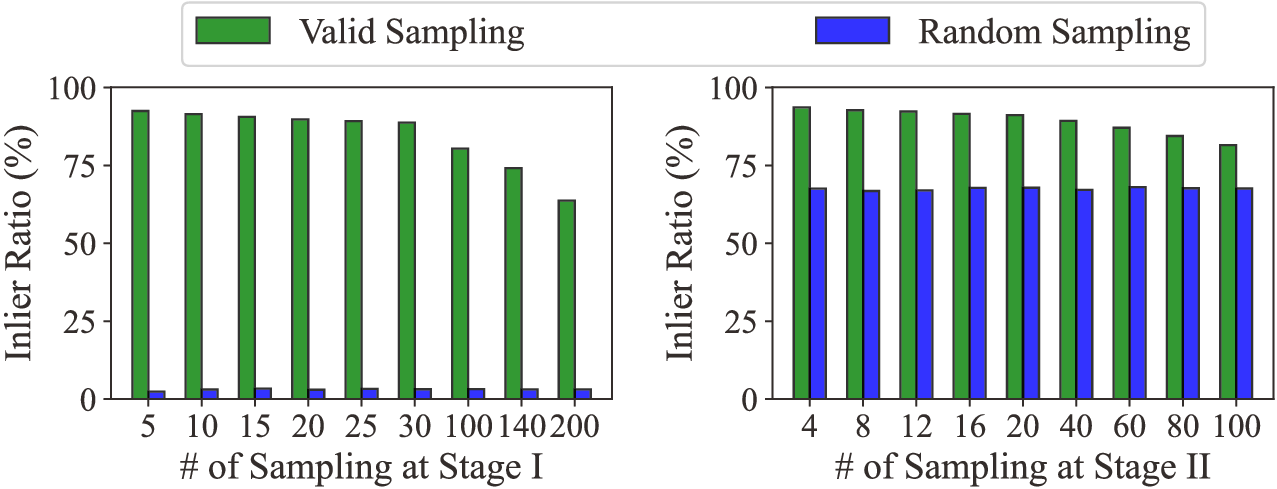}
    \\[-0.3em]
    \makebox[0.22\textwidth]{\footnotesize \quad \ (a)}
    \makebox[0.22\textwidth]{\footnotesize \quad \quad \quad \quad \ \ (b)}  
    \\[0.4em]
    \includegraphics[width=0.215\textwidth]{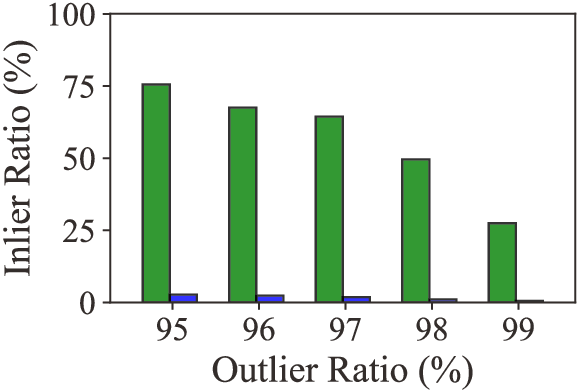}
    \\[-0.4em]
    \makebox[0.22\textwidth]{\footnotesize \quad \quad \ \ (c)}
    \\[-0.8em]
    \caption{Average Inlier ratios of the samples related to different sampling numbers and strategies at (a) stage I and (b) stage II of our method tested on the KITTI dataset~\cite{geiger2012we} with FPFH descriptors~\cite{rusu2009fast}. In addition, (c) reports the average inlier ratios of the samples~(\# = 15) related to different strategies tested on the simulated data with high outlier ratios.}
    \label{fig: abla_samp_inlier}
\end{figure}

\begin{figure*}[!t]
    \centering
    \includegraphics[width=0.975\textwidth]{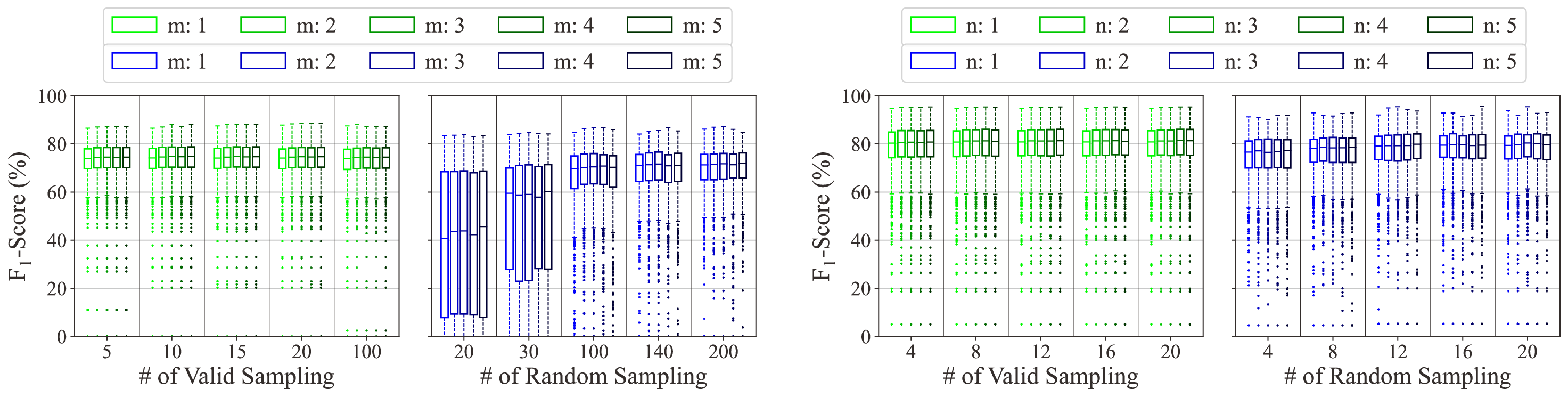}
    \\[-0.4em]
    \makebox[0.22\textwidth]{\footnotesize \quad \quad \quad \quad \quad (a-1)}
    \makebox[0.22\textwidth]{\footnotesize \quad \quad \quad \quad \ \ (a-2)}
    \makebox[0.08\textwidth]{\footnotesize }
    \makebox[0.22\textwidth]{\footnotesize \ \ (b-1)}
    \makebox[0.22\textwidth]{\footnotesize \ (b-2)}
    \\[-0.7em]
    \caption{Boxplots of $F_1$-score under different hyperparameter setups at (a) stage I and (b) stage II of our method tested on the KITTI dataset~\cite{geiger2012we} with FPFH descriptors~\cite{rusu2009fast}. The $F_1$-scores of our valid sampling and the basic random sampling are presented in green and blue, respectively.}
    \label{fig: abla_score}
\end{figure*}


\noindent \textbf{Hyperparameter Setup at Stage I.}
Recall that at stage I, we sample $k_\mathbf{t}$ correspondences and discretize the search region related to each sample to $m$ spherical surfaces. We select different parameter setups of $k_\mathbf{t}$ and $m$ to test our method. 

For the sampling part, we test both the proposed valid sampling strategy and the basic random sampling strategy. Fig.~\ref{fig: abla_samp_inlier}(a) shows the average inlier ratios of the sampled correspondences using different strategies on the KITTI dataset with FPFH descriptors. Our valid sampling strategy results in large inlier ratios even though the sampling number~(i.e., $k_\mathbf{t}$) is smaller than 10. By contrast, the correspondences chosen by random sampling contain few inliers due to sampling uncertainty. To evaluate how $k_\mathbf{t}$ affects the accuracy, Figs.~\ref{fig: abla_score}(a-1) and (a-2) show the $F_1$-score after stage I based on different hyperparameter setups and sampling strategies. 
Since valid sampling provides high inlier ratios, it also yields high $F_1$-score. In particular, its $F_1$-score with $k_\mathbf{t}$ = 5 are even larger than the $F_1$-score of randomly sampling with $k_\mathbf{t}$ = 200.
The reason is that more \textit{true inliers} in the sampled correspondences make the parameter search focus on the regions containing optimal solutions and sidestep the disturbance of outliers. 

To evaluate the performance of our valid sampling regarding high outlier ratios, we test it on the simulated dataset in Section~\ref{subsubsec: simu_extre} with $N=1000$ and the outlier ratios $\rho = \{95\%, 96\%, 97\%, 98\%, 99\%\}$. For each outlier ratio, we set the sampling number as 15 and conduct 100 Monte Carlo runs. As shown in Fig.~\ref{fig: abla_samp_inlier}(c), even with $99\%$ outliers, our valid sampling can lead to averagely more than $1/4$ inliers in the samples, which demonstrates its effectiveness. In addition, experiments in Section~\ref{subsubsec: simu_extre} further present the final registration results under high outlier ratios and prove the robustness of our method based on our valid sampling.



For the discretization part, Figs.~\ref{fig: abla_score}(a-1) and (a-2) show the results under different discretization numbers~(i.e., $m$). With $m$ growing, the $F_1$-score approximately go higher for both sampling strategies.
Even though $m$ is set to be very small~($m\le5$), the $F_1$-score of valid sampling are satisfactory and very stable as $m$ varies from 1 to 5.
The reason is that the thickness of the spherical shell in Fig.~\ref{fig: t_est}(a) is relatively small for the point cloud registration problem~(e.g., the spherical radius $\|\mathbf{x}_i\|$ versus spherical thickness $2\xi$ is averagely larger than 15 in KITTI dataset). Therefore, a few spherical surfaces~($m \le$ 3) are enough to approximate the spherical shell without affecting the accuracy. 
Accordingly, a small $m \le$ 3 can be applied with our valid sampling strategy, which further reduces the time cost of our method. 
  
\noindent \textbf{Hyperparameter Setup at Stage II.} Recall that at stage II, we sample $k_\mathbf{r}$ correspondences and discretize the search region related to each sample to $n$ half-circles. We  select different parameter configurations of $k_\mathbf{r}$ and $n$ for evaluation.

Thanks to the preliminary outlier removal at stage I, the correspondence set passed to stage II contains a large fraction of inliers. Therefore, even random sampling can lead to large fractions of inliers in the sampled correspondences~(see Fig.~\ref{fig: abla_samp_inlier}(b)). However, the proposed valid sampling strategy still achieves higher inlier ratios. Consequently, as shown in Figs.~\ref{fig: abla_score}(b-1) and (b-2), the valid sampling enjoys higher $F_1$-score than the random sampling with the same $k_\mathbf{r}$. In addition, the discretization number $n$ also shows little to no effect on the estimation performance similar to stage I. The reason is that the large fraction of inliers in the sampled correspondences leads to most corresponding search regions containing/near the optimal solution, which eliminates the influence that the optimal solution region is discarded in some search regions due to discretization. 

\begin{figure}[!t]
    \centering
    \includegraphics[width=0.42\textwidth]{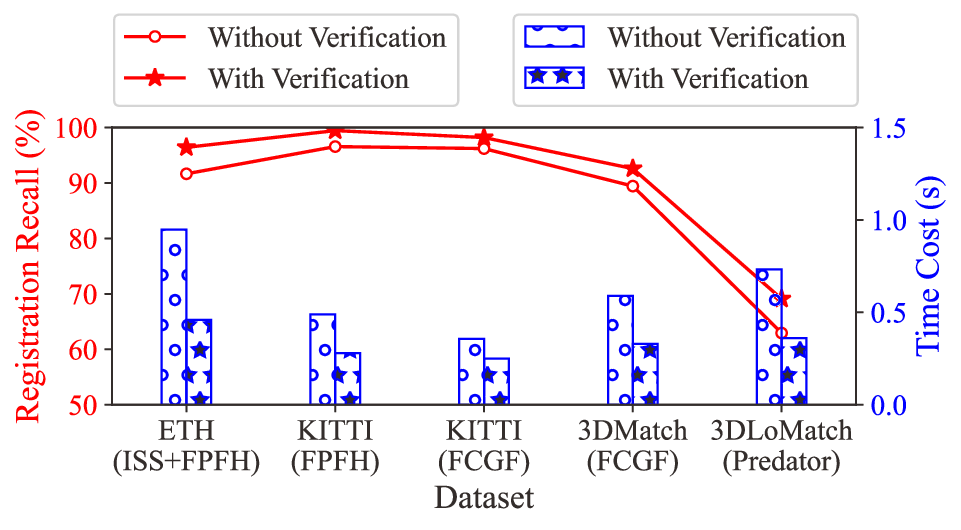}
    \\[-1.4em]
    \caption{Ablation study regarding the compatibility verification strategy on various real-world datasets with different feature descriptors. }
    \label{fig: abla_verfi}
\end{figure}

\subsubsection{Ablation Study on Compatibility Verification}
\label{subsubsec: verfi}
Recall that we design a compatibility verification strategy for our method to reduce the computation cost~(see Section~\ref{subsec: com_interval}). To evaluate its effect, we conduct an ablation study for this design on the five real-world datasets introduced in Section~\ref{subsec: exper_real}. Fig.~\ref{fig: abla_verfi} presents the comparison results on the five datasets regarding the registration recall and time cost. For all five datasets, our design results in higher registration recall. The reason is that it introduces the spatial compatibility constraint and additionally rejects outliers for the original consensus maximization problem. In addition, the verification largely increases the efficiency of our method, e.g., the average time costs in the ETH, 3DMatch, and 3DLoMatch datasets are roughly reduced by half. This benefits from the fact that spatial compatibility avoids our search part from computing intervals for those correspondences that are not compatible with the sampled correspondences. Above all, the proposed compatibility verification improves our method in terms of both robustness and efficiency. 

\section{Conclusion and Future Work}
\label{sec: conc}
In this work, we propose a heuristics-guided parameter search strategy and a three-stage decomposition pipeline to achieve efficient and robust rigid point cloud registration.
Compared to existing methods, our strategy largely reduces the search space and can guarantee accuracy with only a few inlier samples (i.e., heuristics), therefore enjoying a great trade-off between efficiency and robustness.
Note that the current implementation of our method is based on C++ and utilizes the multi-thread parallelism on CPUs, we look forward to extend it to run on the GPUs for higher efficiency.
In addition, our strategy also holds great potential for solving other geometry-based pose estimation problems~\cite{li2020quasi, yang2022certifiably, 9528069}, which will be left as our future work.

\ifCLASSOPTIONcompsoc
  \section*{Acknowledgments}
\else
  \section*{Acknowledgment}
\fi
This work is supported in part by the Shenzhen Portion of Shenzhen-Hong Kong Science and Technology Innovation Cooperation Zone under Grant HZQBKCZYB-20200089, in part by the Hong Kong RGC under T42-409/18-R and 14207119, in part by the Hong Kong Centre for Logistics Robotics, and in part by the VC Fund 4930745 of the CUHK T Stone Robotics Institute.

\ifCLASSOPTIONcaptionsoff
  \newpage
\fi

\bibliographystyle{IEEEtran}
\bibliography{papers}

\clearpage
\appendices

\section{Proof of $\mathbf{R}^{\top}\mathbf{r} = \mathbf{r}$}
Recall that in Section~\textcolor{red}{4.2} of the main manuscript we derive the rotation axis constraint~(i.e., Eq. (\textcolor{red}{7c})) for the second decomposed sub-problem by leveraging the fact $\mathbf{R}^{\top}\mathbf{r} = \mathbf{r}$. In the following we present the proof.

Given a rotation matrix $\mathbf{R}\in$ \textit{SO}(3) with its related rotation axis $\mathbf{r}\in\mathbb{S}^2$ and rotation angle $\theta\in$ [0, $\pi$], for any $\mathbf{x} \in \mathbb{R}^3$, $\mathbf{R}\mathbf{x}$ means rotating the vector $\mathbf{x}$ about the axis $\mathbf{r}$ with the angle $\theta$. Since rotating the axis $\mathbf{r}$ about itself with any angle only results in $\mathbf{r}$, we have $\mathbf{R}\mathbf{r} = \mathbf{r}$. Furthermore, $\mathbf{R}^{\top} = \mathbf{R}^{-1}$ defines the inversion of $\mathbf{R}$. Therefore, the rotation axis and rotation angle of $\mathbf{R}^{\top}$ can be defined by $\mathbf{r}$ and $-\theta$, i.e., rotating about the same axis with the same angle as $\mathbf{R}$ but in inverse direction. Since rotating the axis $\mathbf{r}$ about itself with angle $-\theta$ results in $\mathbf{r}$, we have $\mathbf{R}^{\top}\mathbf{r} = \mathbf{r}$.

\section{Comparison on Theoretical Time Complexity}
The proposed method \textsf{HERE} achieves quadratic time complexity with respect to the correspondence number $N$~(see Section \textcolor{red}{7} of the main manuscript). In the following, we compare it with several highly-robust baseline methods.

Consider two parameter search-based baselines~(most related): \textsf{GORE} has $O(N^3\log N)$ time complexity even without the BnB-based search~\cite[Algorithm 2]{bustos2017guaranteed} since its decomposition increases the problem size quadratically; the recent \textsf{TR-DE}~\cite{chen2022deterministic} leads to $O(2\frac{1}{\psi^3}N)$ time complexity~(note that $\frac{1}{\psi^3}=$1e$^{9}$), where the third power results from its BnB-based search on 3D parameter spaces. In contrast, our decomposition doesn't increase the problem size, and we only conduct BnB-based search on 1D parameter space at stage~I; therefore experiments in Section \textcolor{red}{8} of the main manuscript show that \textsf{HERE} can exhibit up to $10^2\times$ and sometimes exceeding $10^3\times$ speed-up than \textsf{GORE} and \textsf{TR-DE}.

As to \textsf{TEASER++}~\cite{yang2020teaser} and \textsf{MAC}~\cite{Zhang_2023_CVPR}, their operations on the compatibility graph can both lead to exponential time complexity in the worst case: theoretically, mining the maximum clique~(\textsf{TEASER++}) has $O(2^{N/4})$ time complexity~\cite{wu2015review, robson2001finding} and listing all maximal cliques~(\textsf{MAC}) has $O(3^{N/3})$ time complexity~\cite{Moon-IJM1965,Tomita-TCS2006}. Even though certain strategies~(e.g., multi-thread parallelism and graph sparsification) have been adopted in \textsf{TEASER++} and \textsf{MAC} for efficiency improvement, experiments in Section \textcolor{red}{8} of the main manuscript show that our \textsf{HERE} is faster.

In \textsf{SC$^2$-PCR}~\cite{chen2022sc2}, constructing the second-order consistency matrix has $O(N^2 + \rho N^3)$ time complexity, in which $O(N^2)$ is for the first-order matrix, and $O(\rho N^3)$ is for the second-order matrix where $\rho\in[0, 1]$ is a selection ratio~\cite[Section 3.3]{chen2022sc2}. The cubic time complexity here leads to a dozen times slower of \textsf{SC$^2$-PCR} than the proposed \textsf{HERE} with quadratic time complexity in both simulated and real-world datasets~(see Section \textcolor{red}{8} of the main manuscript).

\section{Comparison regarding Different Decomposition Methods}

The proposed method \textsf{HERE} introduces a three-stage decomposition pipeline. Since previous works also present some different decomposition methods, we compare these methods in the following:
\begin{itemize}
    \item \textsf{GORE} adopts exhaustive sampling to decouple the original 6-DoF problem into $N$ sub-problems only concerning the rotation. They solve the sub-problems one by one followed by computing the translation related to each sampled correspondence and choose the transformation with the largest consensus set.
    \item \textsf{TEASER++} leverages the Translation Invariant Measurements~(TIMs) to decouple the 6-DoF problem into a 3-DoF rotation sub-problem followed by a 3-DoF translation sub-problem. Since TIMs require pairwise constraints, their decomposition makes the size of their first sub-problem quadratically larger. 
    \item \textsf{TR-DE} decouples the 6-DoF problem into two 3-DoF sub-problems without increasing the problem size. The former sub-problem is related to the 2-DOF rotation axis and 1-DoF displacement along this rotation axis; the latter is related to the 1-DOF rotation angle and 2-DOF displacement orthogonal to the above rotation axis.
    \item  Our \textsf{HERE} decouples the 6-DoF problem into three lower-dimensional sub-problems with respect to the 3-DoF translation, 2-DoF rotation axis, and 1-DoF rotation angle, respectively~(details can be found in Section \textcolor{red}{4.2} of the main manuscript).
\end{itemize}
Compared to \textsf{GORE} and \textsf{TEASER++}, all three sub-problems in our decomposition pipeline avoid increasing the problem size, which ensures higher solving efficiency. Compared to \textsf{TR-DE}, the parameter space in our decomposed sub-problems is either linear or spherical and can be easily formulated for efficient search. In contrast, the constraints in both sub-problems of \textsf{TR-DE} are highly non-linear, therefore searching related parameter spaces for solutions is generally time-consuming. In addition, our decomposition pipeline enables the application of our heuristics-guided parameter search while other methods could not, which leads to the higher efficiency of our \textsf{HERE}. Experiments in Section \textcolor{red}{8} of the main manuscript show that our \textsf{HERE} can be more than two orders of magnitude faster than \textsf{GORE} and \textsf{TR-DE}, and several times faster than \textsf{TEASER++} while achieving comparable or even higher robustness.

\section{Additional Experimental Results of Different Sampling Strategies}
Recall that in Section~\textcolor{red}{6.3.1} of the main manuscript, we report the comparison results between the basic random sampling strategy and our valid sampling strategy. Based on spatial compatibility, some methods are proposed to directly employ the number of compatible correspondences~(i.e., the correspondence score in Eq.~(\textcolor{red}{24})) as the sampling metrics~\cite{9373914, quan2020compatibility}. We compare our valid sampling with the score-based sampling for further evaluation.

\begin{figure}[!t]
    \centering
    \includegraphics[width=0.485\textwidth]{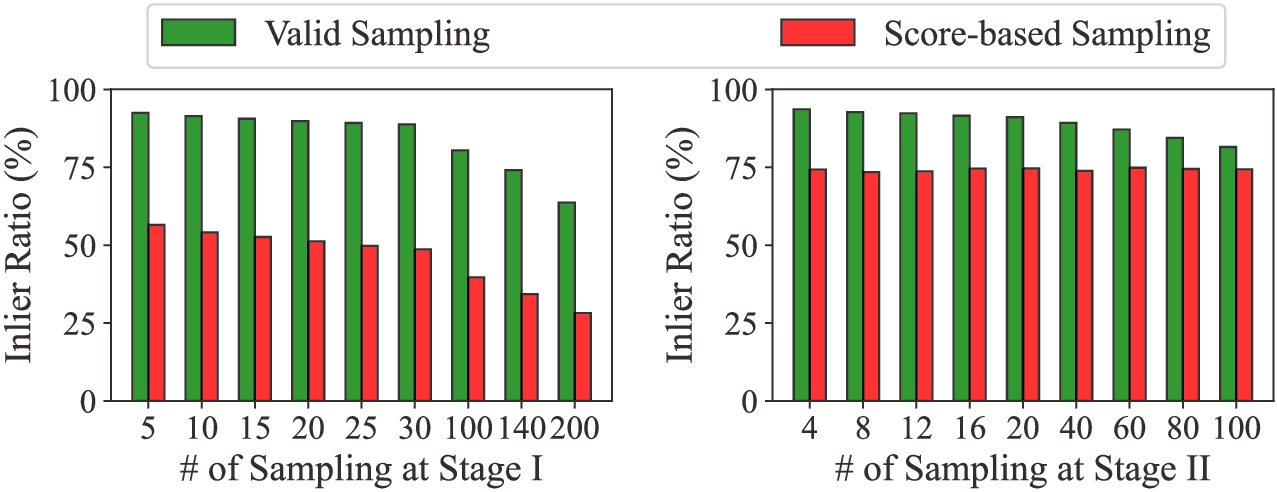}
    \\[-0.3em]
    \makebox[0.22\textwidth]{\footnotesize \quad \ (a)}
    \makebox[0.22\textwidth]{\footnotesize \quad \quad \quad \quad \ \ (b)}  
    \\[0.4em]
    \includegraphics[width=0.22\textwidth]{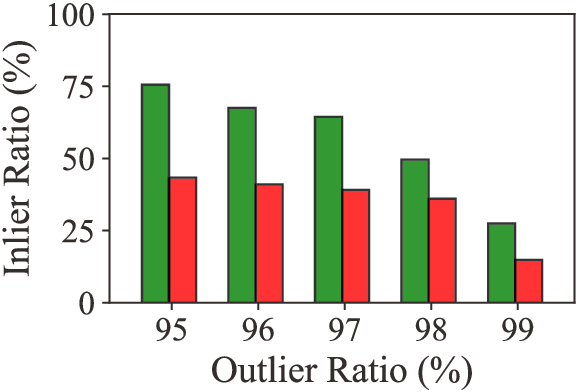}
    \\[-0.4em]
    \makebox[0.22\textwidth]{\footnotesize \quad \quad \ \ (c)}
    \\[-0.8em]
    \caption{Average Inlier ratios of the samples related to different sampling numbers and strategies at (a) stage I and (b) stage II of our method tested on the KITTI dataset~\cite{geiger2012we} with FPFH descriptors~\cite{rusu2009fast}. In addition, (c) reports the average inlier ratios of the samples~(\# = 15) related to different strategies tested on the simulated data with high outlier ratios.}
    \label{fig: abla_samp_inlier_app}
\end{figure}

As shown in Fig.~\ref{fig: abla_samp_inlier_app}, while score-based sampling can select more inliers compared to random sampling~(see Fig.~\textcolor{red}{10} of the main manuscript), our valid sampling shows better performance. Specifically, the average inlier ratios under valid sampling are higher than that of score-based sampling at both stages and sometimes are more than 2 times higher at stage I. The reason is that our valid sampling additionally considers the scores of the compatible correspondences compared to the pure score-based sampling, therefore enjoying better leverage of the spatial compatibility constraint. To evaluate the performance of both sampling strategies regarding high outlier ratios, we test them on the simulated dataset in Section~\textcolor{red}{8.1.2} with $N=1000$ and the outlier ratios $\rho = \{95\%, 96\%, 97\%, 98\%, 99\%\}$. For each outlier ratio, we set the sampling number as 15 and conduct 100 Monte Carlo runs. As shown in Fig.~\ref{fig: abla_samp_inlier_app}(c), the inlier ratios of our valid sampling is roughly 2 times higher than that of score-based sampling, which demonstrates the priority of our valid sampling. 

\begin{figure}[!t]
    \centering
    \includegraphics[width=0.485\textwidth]{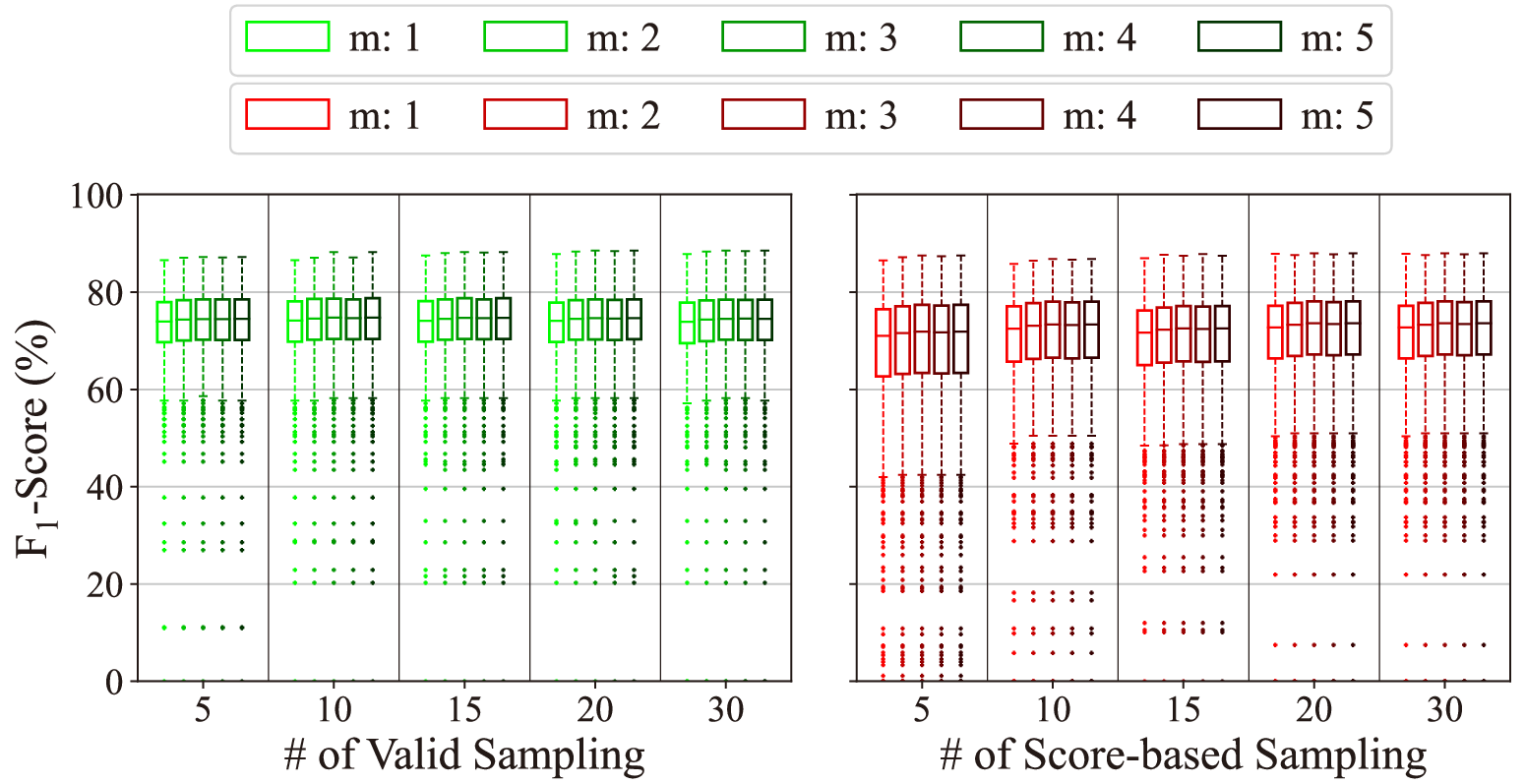}
    \\[-0.3em]
    \makebox[0.22\textwidth]{\footnotesize (a)}
    \\[0.6em]
    \includegraphics[width=0.485\textwidth]
    {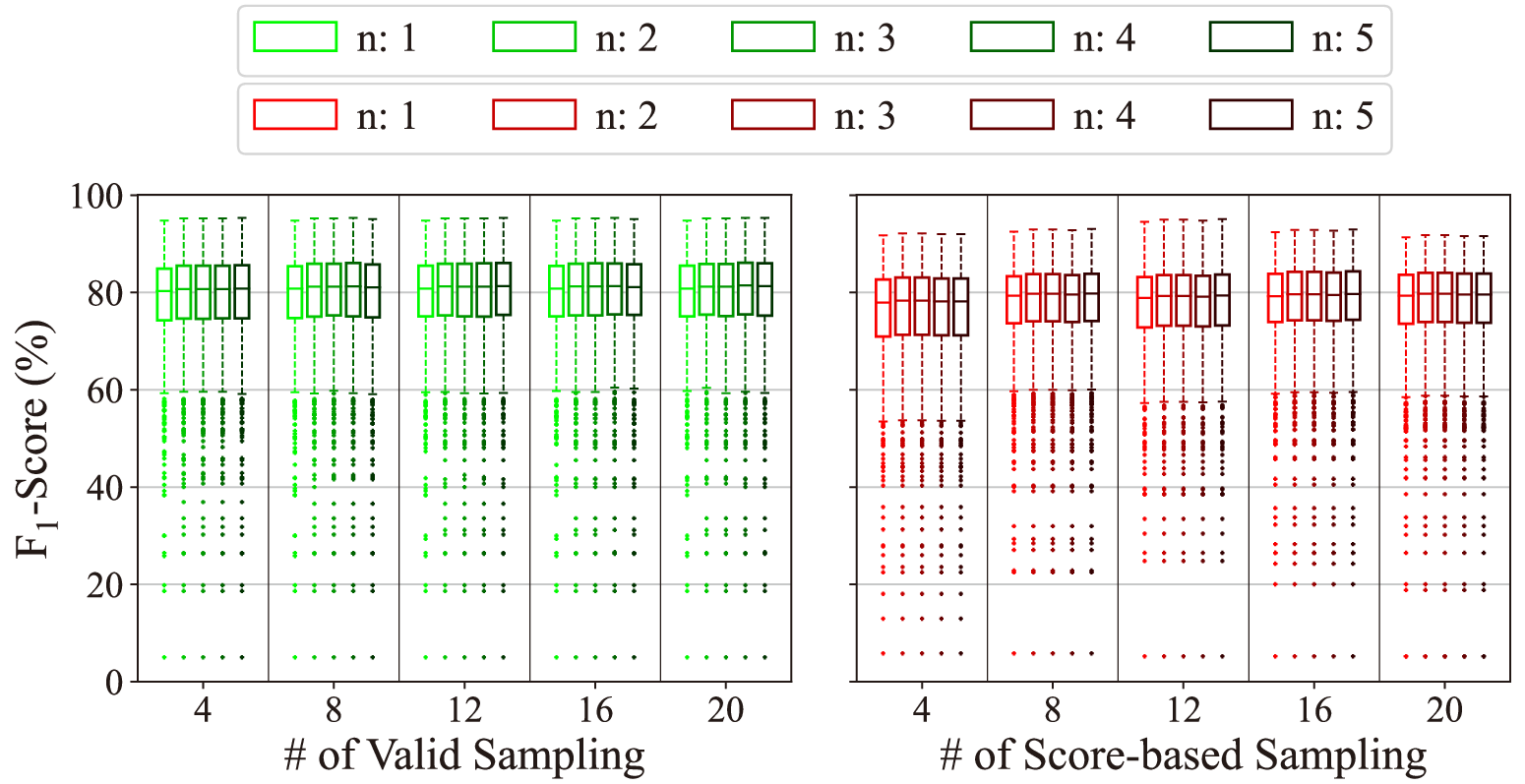}
    \\[-0.3em]
    \makebox[0.22\textwidth]{\footnotesize (b)}
    \\[-0.6em]
    \caption{Boxplots of $F_1$-score under different hyperparameter setups at (a) stage I and (b) stage II of our method tested on the KITTI~\cite{geiger2012we} dataset with FPFH~\cite{rusu2009fast} descriptors. The $F_1$-scores of our valid sampling and the score-based sampling are presented in green and red, respectively.}
    \label{fig: abla_samp_score_app}
\end{figure}

As a consequence, at both stages I and II, our valid sampling results in higher $F_1$-score compared to scored-based sampling under the same sampling numbers~(see Fig.~\ref{fig: abla_samp_score_app}). These results further demonstrate the superiority of the proposed valid sampling strategy. In addition, similarly to the results in Fig.~\textcolor{red}{10} of the main manuscript, the discretization numbers $m$ and $n$ both show little to no effect on the estimation performance of score-based sampling.

\section{Performance of Each Module}
\label{subsubsec: perf_each}
Our method can be roughly divided into five modules: 1) priority computation via spatial compatibility~(see Section~\textcolor{red}{6.1}), 2) outlier removal at stage I regarding the translation constraint~(see Section~(see Section~\textcolor{red}{5.1}), 3) outlier removal at stage II regarding the rotation axis constraint~(see Section~(see Section~\textcolor{red}{5.2}), 4) outlier removal at stage III regarding the rotation angle~(see Section~(see Section~\textcolor{red}{5.3}), and 5) post-refinement based on SVD on the final consensus set. We report the performance of each module on the five real-world datasets introduced in Section~\textcolor{red}{8.2}.

\begin{figure}[!t]
    \centering
    \includegraphics[width=0.45\textwidth]{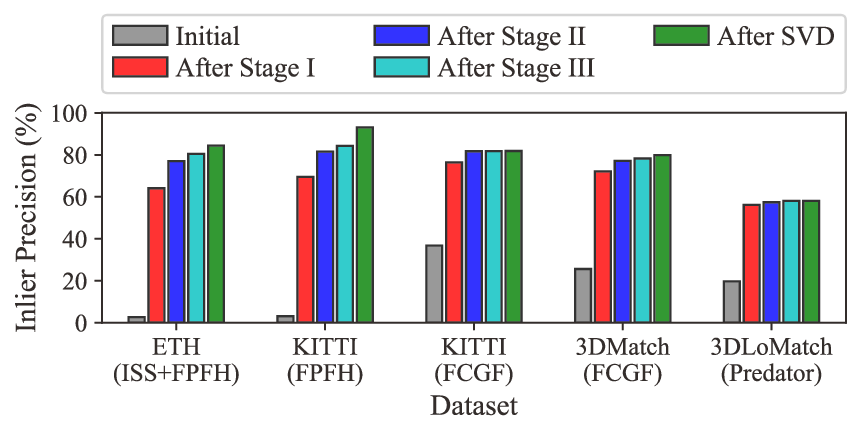}
    \\[-0.7em]
    \makebox[0.2\textwidth]{\footnotesize \quad \quad \ \ (a)}
    \\[0.2em]
    \includegraphics[width=0.45\textwidth]{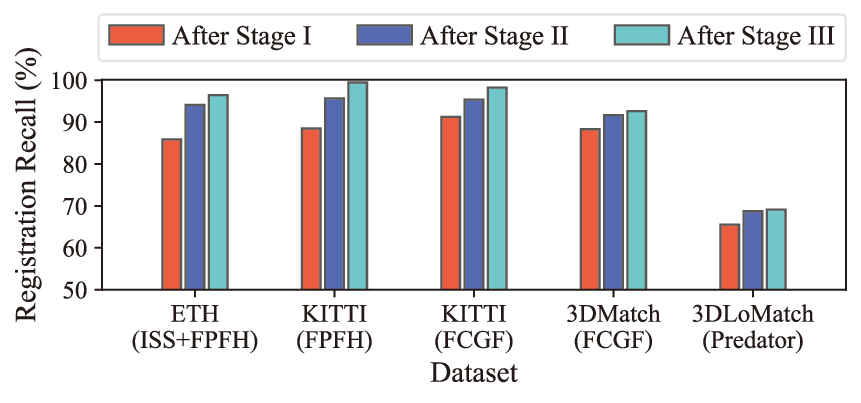}
    \\[-0.7em]
    \makebox[0.2\textwidth]{\footnotesize \quad \quad \ \ (b)}
    \\[-0.5em]
    \caption{Performance of each module in our method on various real-world datasets with different feature descriptors. (a) Inlier precision after each module of our method. Inlier precision of the ``Initial'' module represents the inlier ratio of the putative correspondences. (b) Registration recalls of the three stages in our method. We apply SVD on the correspondence sets after each stage separately to compute the recalls for evaluation.}
    \label{fig: abla_each_acc}
\end{figure}

\begin{figure}[!t]
    \centering
    \includegraphics[width=0.47\textwidth]{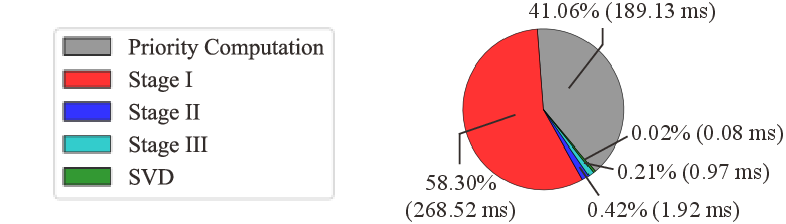}
    \\
    \makebox[0.2\textwidth]{\footnotesize }
    \makebox[0.2\textwidth]{\footnotesize (a) ETH (ISS+FPFH)}
    \\[0.8em]
    \includegraphics[width=0.47\textwidth]{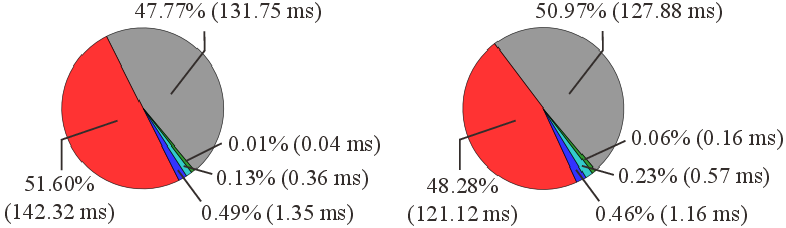}
    \\
    \makebox[0.2\textwidth]{\footnotesize (b) KITTI (FPFH) \ \ }
    \makebox[0.25\textwidth]{\footnotesize (c) KITTI (FCGF)}
    \\[0.8em]
    \includegraphics[width=0.47\textwidth]{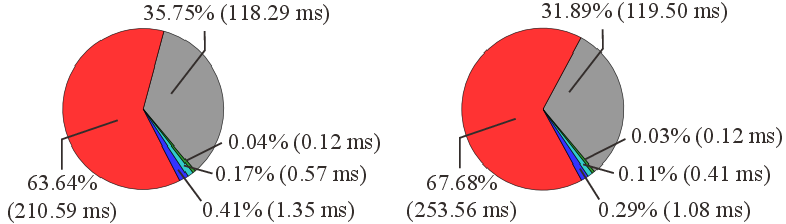}
    \\
    \makebox[0.2\textwidth]{\footnotesize (b) 3DMatch (FCGF) \ \ }
    \makebox[0.25\textwidth]{\footnotesize (c) 3DLoMatch (Predator)}
    \\[-0.6em]
    \caption{Time cost distribution of each module in our method tested on various real-world datasets with different feature descriptors.}
    \label{fig: abla_each_time}
\end{figure}

Figs.~\ref{fig: abla_each_acc} and~\ref{fig: abla_each_time} show the inlier precision, registration recall, and time cost after different modules/stages of our method on the five datasets, respectively. For all datasets, most outliers have been rejected after stage I, and then stages II and III play the role of further refinement~(see Fig.~\ref{fig: abla_each_acc}(a)). Accordingly, stage I takes up roughly the most time cost of our method while the time costs of stages II and III are smaller than 2ms~(see Fig.~\ref{fig: abla_each_time}). In other words, our decomposition and progressive outlier removal reduce the pressure of searching the original 6-DoF parameter space to searching only 3-DoF parameter space. In addition, thanks to our heuristics-guided parameter search and our leverage of compatibility constraints, the time cost of stage I is controlled to be less than 0.3s on average. 
This is the cornerstone of the high efficiency of our method.
Moreover, while the outlier removal at stages II and III seems to be much weaker than stage I, its effect in increasing the registration recall is non-negligible~(see Fig.~\ref{fig: abla_each_acc}(b)).  For example, additionally applying the outlier removal in stages II and III leads to $9.53\%$ higher \textit{RR} in the ETH dataset, $7.96\%$ higher \textit{RR} in the KITTI dataset with FPFH descriptors, and $8.10\%$ higher \textit{RR} in the KITTI dataset with FCGF descriptors compared to only with the outlier removal in stage I. Such improvements make the robustness of our method comparable to or even higher than the state-of-the-art methods~(see Section \textcolor{red}{8.2} of the main manuscript). 

Based on the progressive outlier removal at all three stages, our method achieves excellent accuracy comparable to the state-of-the-art methods and shows much higher efficiency, which has been introduced in Section~\textcolor{red}{8.2}. As to the priority computation module, its time cost is highly relevant to the correspondence set size. While this module takes up roughly the second most time cost in our method, it guarantees the effectiveness of our valid sampling, and further the excellent robustness and efficiency of our method.

\begin{table}[!t]
\centering
    \caption{Detailed information of the ETH dataset~\cite{theiler2014keypoint} with ISS~\cite{zhong2009intrinsic} + FPFH~\cite{rusu2009fast} descriptors. $N$ and $\rho$ in the table represent the number of correspondences and the outlier ratio, respectively.}
    \vspace{-0.6em}
    \label{tab: eth_info}
    \footnotesize
    \renewcommand{\tabcolsep}{4.2pt} 
    \renewcommand\arraystretch{1.4}
    \begin{tabular}{c|ccccc}
        \Xhline{1pt}
        Scene & Arch & Courtyard & Facade & Office & Trees \\
        \Xhline{0.5pt}
        Overlap & 30$\%$-40$\%$ & 40$\%$-70$\%$ & 60$\%$-70$\%$ & $>$80$\%$ & $\approx$50$\%$ \\
        $\#$ of Pairs & 10 & 28 & 21 & 10 & 15\\
        Average of $N$ & 8892 & 9877 & 1356 & 912 & 14987\\
        Average of $\rho$ & 99.44$\%$ & 96.23$\%$ & 95.86$\%$ & 97.92$\%$ & 99.70$\%$\\
        \Xhline{1pt} 
    \end{tabular}
\end{table}

\begin{table}[!t]
\centering
    \caption{Detailed information of the KITTI~\cite{geiger2012we}, 3DMatch~\cite{zeng20173dmatch}, and 3DLoMatch~\cite{huang2021predator} datasets with various descriptors.}
    \vspace{-0.6em}
    \label{tab: other_info}
    \footnotesize
    \renewcommand{\tabcolsep}{3.8pt} 
    \renewcommand\arraystretch{1.4}
    \begin{tabular}{c|ccccc}
        \Xhline{1pt}
        Dataset & Descriptors & $\#$ of Pairs & Average of $N$ & Average of $\rho$ \\
        \Xhline{0.5pt}
        KITTI & FPFH & 555 & 5000 & 96.93$\%$ \\
        KITTI & FCGF & 555 & 5000 & 63.24$\%$ \\
        3DMatch & FCGF & 1623 & 4656 & 74.38$\%$ \\
        3DLoMatch & Predator & 1781 & 4942 & 80.27$\%$\\
        \Xhline{1pt} 
    \end{tabular}
\end{table}

\section{Details of Real-world Datasets}
We present detailed information on various real-world datasets used for evaluation in this paper in Tables~\ref{tab: eth_info} and~\ref{tab: other_info}.

\end{document}